\documentclass[review]{elsarticle}

\usepackage{lineno,hyperref}
\modulolinenumbers[5]

\usepackage{multirow}
\usepackage{multicol}
\usepackage{wrapfig}
\usepackage{array}
\usepackage{amsmath}
\usepackage{color}
\usepackage{graphicx}
\usepackage[caption=false]{subfig}
\usepackage{dirtytalk}
\usepackage{mathrsfs}
\usepackage{textcomp}
\usepackage{booktabs}
\usepackage{amsfonts}
\usepackage{float}

\journal{Journal of \LaTeX\ Templates}

\begin{document}

\begin{frontmatter}

\title{Coarse-to-Fine Salient Object Detection \\with Low-Rank Matrix Recovery}


\author[mymainaddress]{Qi Zheng}
\ead{qiz@hust.edu.cn}

\author[mysecondaryaddress]{Shujian Yu}
\ead{yusjlcy9011@ufl.edu}

\author[mymainaddress]{Xinge You\corref{mycorrespondingauthor}}
\cortext[mycorrespondingauthor]{Corresponding author}
\ead{youxg@hust.edu.cn}



\address[mymainaddress]{Huazhong University of Science and Technology, Wuhan, China}
\address[mysecondaryaddress]{University of Florida, Gainesville, USA}

\begin{abstract}
Low-Rank Matrix Recovery (LRMR) has recently been applied to saliency detection by decomposing image features into a low-rank component associated with background and a sparse component associated with visual salient regions. Despite its great potential, existing LRMR-based saliency detection methods seldom consider the inter-relationship among elements within these two components, thus are prone to generating scattered or incomplete saliency maps. In this paper, we introduce a novel and efficient LRMR-based saliency detection model under a coarse-to-fine framework to circumvent this limitation. First, we roughly measure the saliency of image regions with a baseline LRMR model that integrates a $\ell_1$-norm sparsity constraint and a Laplacian regularization smooth term. Given samples from the coarse saliency map, we then learn a projection that maps image features to refined saliency values, to significantly sharpen the object boundaries and to preserve the object entirety. We evaluate our framework against existing LRMR-based methods on three benchmark datasets. Experimental results validate the superiority of our method as well as the effectiveness of our suggested coarse-to-fine framework, especially for images containing multiple objects.
\end{abstract}


\end{frontmatter}


\section{Introduction}
\label{sec:intro}

Visual saliency has been a fundamental problem in neuroscience, psychology, and computer vision for a long time~\cite{borji2013state,borji2015salient}. It refers to the identification of a portion of essential visual information contained in the original image. Recently, studies of visual saliency have been extended from originally predicting eye-fixation to identifying a region containing salient objects, known as \emph{salient object detection} or \emph{saliency detection}~\cite{wang2017salient}. Tremendous efforts have been made to saliency detection over the past decades owing to its extensive real applications in the realm of computer vision and pattern recognition~\cite{li2019moving,meng2016image}. For example, object detection and recognition become much more efficient and reliable by exploring only those salient locations and ignoring large irrelevant background.

Existing approaches for saliency detection can be divided into two categories: the bottom-up (or stimulus-driven) approaches and the top-down (or task-driven) approaches~\cite{borji2013state}. The bottom-up approaches detect saliency regions only using low-level visual information such as color, texture and localization, without requiring any specific knowledge on the objects and/or background. By contrast, the top-down approaches, including recently proposed deep-learning based methods (e.g.,~\cite{jetley2016end,zhang2017learning,wang2018detect}), utilize high-level human perceptual knowledge such as object labels or semantic information to guide the estimation of saliency maps. Compared with the top-down methods, bottom-up ones require less computational power and exhibit better generality and scalability~\cite{borji2013state,borji2015salient}.

A recent trend is to combine bottom-up cues with top-down priors to facilitate saliency detection using low-rank matrix recovery (LRMR) theory~\cite{candes2011robust}. Generally speaking, these methods (e.g.,~\cite{yan2010visual,lang2012saliency,zou2013segmentation,peng2017salient,zheng2017hierarchical}) assume that a natural scene image consists of visually consistent background regions (corresponding to a highly redundant information component with low-rank structure) and distinctive foreground regions (corresponding to a visually salient component with sparse structure). In~\cite{yan2010visual}, Yan~\emph{et al.} proposed a LRMR-based model using sparse representation of image features as input, where the sparse representation is obtained by learning a dictionary upon image patches. In~\cite{lang2012saliency}, Lang~\emph{et al.} introduced a multitask-sparsity pursuit for saliency detection, where a single low-rank matrix decomposition is replaced by seeking consistently sparse elements from the joint decompositions of multiple-feature matrices into pairs of low-rank and sparse matrices. Despite promising results achieved by various LRMR-based methods, there still remain two challenging problems~\cite{peng2017salient}: 1) Inter-correlations among elements within the sparse component are neglected, causing incompleteness or scattering of detected object; 2) Low-rank matrix recovery model is hard to separate salient objects from background when the background is cluttered or has similar appearance with the salient objects. Therefore, tree-structured sparsity constraint and Laplacian regularization are introduced in~\cite{peng2017salient} to address these two issues respectively.

In this paper, we first argue that the main reason for these two problems is that the spatial relationship among image regions (or super-pixels) is not fully taken into consideration in the original LRMR model. Moreover, the structured-sparse constraint in~\cite{peng2017salient}, actually, cannot effectively preserve such a relationship. To this end, we propose a novel LRMR-based saliency detection method under a coarse-to-fine framework to address the key issue while maintaining high efficiency. Our framework features two modules in a successive manner: a coarse-processing module, in which a Laplacian smooth term is integrated into a $\ell_1$-norm constrained LRMR (baseline) model to roughly detect salient regions; and a refinement module, in which a projection is learned upon the coarse saliency map to enhance object boundaries.

To summarize, our main contributions are threefold:
\begin{itemize}
    \item An effective saliency detection model, integrating $\ell_1$-norm sparsity constrained LRMR and Laplacian regularization, is proposed to roughly detect salient regions. We set this as our baseline model and demonstrate that it performs well in diverse scenes.
    \item A learning-based refinement module is developed to assign more accurate saliency values to such obscure regions, i.e., regions located around object boundaries, thus promoting the entirety of detected salient objects.
    \item Extensive experiments are conducted on three benchmark datasets to demonstrate the superiority of our method against other LRMR-based methods and the efficacy of the proposed coarse-to-fine framework.
\end{itemize}

The remainder of this paper is organized as follows. Section~\ref{sec:re_w} briefly reviews related work. In Section~\ref{sec:prob}, we present our coarse-to-fine framework for salient object detection in details. Section~\ref{sec:exp} shows the experimental results and analysis. Finally, Section~\ref{sec:conclusion} draws the conclusion.

\section{Related Work} \label{sec:re_w}
An extensive review on saliency detection is beyond the scope of this paper. We refer interested readers to two recently published surveys~\cite{borji2013state,borji2015salient} for more details about existing bottom-up and top-down approaches for saliency detection. This section first briefly reviews the prevailing unsupervised bottom-up saliency detection methods, and then introduces several popular LRMR-based methods that are closely related to our work.

\subsection{Popular Bottom-up Saliency Detection Methods}
As a pioneering work, Itti \emph{et al.}~\cite{itti1998model} innovatively suggested using ``Center and Surround" filters to extract image features and to simulate human vision system on multi-scale levels to generate saliency maps. Motivated by Itti's framework, various contrast-based approaches have been developed in past decades, which include local-contrast-based ones (e.g.,~\cite{goferman2012context,jiang2011automatic}), global-contrasts-based ones (e.g., \cite{cheng2015global,perazzi2012saliency,margolin2013makes}), or even those combining both local and global contrasts~(e.g.,~\cite{borji2012exploiting,lu2012saliency,lu2014robust}). Local contrast is estimated by measuring the difference between a ``center" pixel or small region with its neighbors, thus it is sensitive to high frequency changes such as edges and noises. On the contrary, global contrast is much more robust to local textures and edges, but they can fail to distinguish salient objects from the background that shares high similarity with the objects~\cite{borji2015salient,kim2016salient,li2016double}.

On the other hand, frequency domain also provides a reliable avenue for salient object detection. For example, Hou and Zhang~\cite{hou2007saliency} analyzed spectral residual of an image in spectral domain, where the high-frequency components are considered as background. Similar work was presented by Fang \emph{et al.}~\cite{fang2012bottom}, where the standard Fast Fourier Transform (FFT) is substituted with Quaternion Fourier Transform (QFT). Other representative examples include~\cite{li2013visual,imamoglu2013saliency}.

Graph theory-based methods~(e.g.,\cite{yang2013saliency,wang2016grab,zhu2014saliency,jiang2018saliency}) have attracted increasing attention in recent years due to their superior robustness and adaptability. For instance, Yang \emph{et al.}~\cite{yang2013saliency} adopted manifold ranking to rank the similarity of super-pixels with foreground and background seeds. Based on this model, Wang \emph{et al.}~\cite{wang2016grab} suggested detecting saliency by combining local graph structure and background priors together. This way, salient information among different nodes can be jointly exploited. However, a fully-connected graph suffers from high computational cost.

\subsection{LRMR-based Saliency Detection Methods} \label{sec:overview}
The usage of LRMR theory on saliency detection was initiated by Yan \emph{et al.}~\cite{yan2010visual} and then extended in~\cite{shen2012unified}. Generally, the LRMR-based methods assume that an image consists of an information-redundant part and a visually salient part, which are characterized with a low-rank component and a sparse component respectively. Specifically, a given image is firstly divided into small regions or super-pixels $\{B_i\}_{i=1,...,N}$ to reduce computational complexity, where $N$ is the number of regions. Features are extracted for each region, forming a feature matrix $\mathbf{F}=[\mathbf{f}_1,\mathbf{f}_2,\dots,\mathbf{f}_N]$. The LRMR theory is deployed to decompose $\mathbf{F}$ as follows:
\begin{align} \label{eq:lr_decom}
    (&\mathbf{L},\mathbf{S}) = arg\min_{\mathbf{L},\mathbf{S}} \Vert \mathbf{L} \Vert_* + \alpha\Vert \mathbf{S} \Vert_1 \nonumber \\
    &\textrm{s.t.}~~~~\mathbf{F}=\mathbf{L}+\mathbf{S}
\end{align}
where $\Vert \cdot \Vert_{*}$ denotes nuclear norm for the low-rank component and $\Vert \cdot \Vert_1$ denotes $\ell_1$-norm that is used to encourage sparseness. $\alpha>0$ is a trade-off parameter balancing the low-rank term and the sparse term. After the decomposition, a saliency map can be generated from the obtained sparse matrix $\mathbf{S}$:
\begin{equation} \label{eq:sal}
    s_j = \Vert \mathbf{s}_j\Vert_1
\end{equation}
where $\mathbf{s}_j$ denotes the $j$th column of matrix $\mathbf{S}$. Note that $\mathbf{s}_j$ is a vector herein, thus its $\ell_1$-norm is the sum of the absolute value of each entry.

Early LRMR-based methods are data-dependent, i.e., the learned dictionaries or transformations depend heavily on selected training images or image patches, which suffer from limited adaptability and generalization capability. To this end, various approaches are developed in an unsupervised manner by either adopting a multitask scheme~(e.g.,~\cite{lang2012saliency}) or introducing extra priors~(e.g.,~\cite{zou2013segmentation,zhang2017salient}). For example, Lang \emph{et al.}~\cite{lang2012saliency} proposed to jointly decompose multiple-feature matrices instead of directly combining individual saliency maps generated by decomposing each feature matrix. Zou \emph{et al.}~\cite{zou2013segmentation} introduced segmentation priors to cooperate with sparse saliency in an advanced manner. To preserve the entirety of detection objects, saliency fusion models~(e.g.,\cite{li2016double,li2014visual,li2017saliency,huang2015saliency}) were proposed thereafter. For instance, double low-rank matrix recovery (DLRMR) was suggested in~\cite{li2016double} to fuse saliency maps detected by multiple approaches.

Although above extensions improved algorithm robustness to cluttered backgrounds, there still remain two open problems. First, extra priors~\cite{zou2013segmentation} or complicated operations (such as saliency fusion~\cite{li2016double,huang2015saliency}) may incur expensive computational cost. Second, these methods neglect the spatial relationship among image regions, which cannot ensure the entirety of detected objects. The first work that attempts to address above two limitations is the recently proposed structured matrix decomposition (SMD) by Peng \emph{at al.}~\cite{peng2017salient}. Specifically, SMD introduces two new regularization terms to Eq.~(\ref{eq:lr_decom}): a tree-structured sparse constraint that is used to preserve inter-correlations among sparse elements and a Laplacian regularization term that is adopted to enlarge the difference between foreground and background. This way, the spatial relationship among sparse elements and the coherence between low-rank component and sparse component are explicitly modeled and optimized in a unified model. The objective of SMD is formulated as:

\begin{align} \label{eq:smd}
    (\mathbf{L},\mathbf{S})=arg\min_{\mathbf{L},\mathbf{S}}\Vert \mathbf{L}\Vert_*&+\alpha\sum_{i=1}^d \sum_{j=1}^{n_i} \Vert \mathbf{S}_{G_j^i} \Vert_p + \gamma \textrm{tr}(\mathbf{SGS}^T) \nonumber \\
    \textrm{s.t.} ~~~~\mathbf{F\odot P}&=\mathbf{L}+\mathbf{S}
\end{align}
where the matrix $\mathbf{P}$ represents high-level priors~\cite{shen2012unified}, including location prior, semantic prior and color prior. $\odot$ denotes dot-product of matrices. The term $\sum_{i=1}^d \sum_{j=1}^{n_i} \Vert \mathbf{S}_{G_j^i} \Vert_p$ denotes the structured-sparse constraint, $\Vert \cdot \Vert_p$ is the $\ell_p$-norm ($1\leq p\leq\infty$), $d$ is the depth (or layer) of index-tree and $n_i$ is the number of nodes at the $i$-th layer. Here $G_j^i$ denotes the $j$-th node at the $i$-th level of the index-tree such that $G_j^i\cap G_k^i=\emptyset$ ($\forall~2\le i\le d$ and $1\le j< k\le n_i$), $G_j^i\subseteq G_{j0}^{i-1}$, and $\cup_j G_j^i=G_{j0}^{i-1}$, where $j0$ is the indexing at the $(i{-}1)$-th level. $\mathbf{S}_{G_j^i}\in \mathbb{R}^{D\times \vert G_j^i\vert}$ ($\vert \cdot \vert$ denotes set cardinality) is the sub-matrix of $\mathbf{S}$ corresponding to node $G_j^i$. The third term $\gamma \textrm{tr}(\mathbf{SGS}^T)$ is introduced to promote the performance under cluttered background, where $\gamma>0$ is a parameter that balances this regularization and the other two terms. $\mathbf{G}$ is un-normalized graph Laplacian matrix.

Our work is directly motivated by SMD~\cite{peng2017salient}. However, two observations prompt us to propose our method:
\begin{itemize}
    \item SMD uses Laplacian constraint to reduce the coherence between low-rank component and sparse component under cluttered background. In fact, the Laplacian constraint is not novel in saliency detection literature. In our perspective, it performs more like a smooth term (just like it does in previous saliency detection literature) that can hardly increase the discrepancy between foreground and background.
    \item The structured-sparse constraint in SMD cannot effectively preserve spatial relationship among image regions. In fact, it may even disrupt such relationship if we apply this constraint on deep layers (as recommended by the authors).
\end{itemize}

The effects or functionality of Laplacian constraint can trace back to early work on saliency detection (e.g.,~\cite{borji2013state,borji2015salient,zhu2014saliency,lu2014learning}), which uses it as a smooth regularization term to reduce the discrepancy of saliency values from regions that have similar appearance or feature representations. Therefore, in the scenario of cluttered background (i.e., the salient object may be interfered by the background), the Laplacian constraint can hardly increase the discrepancy between foreground and background.

Regarding the second argument, spatial relationship among super-pixels is taken into consideration in the construction of tree nodes $G_j^i$. However, such relationship has not been preserved if we naively impose the $\ell_p$-norm sparse constraint on these nodes. It should be pointed out that in the deepest level of the tree, one node is composed of a single super-pixel, whereas in the shallowest level, one node is composed of all the super-pixels. According to scale theory, there exists an optimal scale for an object~\cite{lindeberg1998feature}. However, in tree-structured sparsity constraint, nodes in different levels contribute equally to final sparsity, which does not emphasize or highlight spatial relationship among image regions. Moreover, one should note that the $\ell_\infty$-norm and the $\ell_1$-norm in a specific node lead to row-sparsity and column-sparsity respectively, which contributes little to maintain the spatial structure.

\begin{figure*}[!t]
    \centering
    \includegraphics[width=\textwidth]{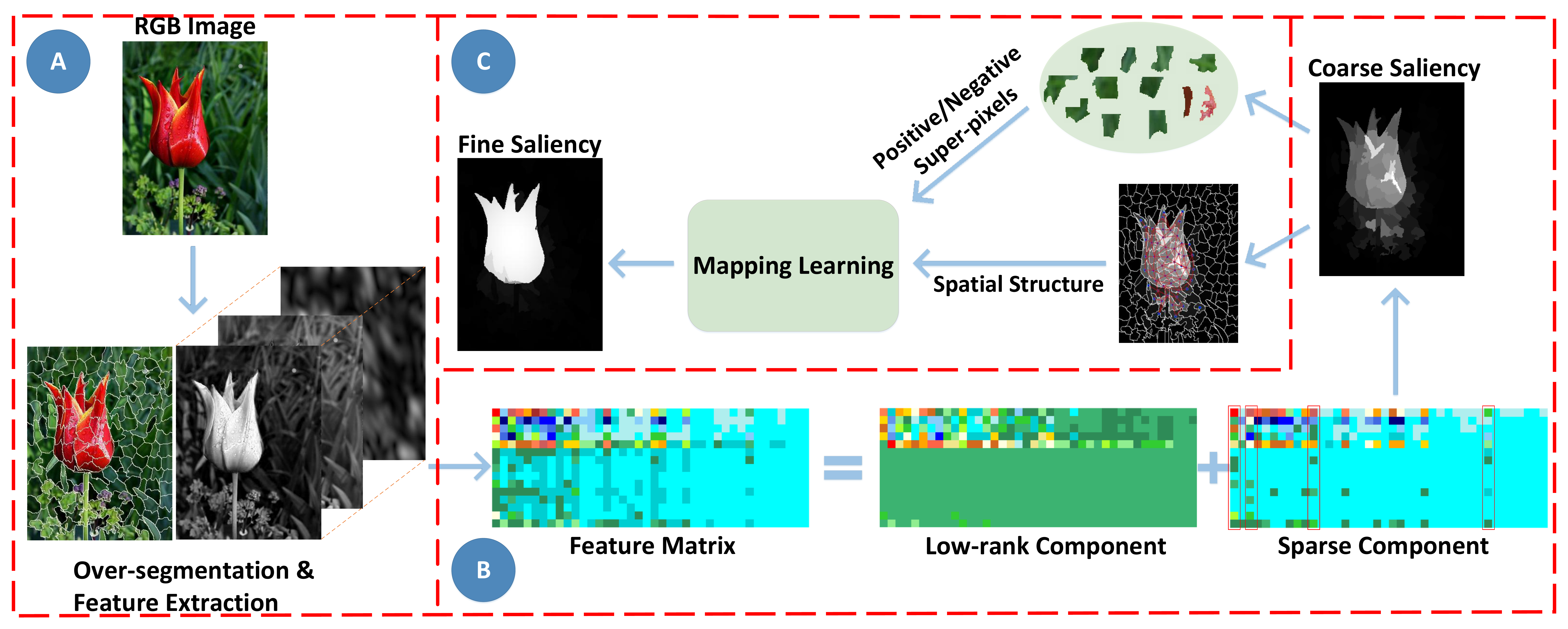}
    \caption{The general coarse-to-fine framework of our proposed LRMR-based saliency detection method. Given an input image, we first conduct over-segmentation and feature extraction (module (A)), and then generate coarse saliency map via applying low-rank matrix decomposition to the feature matrix (module (B)). We finally learn a projection, using super-pixels in the coarse saliency map, to map raw features to their refined saliency values (module (C)).}
    \label{fig:framework}
\end{figure*}

\section{Our Method} \label{sec:prob}
This paper proposed a novel LRMR-based saliency detection method under a coarse-to-fine framework that can effectively preserve object entirety, even in the scenarios of multiple objects or cluttered background. To this end, we integrate the basic LRMR model in Eq.~(\ref{eq:lr_decom}) and Laplacian regularization to generate a coarse saliency map. Then, we learn a projection on top of super-pixels sampled from the coarse saliency map to obtain final saliency. By exploiting the spatial relationship among super-pixels in the refinement module, the proposed method is robust to cluttered background. The overall flowchart of our method is illustrated in Fig.~\ref{fig:framework}.

\subsection{The Limitation of Tree-Structured Sparsity in SMD}
In Section~\ref{sec:overview}, we pointed out that tree-structured regularization in SMD is not suitable for salient object detection. In this section, we further propose two arguments to specify the limitations of tree-structured regularization: (1) for images containing only a single object, the regularization imposed on shallow layers of the index-tree is sufficient to render satisfactory performance, and (2) for images containing multiple objects or complex scenes, the regularization imposed on deeper layers will destroy the spatial structure of a group of objects, thus disrupting the entirety of detected saliency regions.

\begin{figure*}[!ht]
    \centering
    \includegraphics[width=\textwidth]{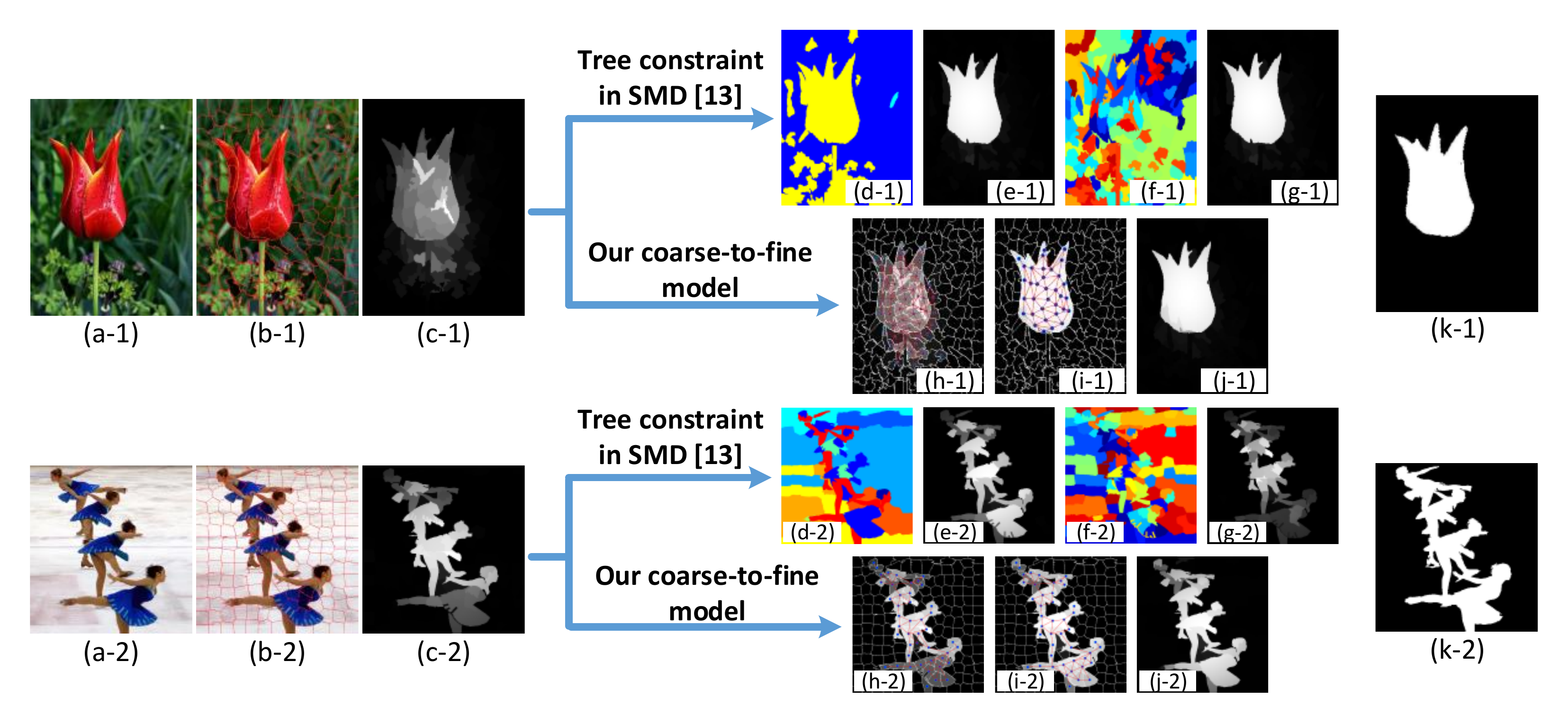}
    \caption{Comparison of a four-layer-based index-tree structured constraint in SMD~\cite{peng2017salient} and our coarse-to-fine architecture. In both two examples, (a) shows the raw image; (b) shows the over-segmented super-pixels; (c) shows the coarse saliency map obtained by our baseline model (i.e., Eq. (\ref{eq:lr_lg})); (d) shows the merged graph in the $2$-nd layer of the index-tree; (e) shows the saliency map obtained by incorporating tree-constraint in both the $1$-st layer and the $2$-nd layer; (f) shows the merged graph in the $4$-th layer of the index-tree; (g) shows saliency map obtained by incorporating tree-constraints from the $1$-st layer to the $4$-th layer; (h) shows the coarse graph constructed with salient super-pixels from the rough saliency map in (c); (i) shows the refined salient graph; (j) shows the refined saliency map given by refined salient graph in (i); (k) shows the ground truth. The tree-structured constraint in shallow layers can effectively preserve the spatial relationship and the entirety of detected object. However, this functionality disappears with respect to deeper layers in the scenario of multiple objects (or complex backgrounds as shown in the supplementary material). By contrast, our coarse-to-fine model enhances object entirety in a designated way, with much clearer boundaries, regardless of the number and the size of objects in the image. More examples are available in our supplementary material.}
    \label{fig:c_to_f}
\end{figure*}

To experimentally validate the effects of structured-sparse regularization in Eq.~(\ref{eq:smd}) and our coarse-to-fine architecture, we give two examples in Fig.~\ref{fig:c_to_f}\footnote{More examples are shown in supplementary material}. Specifically, we construct a four-layer index-tree for validation. It is worth noting that the bottom layer (the 4-th layer) of index-tree is composed of $N$ graphs, each containing a super-pixel, whereas the top layer (the 1-st layer) of index-tree only contains one graph that incorporating all $N$ super-pixels. The $\ell_p$-norm constraint is applied to each graph separately and then the results are summed.

The first image is presented to illustrate the case of single object in pure background. Comparing Fig.~\ref{fig:c_to_f}(c-1) with Fig.~\ref{fig:c_to_f}(e-1) and Fig.~\ref{fig:c_to_f}(g-1) respectively, we can observe that adding constraint to the 2-nd layer eliminates irrelevant background, while deeper constraint is unnecessary for preserving spatial structure of the flower. Meanwhile, comparing Fig.~\ref{fig:c_to_f}(c-1) with Fig.~\ref{fig:c_to_f}(i-1), we can see that our coarse-to-fine architecture is also able to remove irrelevant background, e.g., regions below the flower.

The second image is presented to illustrate the case of multiple objects. Comparing Fig.~\ref{fig:c_to_f}(c-2) with Fig.~\ref{fig:c_to_f}(e-2) and Fig.~\ref{fig:c_to_f}(g-2), we can observe that adding constraint to the 2-nd layer promotes the structural entirety of objects to some extent, while deeper constraint destroys the spatial structure of the bodies. On the contrary, comparing Fig.~\ref{fig:c_to_f}(c-2) with Fig.~\ref{fig:c_to_f}(i-2), we can see that our coarse-to-fine architecture produces more accurate saliency of super-pixels around object boundaries, e.g., super-pixels in leg areas adjacent to image boundary, thus improves the entirety of salient objects.

\subsection{Coarse Saliency from Low-Rank Matrix Recovery} \label{sec:coarse_s}
Due to the limitations of tree-structured sparsity, we revert to the original $\ell_1$-norm sparsity constraint, yielding sparsity by treating each element individually. Specifically, we roughly measure saliency of image regions using
\begin{align} \label{eq:lr_lg}
    (\mathbf{L},\mathbf{S}) = arg\min_{\mathbf{L},\mathbf{S}} &\Vert \mathbf{L} \Vert_* + \alpha\Vert \mathbf{S} \Vert_1 + \gamma \textrm{tr}(\mathbf{SGS}^T) \nonumber \\
    \textrm{s.t.}~~~~\mathbf{F\odot P}&=\mathbf{L}+\mathbf{S}
\end{align}
where matrices $\mathbf{F},\mathbf{P},\mathbf{L},\mathbf{S} \in \mathbb{R}^{D\times N}$, $\mathbf{G}$ is un-normalized graph Laplacian matrix. Once the low-rank matrix $\mathbf{L}$ and sparse matrix $\mathbf{S}$ are determined, saliency value $s_j$ of the $j$th super-pixel can be calculated by Eq.~(\ref{eq:sal}). \\

\noindent\textbf{Optimization:}
The optimization problem in Eq.~(\ref{eq:lr_lg}) can be efficiently solved via the alternating direction method of multipliers (ADMMs)~{\cite{lin2011linearized}}. For simplification, we denote the projected feature matrix $\mathbf{F\odot P}$ as $\mathbf{F}$. An auxiliary variable $\mathbf{Z}$ is introduced and problem Eq.~(\ref{eq:lr_lg}) becomes
\begin{align} \label{eq:lr_lg2}
    &(\mathbf{L},\mathbf{S}) = arg\min_{\mathbf{L},\mathbf{S}} \Vert \mathbf{L} \Vert_* + \alpha\Vert \mathbf{S} \Vert_1 + \gamma \textrm{tr}(\mathbf{ZGZ}^T) \nonumber \\
    &\textrm{s.t.}~~~~\mathbf{F}=\mathbf{L}+\mathbf{S}, ~~~\mathbf{S}=\mathbf{Z}
\end{align}
Lagrange multipliers $\mathbf{Y}_1$ and $\mathbf{Y}_2$ are introduced to remove the equality constraints, and the augmented Lagrangian function is constructed as
\begin{align} \label{eq:lr_l3}
    \mathcal{L}(\mathbf{L},\mathbf{S},\mathbf{Z},\mathbf{Y}_1,\mathbf{Y}_2,\mu)&=\Vert \mathbf{L}\Vert_*+\alpha\Vert \mathbf{S}\Vert_1+\gamma \textrm{tr}(\mathbf{ZGZ}^T) \nonumber\\
    &+\textrm{tr}(\mathbf{Y}_1^T(\mathbf{F}-\mathbf{L}-\mathbf{S}))+\textrm{tr}(\mathbf{Y}_2^T(\mathbf{S}-\mathbf{Z})) \nonumber\\
    &+\frac{\mu}{2}(\Vert \mathbf{F}-\mathbf{L}-\mathbf{S}\Vert_F^2+\Vert \mathbf{S}-\mathbf{Z}\Vert_F^2)
\end{align}
where $\mu>0$ is the penalty parameter.

Iterative steps of minimizing the Lagrangian function are utilized to optimize Eq.~(\ref{eq:lr_l3}), and stop criteria at step $k$ are given by Eq.~(\ref{eq:stop1}) and Eq.~(\ref{eq:stop2})
\begin{align}
    \Vert \mathbf{F}-\mathbf{L}_k-\mathbf{S}_k\Vert_F/\Vert \mathbf{F}\Vert_F &< \varepsilon_1 \label{eq:stop1}\\
    \max(\Vert \mathbf{S}_k-\mathbf{S}_{k-1}\Vert_F, \Vert \mathbf{L}_k-\mathbf{L}_{k-1}\Vert_F) &< \varepsilon_2 \label{eq:stop2}
\end{align}

The variables $\mathbf{L},~\mathbf{S},~\mathbf{Z},~\mathbf{Y}_1,~\mathbf{Y}_2$ and $\mu$ can be alternately updated by minimizing the augmented Lagrangian function $\mathcal{L}$ with other variables fixed. In this model, each variable can be updated with a closed form solution. With respect to $\mathbf{L}$ and $\mathbf{S}$, they can be updated as follows
\begin{align} \label{eq:up_l}
\mathbf{L}^{k+1}&=arg\min_\mathbf{L} \frac{1}{\mu^k}\Vert \mathbf{L}\Vert_*{+}\frac{1}{2}\Vert \mathbf{L}{-}(\mathbf{F}{-}\mathbf{S}^k{+}\frac{\mathbf{Y}_1^k}{\mu})\Vert_F^2 \nonumber \\
&=\mathbf{U}\Gamma_{\frac{1}{\mu^k}}[\Sigma]\mathbf{V}^T
\end{align}
\begin{align} \label{eq:up_s}
\mathbf{S}^{k+1}&=arg\min_\mathbf{S} \mathcal{L}(\mathbf{L}^{k+1},\mathbf{S},\mathbf{Z}^k,\mathbf{Y}_1^k,\mathbf{Y}_2^k,\mu^k) \nonumber \\
&=arg\min_\mathbf{S} \alpha\Vert \mathbf{S}\Vert_1+\textrm{tr}((\mathbf{Y}_1^k)^T(\mathbf{F}-\mathbf{L}^k-\mathbf{S}^k)) \nonumber \\
&+\textrm{tr}((\mathbf{Y}_2^k)^T(\mathbf{S}{-}\mathbf{Z}^k)){+}\frac{\mu^k}{2}(\Vert \mathbf{F}{-}\mathbf{L}^{k{+}1}{-}\mathbf{S}^k\Vert_F^2{+}\Vert \mathbf{S}{-}\mathbf{Z}^k\Vert_F^2) \nonumber \\
&=arg\min_\mathbf{S} \frac{\alpha}{2\mu^k}\Vert \mathbf{S}\Vert_1{+}\frac{1}{2}\Vert \mathbf{S}{-}(\mathbf{F}{-}\mathbf{L}^{k+1}+\mathbf{Z}^k{-}\frac{\mathbf{Y}_2^k{-}\mathbf{Y}_1^k}{\mu^k})\Vert_F^2 \nonumber \\
&=\Gamma_{\frac{\alpha}{2\mu}}[\mathbf{F}{-}\mathbf{L}^{k+1}+\mathbf{Z}^k{-}\frac{\mathbf{Y}_2^k{-}\mathbf{Y}_1^k}{\mu^k}]
\end{align}
where the soft-thresholding operator $\Gamma_\epsilon$ is defined by
\begin{displaymath}
\Gamma_\epsilon[x]=\left\{ \begin{array}{ll}
x-\epsilon, & \textrm{if $x>\epsilon$} \\
x+\epsilon, & \textrm{if $x<-\epsilon$} \\
0, & \textrm{otherwise}
\end{array}\right.
\end{displaymath}
and $[\mathbf{U},~\Sigma,~\mathbf{V}^T]=\textrm{SVD}(\mathbf{F}-\mathbf{S}+\frac{\mathbf{Y}_1}{\mu})$, where SVD is the \emph{singular value decomposition}.

Regarding $\mathbf{Z},~\mathbf{Y}_1,~\mathbf{Y}_2$ and $\mu$, we can update them as follows
\begin{align}
\mathbf{Z}^{k+1}&=(\mu^k\mathbf{S}^{k+1}+\mathbf{Y}_2^k)(\mu^kI+2\gamma \mathbf{G})^{-1} \label{eq:up_z} \\
\mathbf{Y}_1^{k+1}&=\mathbf{Y}_1^k+\mu^k(\mathbf{F}-\mathbf{L}^{k+1}-\mathbf{S}^{k+1}) \label{eq:up_y1} \\
\mathbf{Y}_2^{k+1}&=\mathbf{Y}_2^k+\mu^k(\mathbf{S}^{k+1}-\mathbf{Z}^{k+1}) \label{eq:up_y2} \\
\mu^{k+1}&=\min(\rho \mu^k,~\mu_{max})
\end{align}
where the parameter $\rho>0$ controls the convergence speed.

\subsection{Learning-based Saliency Refinement} \label{sec:fine_s}
As we have discussed in Section~\ref{sec:overview}, the coarse saliency map generated by LRMR-based approaches ignores spatial relationship among adjacent super-pixels. To further improve the detection results, we refine the coarse saliency by learning a projection from image features to saliency values.

\begin{figure} [!h]
    \centering
    \begin{tabular}{cccc}
    \includegraphics[width=0.2\textwidth]{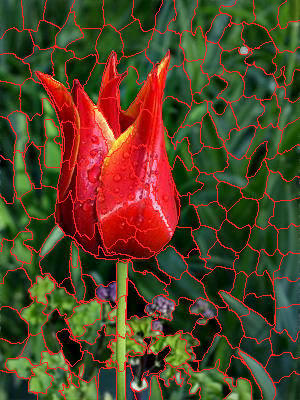} & \includegraphics[width=0.2\textwidth]{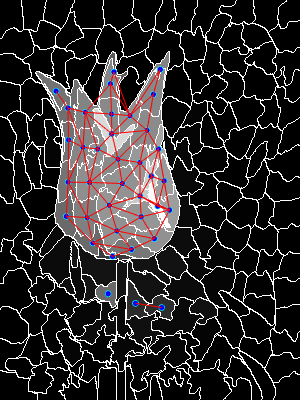} & \includegraphics[width=0.2\textwidth]{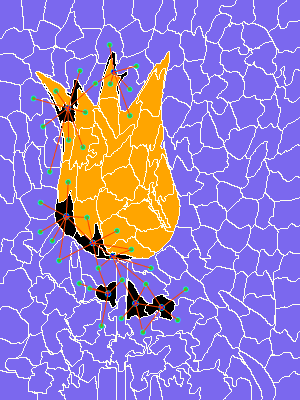} & \includegraphics[width=0.2\textwidth]{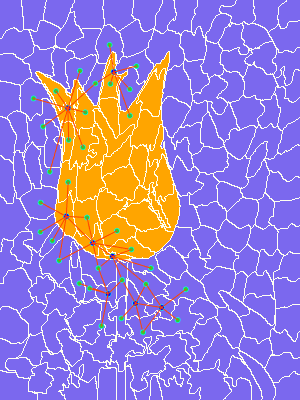} \\
    \includegraphics[width=0.2\textwidth]{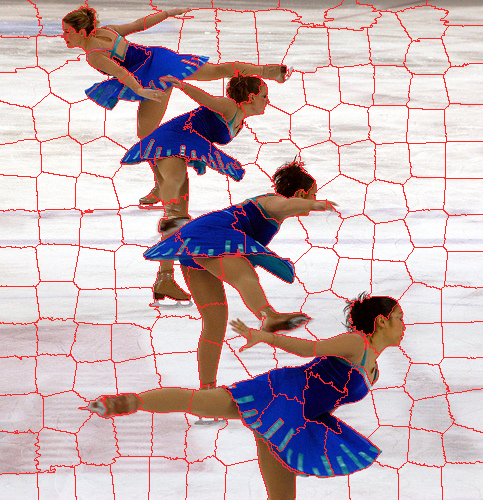} & \includegraphics[width=0.2\textwidth]{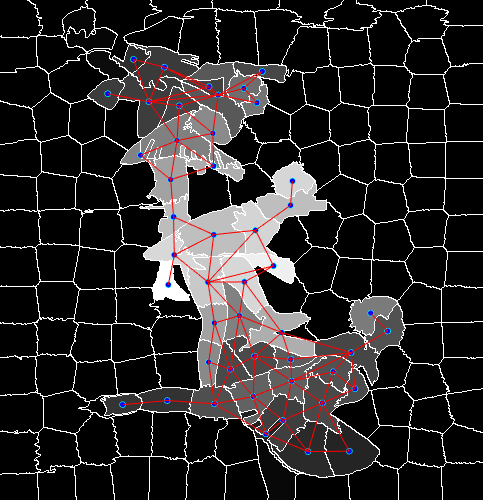} & \includegraphics[width=0.2\textwidth]{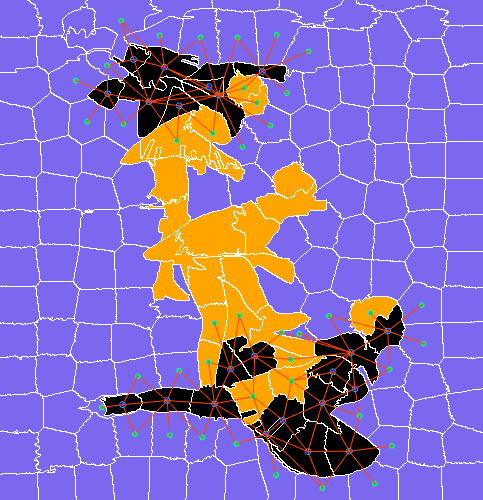} & \includegraphics[width=0.2\textwidth]{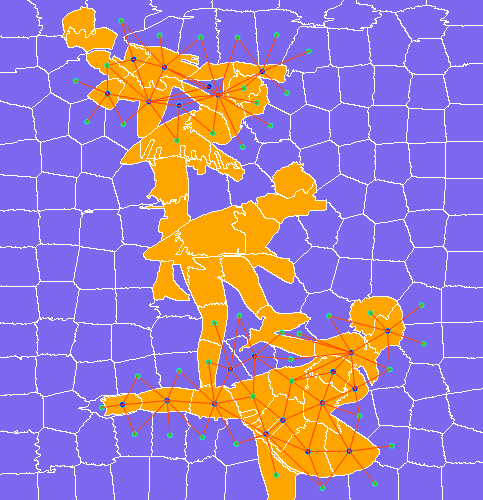} \\
    (a) & (b) & (c) & (d)
    \end{tabular}
    \caption{Illustration for the process of learning-based saliency refinement. (a) Over-segmented RGB images ($N=200$). (b) Coarse saliency maps and corresponding graph structure of salient super-pixels. (c) Positive samples (in orange), negative samples (in purple) and tough samples (in black) generated from coarse saliency map. The line-connections demonstrate spatial relationship around those tough samples. (d) Refined saliency of those tough samples and their spatial relationship.}
    \label{fig:eg_samp}
\end{figure}

Given the coarse saliency $s_i,~i \in \{1,...,N\}$ calculated using Eq.~(\ref{eq:sal}), we can roughly distinguish salient regions from background. In order to obtain common interior feature of foreground and background respectively, we choose confident super-pixels based on their coarse saliency value. Specifically, we set two thresholds $\tau_1,~\tau_2~(\tau_1<\tau_2)$ to select confident super-pixel samples for background and for foreground respectively, i.e., super-pixels with saliency value lower than $\tau_1$ are considered as negative samples, and super-pixels with saliency value higher than $\tau_2$ are considered as positive ones. We denote $\mathbf{A}\in \mathbb{R}^{N_1\times D}$ as the sample matrix composed of both positive and negative samples, and $\mathbf{Y}=[\mathbf{y}_1,\mathbf{y}_2,\dots,\mathbf{y}_{N_1}]\in \mathbb{R}^{N_1\times 2}$ as corresponding label matrix, where $N_1$ is the total number of confident samples. For the $i$th positive sample, its label vector is $\mathbf{y}_i=[1,0]$, while for the $j$th negative sample, its label vector is $\mathbf{y}_j=[0,1]$. See Fig.~\ref{fig:eg_samp} for more intuitive examples.

In order to determine the saliency of those tough samples $\mathbf{A}^t$, we utilize their spatial relationship with these confident samples, as shown in Fig.~\ref{fig:eg_samp}. Based on the coarse saliency and adjacency, rough saliency for the $j$th tough sample $\mathbf{A}^t_j$ is generated by
\begin{align}
    s_j = \frac{\sum_{k=1}^Ks_k|P_k|}{\sum_{k=1}^K|P_k|}
\end{align}
where $K$ is the number of super-pixels adjacent to the $j$th tough sample $\mathbf{A}^t_j$, and $|P_k|$ denotes the number of pixels contained in the $k$th super-pixel. Similarly, we formulate label vector of $\mathbf{A}^t_j$ as $\mathbf{y}^t_j=[s_j,1-s_j]$, and the label matrix $\mathbf{Y}^t=[\mathbf{y}^t_1,\mathbf{y}^t_2,\dots,\mathbf{y}^t_{N_2}]\in \mathbb{R}^{N_2\times 2}$, where $N_2$ is the number of tough samples.

\setlength{\tabcolsep}{1pt}

Combining the coarse saliency for confident samples and tough samples, we build our saliency refining model as follows
\begin{align} \label{eq:mapping}
    \mathbf{M}=\arg\min_\mathbf{M} \{\frac{1}{2}\Vert \mathbf{M} \Vert_F^2+\frac{\lambda_1}{2}\Vert \mathbf{AM}-\mathbf{Y} \Vert_F^2  +\frac{\lambda_2}{2}\Vert \mathbf{A}^t\mathbf{M}-\mathbf{Y}^t\Vert_F^2\}
\end{align}
where $\mathbf{A}^t \in \mathbb{R}^{N_2\times D}$ and $\mathbf{Y}^t \in \mathbb{R}^{N_2\times 2}$ represent tough samples and corresponding labels, respectively. $\mathbf{M}\in \mathbb{R}^{D\times 2}$ is the projection to be learned, and $\lambda_1,~\lambda_2$ are regularization parameters. The first term imposes regularization on $\mathbf{M}$ to avoid over-fitting, whereas the second and third terms require respectively labeled confident and tough samples. Once the projection is learned, saliency of those tough super-pixels is given by the first column of matrix $\mathbf{A}^t\mathbf{M}$.

Despite the simplicity of Eq.~(\ref{eq:mapping}), one should note that background region is typically much larger than salient region. This leads to the issue of learning in the circumstance of imbalanced data. In order to overcome this limitation, we introduce sample-wise weights to balance the contributions of positive and negative samples in projection learning, which is formulated as follows
\begin{align}\label{eq:w_mapping}
    \mathbf{M}=\arg\min_\mathbf{M} \{\frac{1}{2}\Vert \mathbf{M}\Vert_F^2+\frac{\lambda_1}{2}\sum_{i=1}^{N_1}w_{i}\Vert \mathbf{A}_i \mathbf{M}-\mathbf{Y}_i\Vert_F^2 +\frac{\lambda_2}{2}\Vert \mathbf{A}^t\mathbf{M}-\mathbf{Y}^t\Vert_F^2\}
\end{align}
where $w_i$ is the weight for the $i$th confident sample. Now the second term distinguishes the importance of positive samples from that of negative ones. In fact, we can simplify Eq.~(\ref{eq:w_mapping}) by combining the second term and the third term with generalized weights $\tilde{w}_i$ as follows
\begin{align} \label{eq:unified}
    \mathbf{M}=\arg\min_\mathbf{M}{\frac{1}{2}\Vert \mathbf{M}\Vert_F^2+\frac{\lambda}{2}\sum_{i=1}^{N_1+N_2}\tilde{w}_i\Vert \widetilde{\mathbf{A}}_i \mathbf{M}-\widetilde{\mathbf{Y}}_i\Vert_F^2}
\end{align}
where $\tilde{w}_i$ is the weight for the $i$th sample, either positive one, negative one or tough one. Given that there are much more negative samples than the positive ones, we adopt the weighting strategy that is widely used in imbalanced date problems~\cite{sun2009strategies} to leverage the effect of positive and negative samples, i.e, $w_i/w_j=N_n/N_p$, where $w_i$ and $w_j$ are the weights of the $i$th positive sample and the $j$th negative sample, respectively. $N_n$ and $N_p$ denote the number of negative and positive samples. Moreover, noting that labels of positive/negative samples are more reliable than that of tough ones, the weight of a tough sample is set to be half of that for a confident sample. To summarize, the weighting scheme is given by
\begin{displaymath}
\tilde{w}_i = \left\{ \begin{array}{ll}
    0.5 & \textrm{if $\widetilde{\mathbf{A}}_i$ is a tough sample} \\
    1.0 & \textrm{if $\widetilde{\mathbf{A}}_i$ is a negative sample} \\
    N_n/N_p & \textrm{if $\widetilde{\mathbf{A}}_i$ is a positive sample}
\end{array} \right.
\end{displaymath}
where $N_n+N_p=N_1$. Optimization problem in Eq.~(\ref{eq:unified}) can be efficiently solved by
\begin{align} \label{eq:mapping_sol}
    \mathbf{M} = (\mathbf{I} {+} \lambda \widetilde{\mathbf{A}}^T\widetilde{\mathbf{W}}\widetilde{\mathbf{A}})^{-1}(\lambda \widetilde{\mathbf{A}}^T\widetilde{\mathbf{W}}\widetilde{\mathbf{Y}})
\end{align}
where $\widetilde{\mathbf{W}}$ is a diagonal matrix with $\widetilde{\mathbf{W}}_{ii}=\tilde{w}_i$, and $\mathbf{I}$ is an identity matrix.

\subsection{Complexity analysis and discussion}
Here we briefly discuss the computational complexity of optimization in Section~\ref{sec:coarse_s} and Section~\ref{sec:fine_s} respectively, and we have $\mathbf{F},~\mathbf{L},~\mathbf{S},~\mathbf{Z},~\mathbf{Y}_1,~\mathbf{Y}_2\in \mathbb{R}^{D\times N}$, $\widetilde{\mathbf{A}}\in \mathbb{R}^{N\times D},~\widetilde{\mathbf{W}}\in \mathbb{R}^{N\times N},~\widetilde{\mathbf{Y}}\in\mathbb{R}^{N\times 2}$.

We set the $k$th iteration for coarse saliency generation as an example. The time consumption mainly involves three kinds of operations, i.e., SVD, matrix inversion and matrix multiplication. Specifically, update for $\mathbf{L}$ and $\mathbf{S}$ is addressed by SVD, with the complexity of $\mathcal{O}(2DN^2+D^2N)$ and $\mathcal{O}(DN)$, respectively. While major operations in updating $\mathbf{Z}$ include matrix inversion and matrix multiplication, with complexity of $\mathcal{O}(N^3+DN^2)$. Considering $N>D$, the final computational complexity is $\mathcal{O}(N^3)$. Compared with this, the optimization for the tree-structured sparsity in~\cite{peng2017salient} requires no extra computational complexity. However, multi-scale segmentation in constructing the index-tree introduces computational cost thus slows down the speed, as listed in Table~\ref{tab:time}.

For saliency refinement, the solution in Eq.~(\ref{eq:mapping_sol}) involves matrix inversion and matrix multiplication, with the complexity of $\mathcal{O}(2DN^2+D^2N+2DN+2D^2)$ and $\mathcal{O}(D^3)$, respectively. Considering $N>D$, the final computational complexity is $\mathcal{O}(DN^2)$.

Since human-annotated ground-truth labels are not required in either coarse module or refinement module, our proposed model may benefit existing deep-learning based saliency detection methods (e.g., \cite{quan2017unsupervised,zhang2019synthesizing}) as well. Specifically, the saliency map generated in the coarse module can serve as external priors and provide weak supervision during the initial training process of the deep neural networks. Moreover, in the successive fusion process, the refinement module provides an alternative to improve the entirety of the detected salient objects without referring to other images.

\section{Experiments} \label{sec:exp}
In this section, extensive experiments are conducted to demonstrate the effectiveness and superiority of our method. We first introduce the quantitative metrics and the implementation details of our method in Section~\ref{sec:exp_setup}. Then in Section~\ref{sec:com_contr}, we compare our method (including our baseline model) with other LRMR-based methods to emphasize the effectiveness and advantage of our coarse-to-fine architecture. In Section~\ref{sec:com_app}, we present a systematic comparison with state-or-the-arts to show the superiority of our method. Finally in Section~\ref{sec:param_sensi}, we analyze the effects of different parameters in our method. Three benchmark datasets are selected: MSRA10K~\cite{cheng2015global} contains 10,000 images with a single object per image, iCoSeg~\cite{batra2011interactively} contains 643 images with multiple objects per image, and ECSSD~\cite{perazzi2012saliency} contains 1,000 images with cluttered backgrounds. We also select $12$ state-of-the-art methods for comparison. Among them, three methods are LRMR-based, i.e., SMD~\cite{peng2017salient}, SLR~\cite{zou2013segmentation} and ULR~\cite{shen2012unified}. Moreover, we select five state-of-the-art methods that use contrast or incorporating priors, i.e., RBD~\cite{zhu2014saliency}, PCA~\cite{margolin2013makes}, HS~\cite{yan2013hierarchical}, HCT~\cite{kim2016salient} and DSR~\cite{li2013saliency}. The four remaining methods are MR~\cite{yang2013saliency}, SS~\cite{hou2012image}, FT~\cite{achanta2009frequency}, and DRFI~\cite{wang2017salient}. All the experiments in this paper were conducted with MATLAB2016b on an Intel i5-6500 3.2GHz Dual Core PC with 16GB RAM.

\subsection{Experimental Setup} \label{sec:exp_setup}
We follow the same experimental setup in SMD~\cite{peng2017salient} to compare the performance of different methods. The quantitative metrics include precision-recall (PR) curve, receiver operating characteristic (ROC) curve, area under the ROC curve (AUC), weighted $F_\beta^w$-measure (WF), overlapping ratio (OR) and mean absolute error (MAE). Supposing saliency values are normalized to the range of $[0,1]$, the generated saliency map can be binarized with a given threshold, i.e., salient or non-salient. PR curve is obtained by setting a series of discrete threshold ranging from $0$ to $1$ on the grayscale saliency map. ROC curve is obtained in a similar way, the only difference is that ROC measures hit-rate (recall) and false-alarm. WF is proposed in~\cite{margolin2014evaluate} to achieve a trade-off between precision and recall $F_\beta^w=((1+\beta^2)P^w\times R^w)/(\beta^2 P^w+R^w)$, with $\beta^2=0.3$ in previous work~\cite{wang2017salient,zhu2014saliency}. OR measures the intersection between predicted (binarized) saliency map (S) and the ground-truth saliency map (G), $\textrm{OR}=\frac{\vert S \cap G \vert}{\vert S \cup G \vert}$. MAE gives a numerical difference between the continuous saliency map and the true saliency map.

For our method, we adopt \emph{simple linear iterative clustering} (SLIC) algorithm~\cite{achanta2012slic} ($N=200$) for over-segmentation and extract the widely used $53$-dimensional features (i.e., color, responses of steerable pyramid filters, responses of Gabor filters) as conducted in previous approaches~\cite{zou2013segmentation,peng2017salient,shen2012unified}. Initialization for variables and parameters in the coarse module are set as $\mathbf{L}^0=\mathbf{S}^0=\mathbf{Z}^0=\mathbf{Y}_1^0=\mathbf{Y}_2^0=0,~\mu^0=0.1,~\mu_{max}=1e10,~\rho=1.1$. Regularization parameters for coarse saliency generation are set as optimal ones, i.e., $\alpha=0.35,~\gamma=1.1$ through out the experiments except for parametric analysis. For the refinement module, we set $\lambda=10$ and corresponding parametric sensitivity is provided in Section~\ref{sec:param_sensi}. As for homogenization, we consider location, contrast and background priors as done in~\cite{peng2017salient}.

\setlength{\tabcolsep}{1pt}
\setlength\extrarowheight{3pt}
\begin{table*}[!h]
    \centering
    \caption{Comparison with the other low-rank methods and performance boost with different baselines on three datasets. The best two results are marked with red and blue respectively. The sign $^+$ denotes method with refinement.}
    \begin{tabular}{c |c c c c|c c c c|c c c c}
        \hline
        \multirow{2}{*}{Dataset} & \multicolumn{4}{c|}{MSRA10K} & \multicolumn{4}{c|}{iCoSeg} & \multicolumn{4}{c}{ECSSD} \\
        \cline{2-13}
         & WF$\uparrow$ & OR$\uparrow$ & AUC$\uparrow$ & MAE$\downarrow$ & WF$\uparrow$ & OR$\uparrow$ & AUC$\uparrow$ & MAE$\downarrow$ & WF$\uparrow$ & OR$\uparrow$ & AUC$\uparrow$ & MAE$\downarrow$ \\ \hline
         ULR~\cite{shen2012unified} & 0.425 & 0.524 & 0.831 & 0.224 & 0.379 & 0.443 & 0.814 & 0.222 & 0.351 & 0.369 & 0.788 & 0.274\\
         SLR~\cite{zou2013segmentation} & 0.601 & 0.691 & 0.840 & 0.141 &0.473 & 0.505 & 0.805 & 0.179 & 0.402 & 0.486 & 0.805 & 0.226\\
         SMD~\cite{peng2017salient} & {\color{red}0.704} & {\color{red}0.741} & {\color{red}0.847} & {\color{red}0.104} & {\color{blue}0.611} & {\color{blue}0.598} & {\color{blue}0.822} & {\color{blue}0.138} & {\color{red}0.544} & {\color{red}0.563} & {\color{red}0.813} & {\color{red}0.174}\\
         Ours (C) & {\color{blue}0.688} & {\color{blue}0.734} & {\color{blue}0.844} & {\color{blue}0.108} & {\color{red}0.614} & {\color{red}0.599} & {\color{red}0.823} & {\color{red}0.137} & {\color{blue}0.535} & {\color{blue}0.557} & {\color{blue}0.810} & {\color{blue}0.175}\\ \hline

         ULR$^+$ & 0.532 & 0.597 & 0.846 & 0.195 & 0.439 & 0.459 & 0.814 & 0.219 & 0.421 & 0.418 & 0.801 & 0.262 \\
         SLR$^+$ & 0.681 & 0.726 & 0.847 & 0.122 & 0.602 & 0.587 & 0.816 & 0.161 & 0.519 & 0.542 & 0.814 & 0.199 \\
         SMD$^+$ & {\color{red}0.706} & {\color{red}0.753} & {\color{red}0.854} & {\color{red}0.103} & {\color{blue}0.630} & {\color{blue}0.618} & {\color{red}0.838} & {\color{blue}0.132} & {\color{red}0.546} & {\color{red}0.571} & {\color{red}0.820} & {\color{red}0.175} \\
         Ours & {\color{blue}0.705} & {\color{blue}0.751} & {\color{red}0.854} & {\color{blue}0.104} & {\color{red}0.634} & {\color{red}0.624} & {\color{red}0.838} & {\color{red}0.131} & {\color{blue}0.545} & {\color{red}0.571} & {\color{red}0.820} & {\color{blue}0.176}\\ \hline
    \end{tabular}
    \label{tab:baseline}
\end{table*}

\setlength{\tabcolsep}{3.0pt}
\begin{figure*}[!h]
    \centering
    \begin{tabular}{cccccccccc}
        \includegraphics[width=0.08\textwidth]{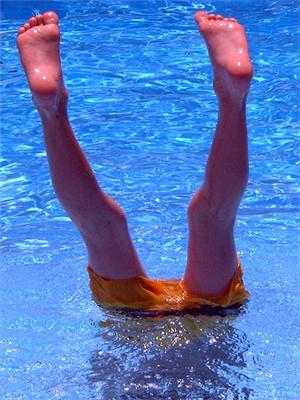} &
        \includegraphics[width=0.08\textwidth]{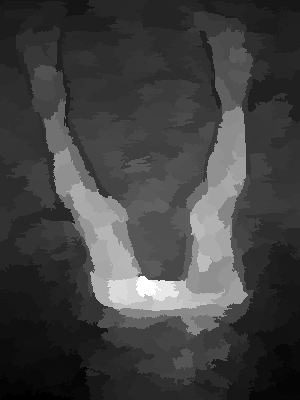} &
        \includegraphics[width=0.08\textwidth]{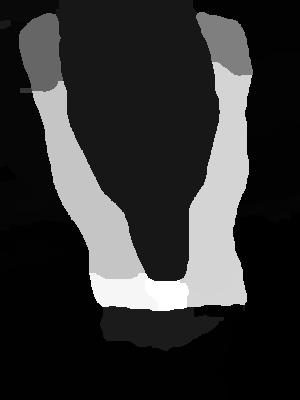} &
        \includegraphics[width=0.08\textwidth]{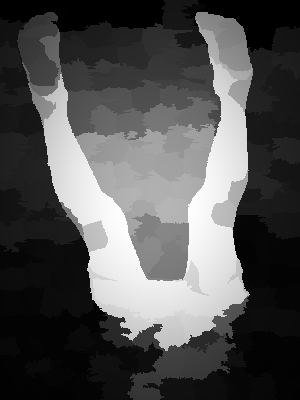} &
        \includegraphics[width=0.08\textwidth]{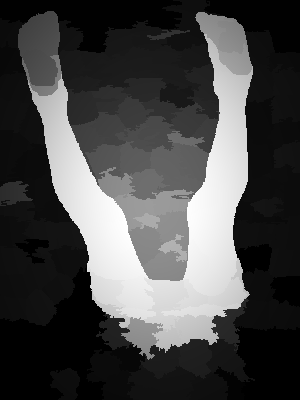} &
        \includegraphics[width=0.08\textwidth]{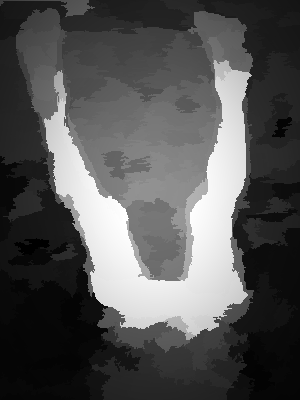} &
        \includegraphics[width=0.08\textwidth]{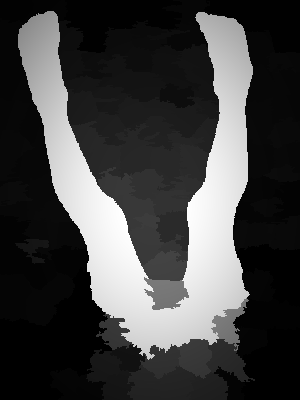} &
        \includegraphics[width=0.08\textwidth]{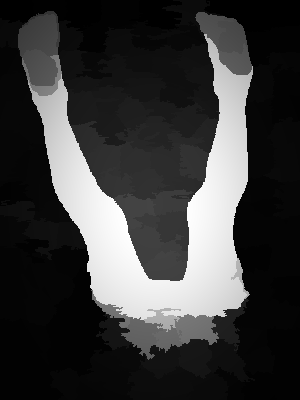} &
        \includegraphics[width=0.08\textwidth]{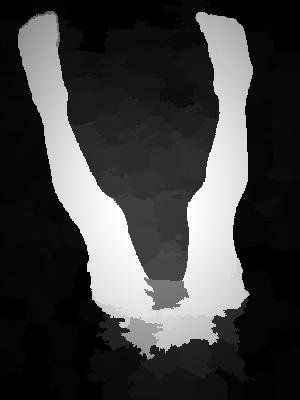} &
        \includegraphics[width=0.08\textwidth]{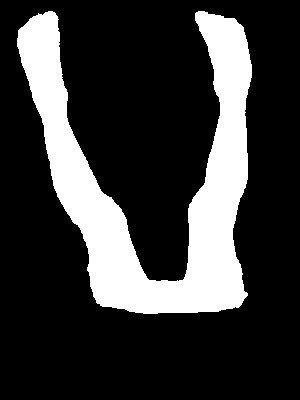} \\

        \includegraphics[width=0.08\textwidth]{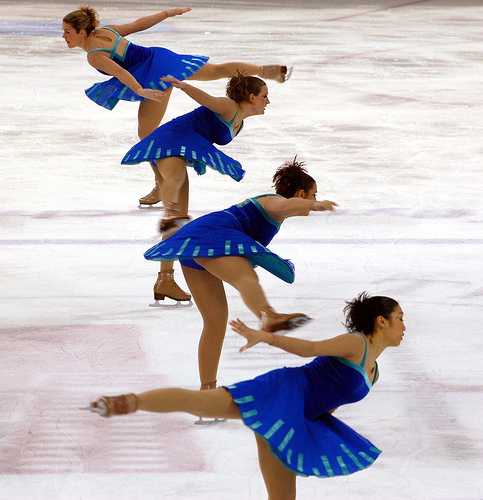} &
        \includegraphics[width=0.08\textwidth]{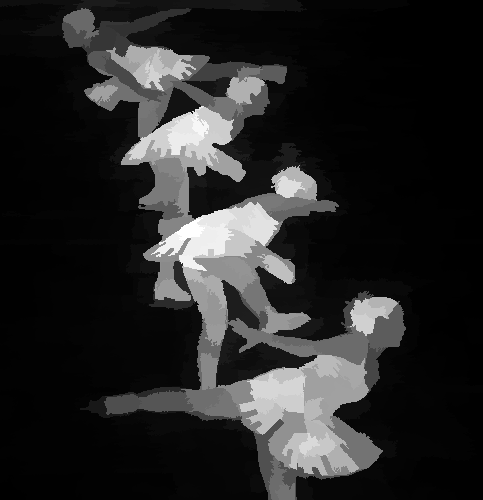} &
        \includegraphics[width=0.08\textwidth]{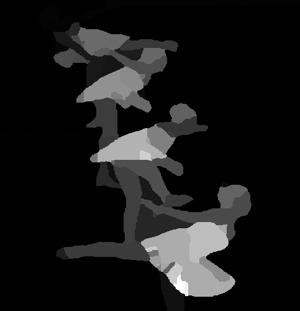} &
        \includegraphics[width=0.08\textwidth]{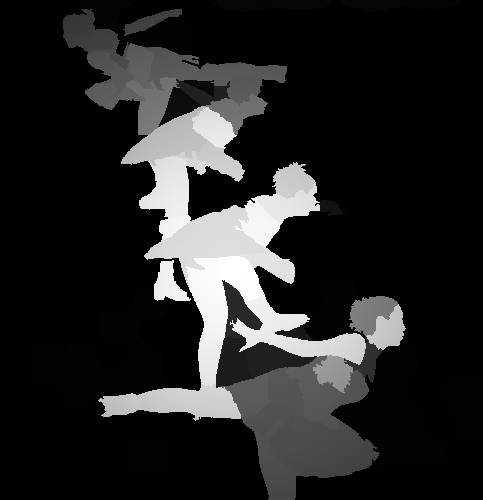} &
        \includegraphics[width=0.08\textwidth]{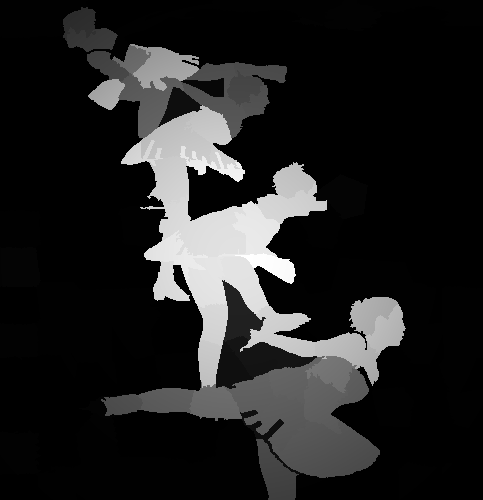} &
        \includegraphics[width=0.08\textwidth]{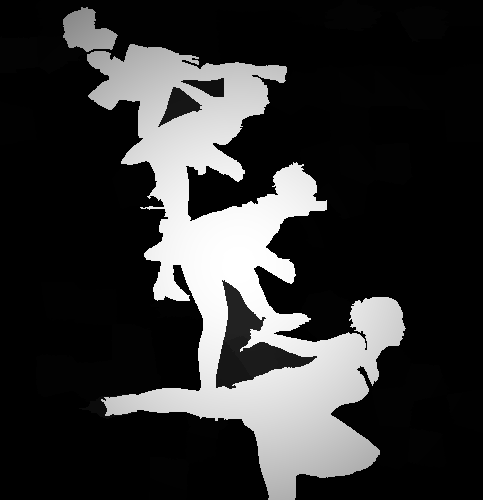} &
        \includegraphics[width=0.08\textwidth]{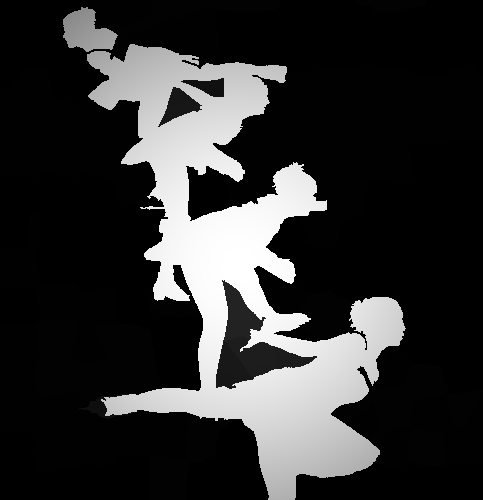} &
        \includegraphics[width=0.08\textwidth]{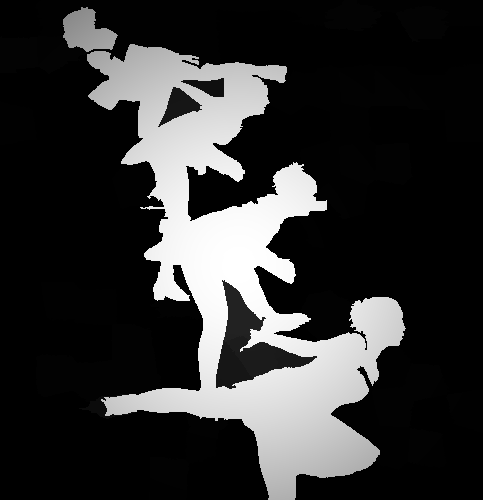} &
        \includegraphics[width=0.08\textwidth]{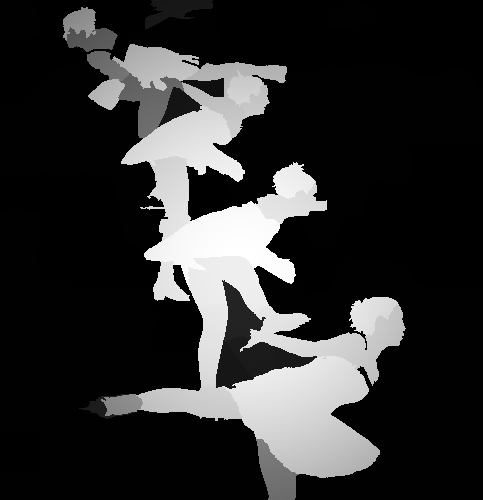} &
        \includegraphics[width=0.08\textwidth]{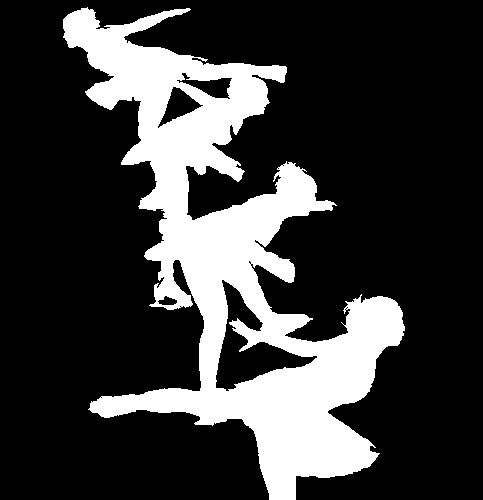} \\

        \includegraphics[width=0.08\textwidth]{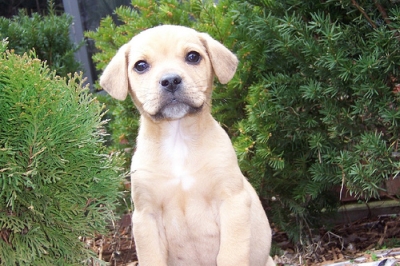} &
        \includegraphics[width=0.08\textwidth]{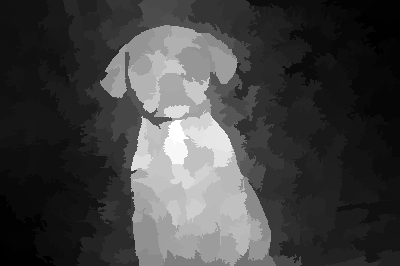} &
        \includegraphics[width=0.08\textwidth]{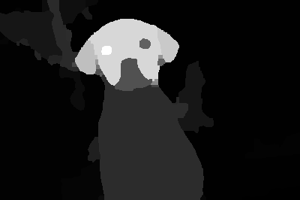} &
        \includegraphics[width=0.08\textwidth]{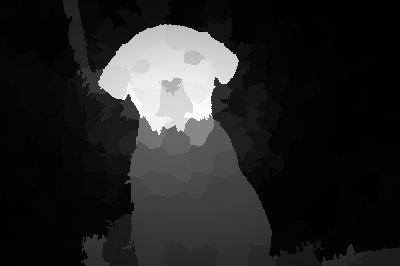} &
        \includegraphics[width=0.08\textwidth]{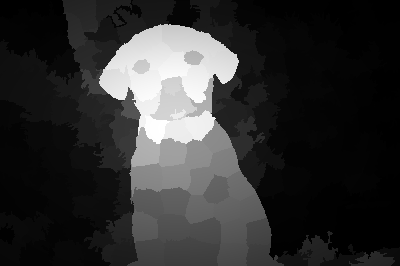} &
        \includegraphics[width=0.08\textwidth]{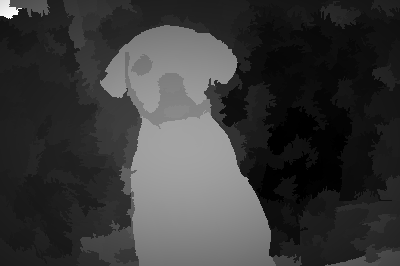} &
        \includegraphics[width=0.08\textwidth]{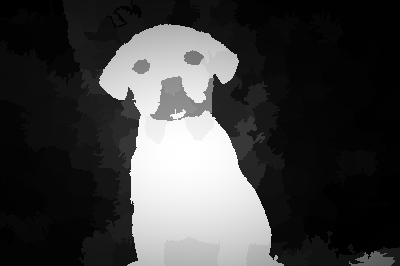} &
        \includegraphics[width=0.08\textwidth]{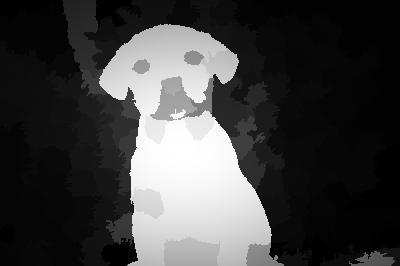} &
        \includegraphics[width=0.08\textwidth]{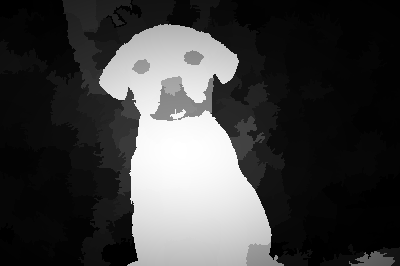} &
        \includegraphics[width=0.08\textwidth]{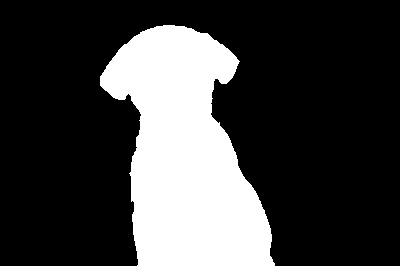} \\
        \small{RGB} & \small{ULR~\cite{shen2012unified}} & \small{SLR~\cite{zou2013segmentation}} & \small{SMD~\cite{peng2017salient}} & \small{Ours (C)} & \small{ULR$^+$} & \small{SLR$^+$} & \small{SMD$^+$} & \small{Ours} & \small{GT}
    \end{tabular}
    \caption{Visual comparison of our method (the coarse and the fine) with the other low-rank involved approaches. The sign $^+$ denotes method with refinement. The three images are randomly selected from MSRA10K, iCoSeg and ECSSD datasets, respectively.}
    \label{fig:lr_comp}
\end{figure*}

\subsection{Comparison with LRMR-based methods} \label{sec:com_contr}

\subsubsection{The effectiveness of our baseline model}
To evaluate the performance of our baseline model, i.e., the low-rank decomposition model with Laplacian constraint in Eq.~(\ref{eq:lr_lg}), a thorough comparison with other LRMR-based methods including ULR~\cite{shen2012unified}, SLR~\cite{zou2013segmentation} and SMD~\cite{peng2017salient} is provided in Table~\ref{tab:baseline} and Fig.~\ref{fig:lr_comp}. From the qualitative comparison in Fig.~\ref{fig:lr_comp}, we can see that methods such as ULR and SLR fail to generate uniform detection results. By contrast, salient objects detected by SMD~\cite{peng2017salient} and our baseline model are much smoother. This results further validate our argument that the Laplacian regularization plays more like a smooth term, rather than increasing the discriminancy around object boundaries as claimed in~\cite{peng2017salient}. From quantitative comparison in Table~\ref{tab:baseline}, we can see that our baseline model and SMD~\cite{peng2017salient} outperform ULR~\cite{shen2012unified} and SLR~\cite{zou2013segmentation} by a large margin. It is worth noting that our baseline model is only slightly outperformed by SMD~\cite{peng2017salient} on MSRA10K and ECSSD datasets. While on iCoSeg dataset, our baseline model achieves even better result than SMD~\cite{peng2017salient} in terms of all the four metrics. Two key conclusions can be drawn from the experimental results. First, the basic $\ell_1$-norm sparsity constraint performs almost equally to the structured-sparse regularization, which indicates that the latter can hardly preserve spatial relationship among elements within the sparse component. Second, tree-structured sparsity constraint is not suitable in the scenario of multiple objects.


\subsubsection{The advantage of our coarse-to-fine framework}
It can be observed in Fig.~\ref{fig:lr_comp} that salient objects detected by these LRMR-based approaches are not entire enough, and even contain irrelevant background regions. This is because the basic LRMR model ignores the spatial relationship of object parts. Though SMD~\cite{peng2017salient} attempts to handle this issue by replacing original $\ell_1$-norm sparsity constraint with structured-sparse constraint, it can hardly achieve the goal as aforementioned. Instead, we address the issue by cascading a learned projection to produce finer saliency maps. We can see that our method generates more entire saliency result compared with our baseline model, e.g., the persons in the second image and the dog in the third image. Besides, the refinement module also helps eliminate irrelevant background, e.g., blue water in the first image. With quantitative comparison listed in Table~\ref{tab:baseline}, we can see an obvious boost of performance of our model on all the three benchmark datasets, compared with that of our baseline model.

To further verify the general effectiveness of our coarse-to-fine architecture, we conduct more experiments with different LRMR baseline models, i.e., ULR~\cite{shen2012unified}, SLR~\cite{zou2013segmentation} and SMD~\cite{peng2017salient}. Test results are also summarized in Table~\ref{tab:baseline}. Comparing with original baselines, models with refinement show an improvement on all the three datasets. The best performance is achieved by our method and also by the SMD~\cite{peng2017salient} model with refinement. Similar visual improvement as discussed above can be observed in Fig.~\ref{fig:lr_comp}. It is especially obvious for the ULR~\cite{shen2012unified} baseline, where clearer and more entire saliency maps are generated after refinement.

\subsection{Comparison with State-of-the-Arts} \label{sec:com_app}
\begin{figure*}[!ht]
\centering
\begin{tabular}{ccc}
    \includegraphics[width=0.32\textwidth]{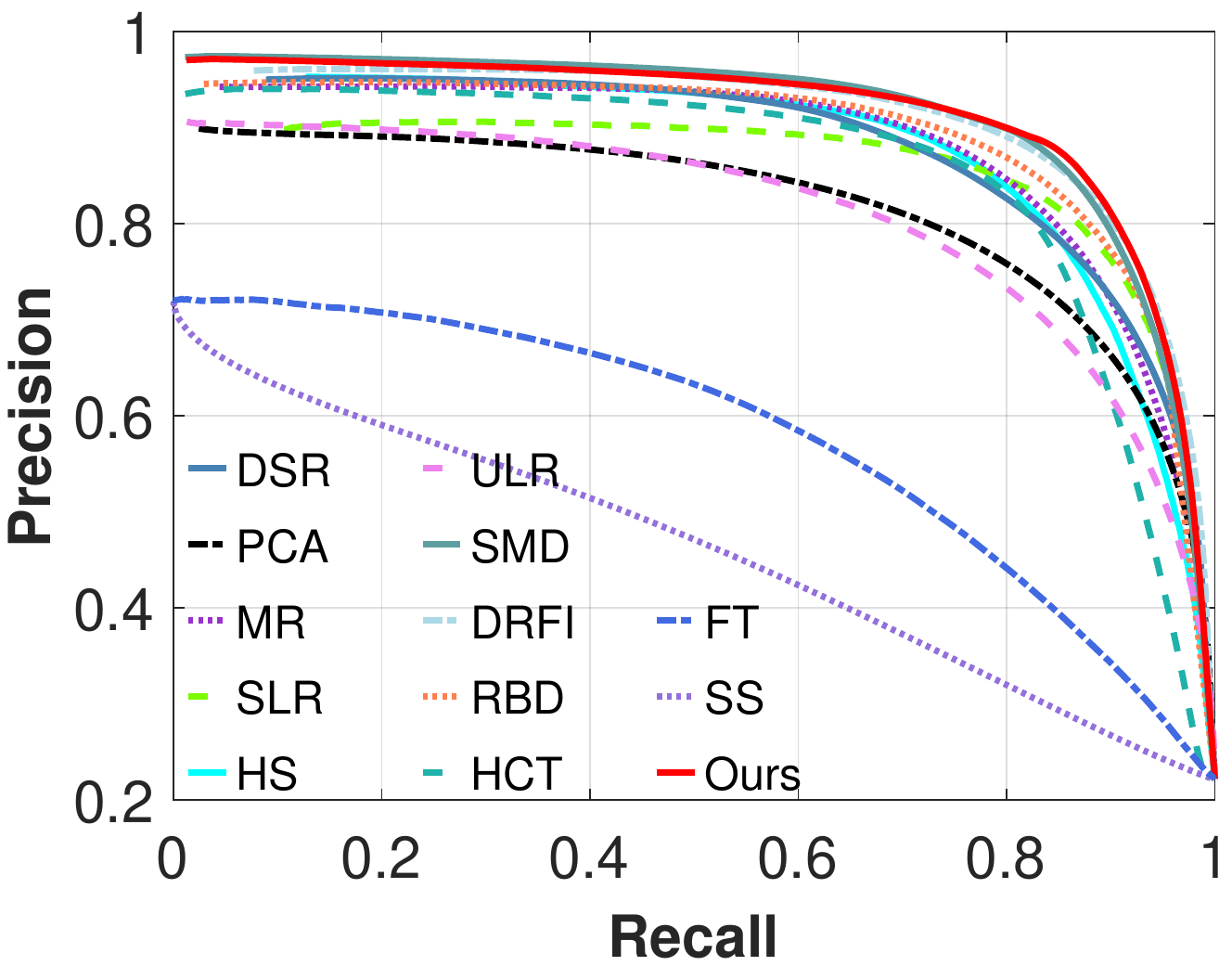} & \includegraphics[width=0.32\textwidth]{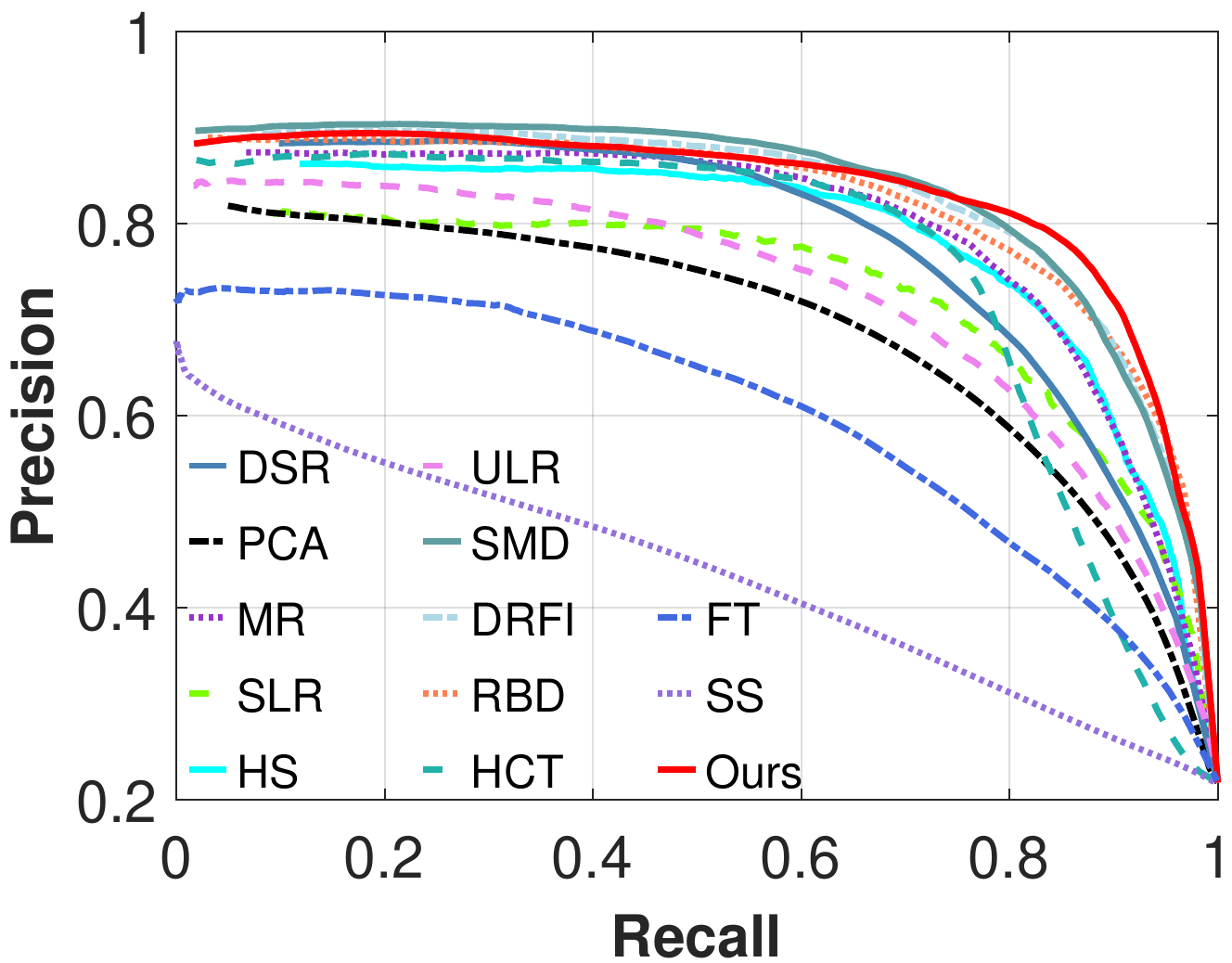} &
    \includegraphics[width=0.32\textwidth]{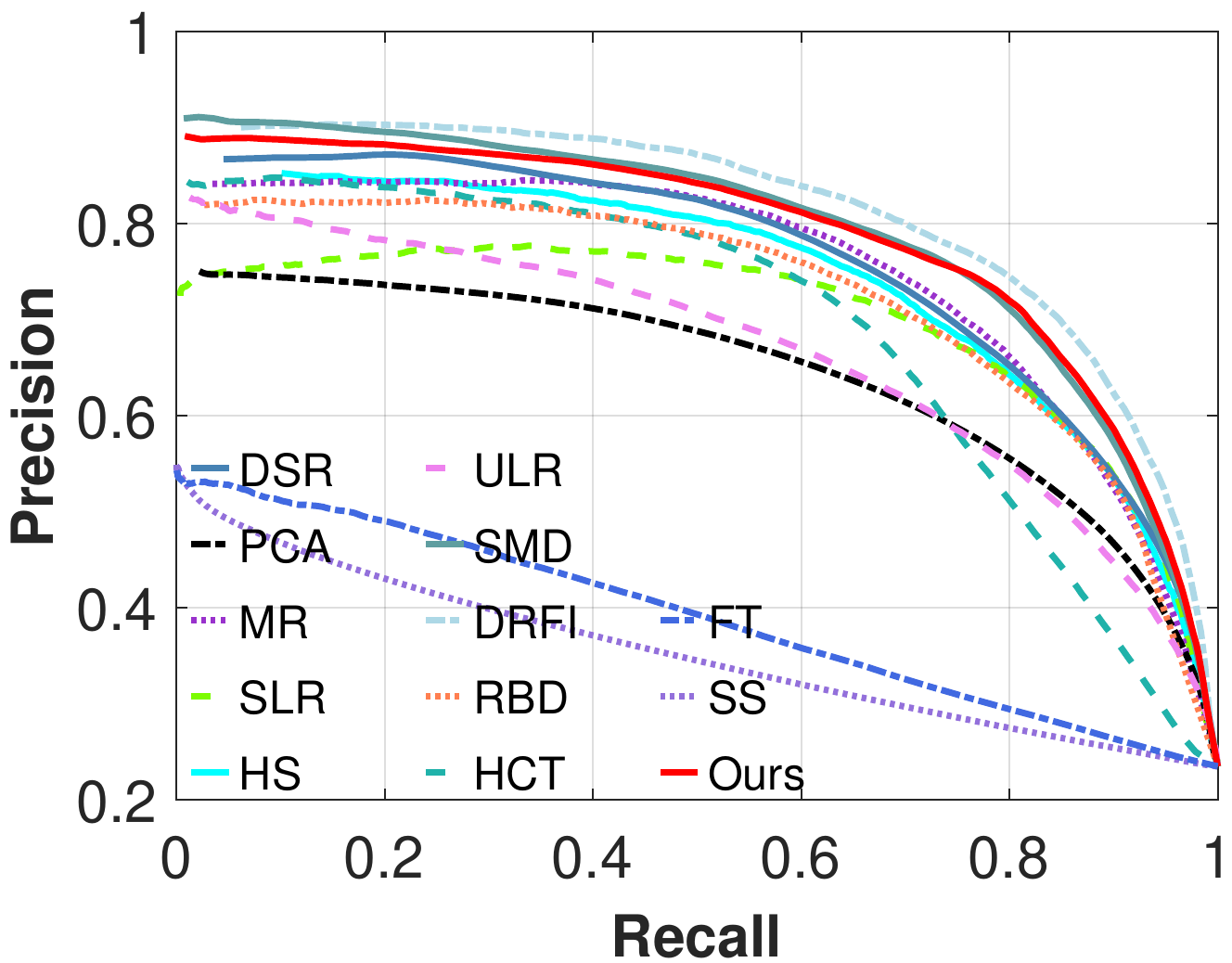} \\
    (a)&(b)&(c)
\end{tabular}
\caption{PR curve of all methods. (a) results on MSRA10K dataset. (b) results on iCoSeg dataset. (c) results on ECSSD dataset}
\label{fig:pr_curve}
\end{figure*}

\begin{figure*}[!ht]
\centering
\begin{tabular}{ccc}
    \includegraphics[width=0.32\textwidth]{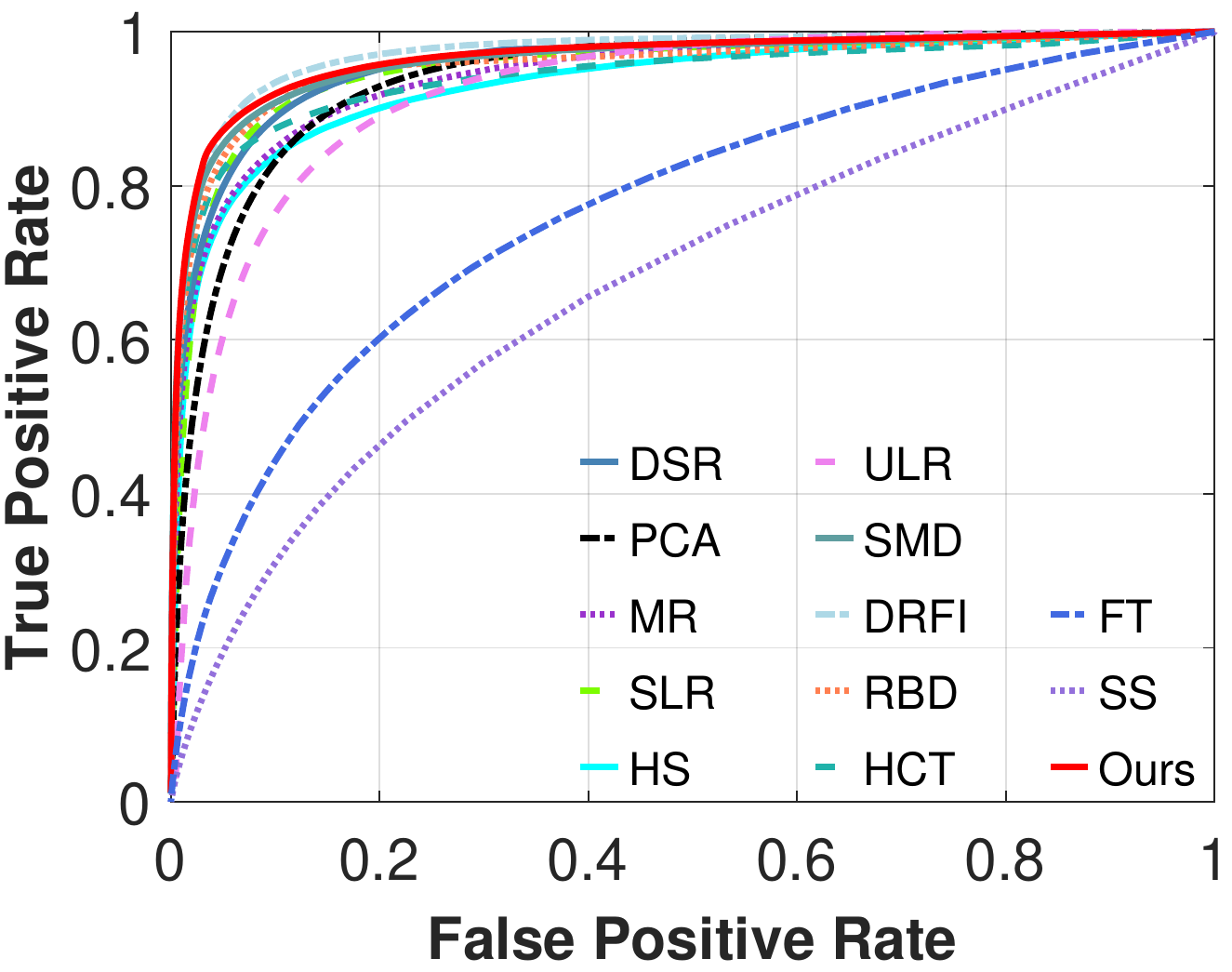} & \includegraphics[width=0.32\textwidth]{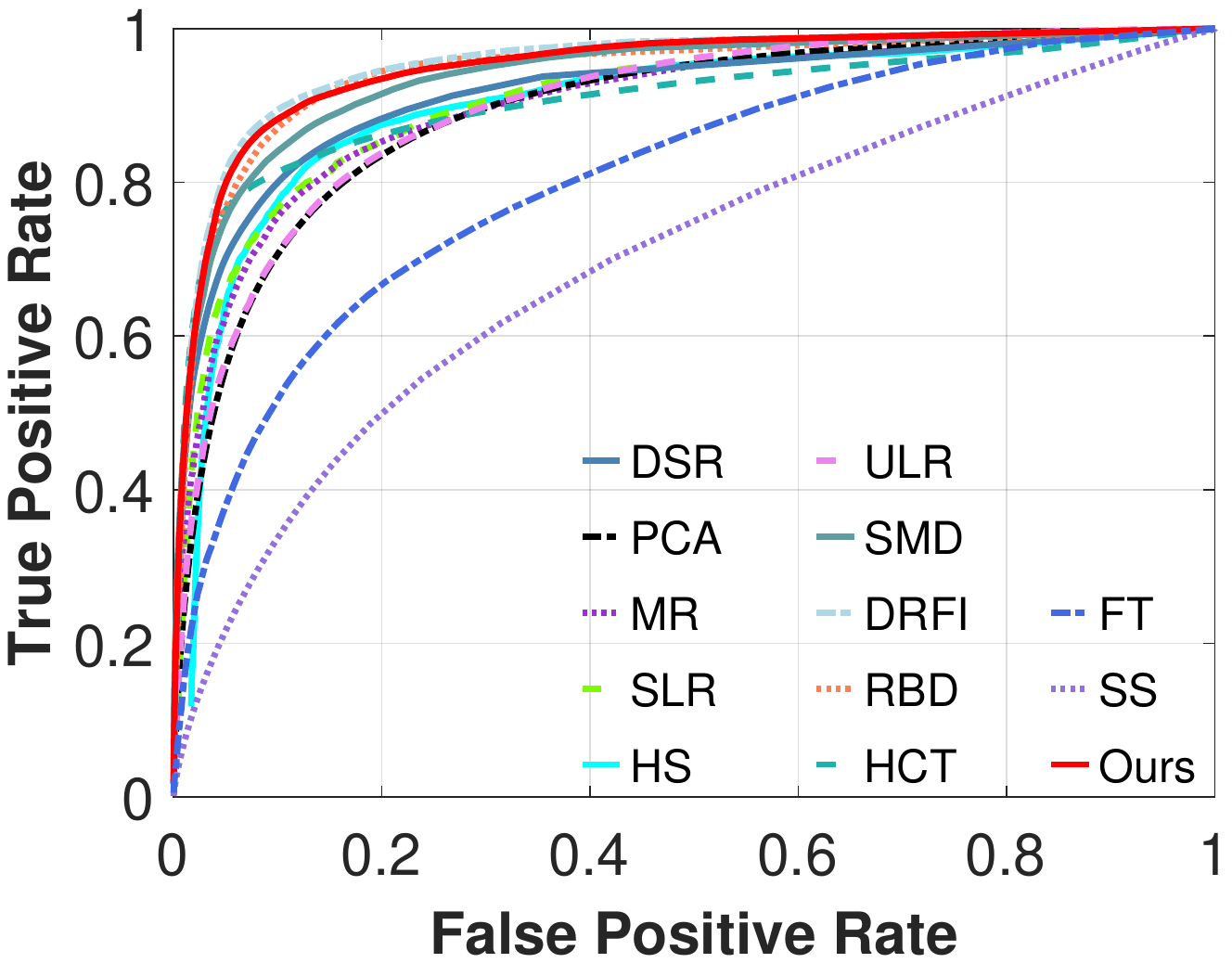} &
    \includegraphics[width=0.32\textwidth]{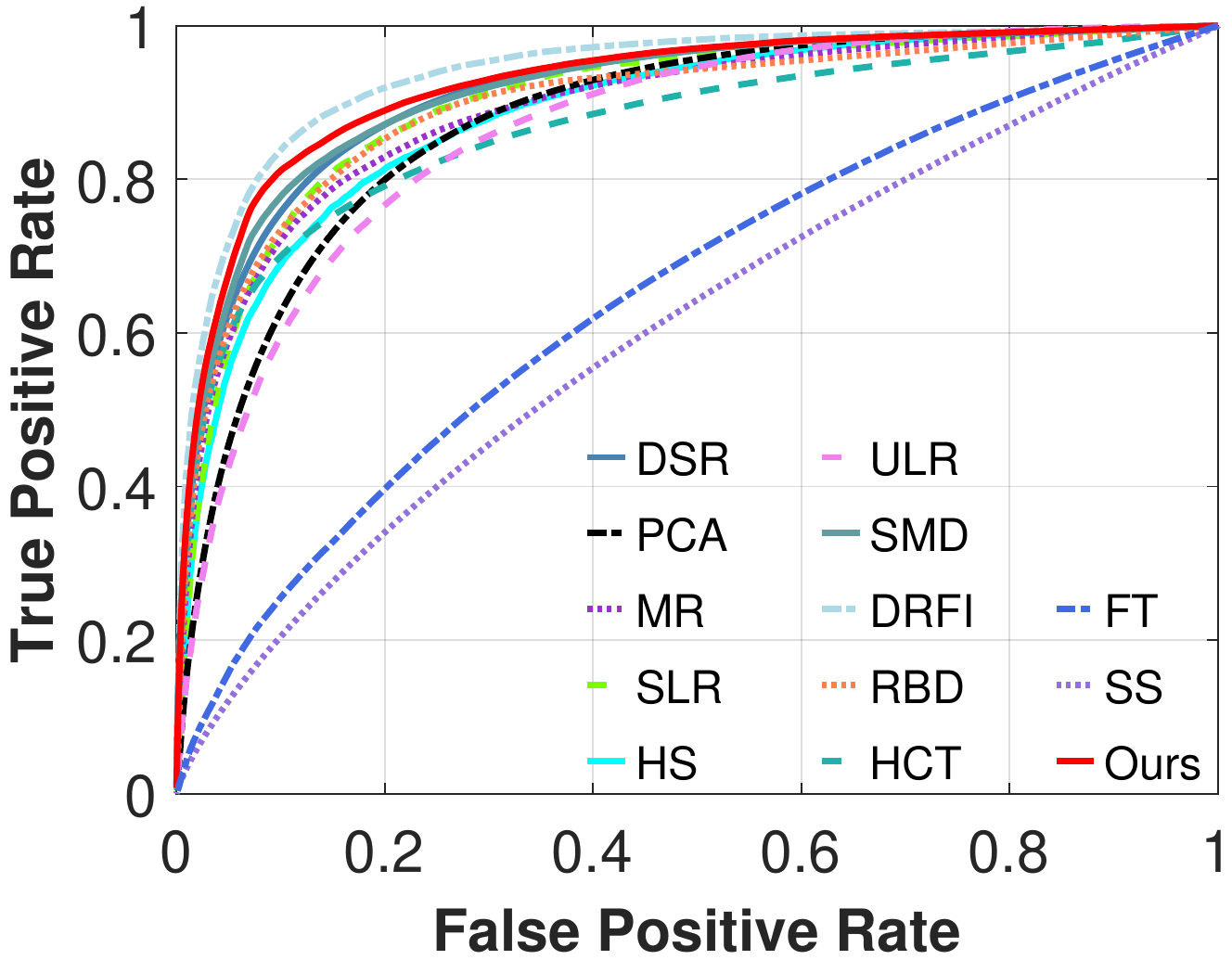} \\
    (a)&(b)&(c)
\end{tabular}
\caption{ROC curve of all methods. (a) results on MSRA10K dataset. (b) results on iCoSeg dataset. (c) results on ECSSD dataset}
\label{fig:roc}
\end{figure*}

To evaluate the superiority of our coarse-to-fine model, we systematically compare it with the other twelve state-of-the-arts. PR curves on three datasets are shown in Fig.~\ref{fig:pr_curve}, ROC curves are shown on Fig.~\ref{fig:roc}, and results of four metrics mentioned above are listed in Table~\ref{tab:res}. Besides, qualitative comparisons are provided in Fig.~\ref{fig:vis}. From the results we can see that, in most cases, our model ranks first or second on the three datasets under different criteria. It is worth noting that we report the result of DRFI~\cite{wang2017salient} as a reference, which belongs to top-down methods with supervised training.

\setlength{\tabcolsep}{0.5pt}
\setlength\extrarowheight{3pt}
\begin{table*}[!ht]
    \centering
    \caption{WF, OR, AUC, MAE of all methods on (a) MSRA10K, (b) iCoSeg and (c) ECSSD. \protect\linebreak The best three results are marked with red, green and blue respectively.}
    \begin{tabular}{c|c|c c c c c c c c c c c c c}
    \hline
        \multirow{5}{*}{(a)} & \footnotesize{Metric} ~&~ \footnotesize{\textbf{Ours}} & \footnotesize{SMD\cite{peng2017salient}} & \footnotesize{DRFI\cite{wang2017salient}} & \footnotesize{RBD\cite{zhu2014saliency}} & \footnotesize{HCT\cite{kim2016salient}} & \footnotesize{DSR\cite{li2013saliency}} & \footnotesize{PCA\cite{margolin2013makes}} & \footnotesize{MR\cite{yang2013saliency}} & \footnotesize{SLR\cite{zou2013segmentation}} & \footnotesize{SS\cite{hou2012image}} & \footnotesize{ULR\cite{shen2012unified}} & \footnotesize{HS\cite{yan2013hierarchical}} & \footnotesize{FT\cite{achanta2009frequency}} \\ \cline{2-15}
         & WF$\uparrow$ & {\color{red}0.705} & {\color{green}0.704} & 0.666 & {\color{blue}0.685} & 0.582 & 0.656 & 0.473 & 0.642 & 0.601 & 0.137 & 0.425 & 0.604 & 0.277\\
         & OR$\uparrow$ & {\color{red}0.751} & {\color{green}0.741} & {\color{blue}0.723} & 0.716 & 0.674 & 0.654 & 0.576 & 0.693 & 0.691 & 0.148 & 0.524 & 0.656 & 0.379\\
         & AUC$\uparrow$ & {\color{green}0.854} & {\color{blue}0.847} & {\color{red}0.857} & 0.834 & {\color{blue}0.847} & 0.825 & 0.839 & 0.601 & 0.840 & 0.801 & 0.831 & 0.833 & 0.690\\
         & MAE$\downarrow$ & {\color{red}0.104} & {\color{red}0.104} & 0.114 & {\color{green}0.108} & 0.143 & 0.121 & 0.185 & 0.125 & 0.141 & 0.255 & 0.224 & 0.149 & 0.231\\ \hline
         \hline
         \multirow{5}{*}{(b)} & \footnotesize{Metric} ~&~ \footnotesize{\textbf{Ours}} & \footnotesize{SMD\cite{peng2017salient}} & \footnotesize{DRFI\cite{wang2017salient}} & \footnotesize{RBD\cite{zhu2014saliency}} & \footnotesize{HCT\cite{kim2016salient}} & \footnotesize{DSR\cite{li2013saliency}} & \footnotesize{PCA\cite{margolin2013makes}} & \footnotesize{MR\cite{yang2013saliency}} & \footnotesize{SLR\cite{zou2013segmentation}} & \footnotesize{SS\cite{hou2012image}} & \footnotesize{ULR\cite{shen2012unified}} & \footnotesize{HS\cite{yan2013hierarchical}} & \footnotesize{FT\cite{achanta2009frequency}} \\ \cline{2-15}
         & WF$\uparrow$ & {\color{red}0.634} & {\color{green}0.611} & 0.592 & {\color{blue}0.599} & 0.464 & 0.548 & 0.407 & 0.554 & 0.473 & 0.126 & 0.379 & 0.563 & 0.289\\
         & OR$\uparrow$ & {\color{red}0.624} & {\color{green}0.598} & 0.582 & {\color{blue}0.588} & 0.519 & 0.514 & 0.427 & 0.573 & 0.505 & 0.164 & 0.443 & 0.537 & 0.387\\
         & AUC$\uparrow$ & {\color{green}0.838} & 0.822 & {\color{red}0.839} & 0.827 & {\color{blue}0.833} & 0.801 & 0.798 & 0.795 & 0.805 & 0.630 & 0.814 & 0.812 & 0.717\\
         & MAE$\downarrow$ & {\color{red}0.131} & {\color{green}0.138} & 0.139 & {\color{green}0.138} & 0.179 & 0.153 & 0.201 & 0.162 & 0.179 & 0.253 & 0.222 & 0.176 & 0.223 \\ \hline
         \hline
         \multirow{5}{*}{(c)} & \footnotesize{Metric} ~&~ \footnotesize{\textbf{Ours}} & \footnotesize{SMD\cite{peng2017salient}} & \footnotesize{DRFI\cite{wang2017salient}} & \footnotesize{RBD\cite{zhu2014saliency}} & \footnotesize{HCT\cite{kim2016salient}} & \footnotesize{DSR\cite{li2013saliency}} & \footnotesize{PCA\cite{margolin2013makes}} & \footnotesize{MR\cite{yang2013saliency}} & \footnotesize{SLR\cite{zou2013segmentation}} & \footnotesize{SS\cite{hou2012image}} & \footnotesize{ULR\cite{shen2012unified}} & \footnotesize{HS\cite{yan2013hierarchical}} & \footnotesize{FT\cite{achanta2009frequency}} \\ \cline{2-15}
         & WF$\uparrow$    & {\color{green}0.545} & {\color{blue}0.544} & {\color{red}0.547} & 0.513 & 0.446 & 0.514 & 0.364 & 0.496 & 0.402 & 0.128 & 0.351 & 0.454 & 0.195 \\
         & OR$\uparrow$      & {\color{red}0.571} & {\color{blue}0.563} & {\color{green}0.568} & 0.526 & 0.486 & 0.514 & 0.395 & 0.523 & 0.486 & 0.103 & 0.369 & 0.458 & 0.216 \\
         & AUC$\uparrow$     & {\color{red}0.820} & {\color{blue}0.813} & {\color{green}0.817} & 0.781 & 0.785 & 0.785 & 0.791 & 0.793 & 0.805 & 0.567 & 0.788 & 0.801 & 0.607 \\
         & MAE$\downarrow$ & 0.176 & 0.174 & {\color{red}0.160} & {\color{green}0.171} & 0.198 & {\color{green}0.171} & 0.247 & 0.186 & 0.226 & 0.278 & 0.274 & 0.227 & 0.270 \\ \hline
    \end{tabular}
    \label{tab:res}
\end{table*}

\subsubsection{Results on single-object images}
The MSRA10K dataset contains images with diverse objects of varying size, and with only one object in each image. From Fig.~\ref{fig:pr_curve} (a), Fig.~\ref{fig:roc} (a) and Table~\ref{tab:res} (a), we can see that our method achieves the best result with the highest weighted F-measure, overlapping ratio and the lowest mean average error, while DRFI~\cite{wang2017salient} obtains the highest AUC score. It is worth noting that, our method even outperforms DRFI~\cite{wang2017salient} with just simple features and no supervision. Frequency-based methods like FT~\cite{achanta2009frequency} perform badly, as it is difficult to choose a proper scale to suppress background without knowing of object size. While SS~\cite{hou2012image} considers sparsity directly in standard spatial space and DCT space, it can only give a rough result of detected objects. In PR curves, our method shows an obvious superiority to other approaches. While in ROC curves, DRFI~\cite{wang2017salient} and our method are the best two among those competitive methods.

\subsubsection{Results on multiple-object images}
The iCoSeg dataset contains images with multiple objects, separate or adjacent. From Fig.~\ref{fig:pr_curve} (b), Fig.~\ref{fig:roc} (b) and Table~\ref{tab:res} (b), we can see that our method also achieves the highest weighted F-measure, overlapping ratio and the lowest mean average error, which shows that our method is effective under cases of multiple objects. However, the performance of PCA~\cite{margolin2013makes}, SLR~\cite{zou2013segmentation}, DSR~\cite{li2013saliency} and ULR~\cite{shen2012unified} decrease heavily. As PCA~\cite{margolin2013makes} considers the dissimilarity between image patches and SLR~\cite{zou2013segmentation} introduces a segmentation prior, they are more sensitive to the quantity of object within a scene. As for DSR~\cite{li2013saliency}, its precision drops dramatically with the increase of recall due to its dependence on background templates. This is because in the scenario of multiple objects, salient objects are more likely to overlap with image boundary regions. ULR~\cite{shen2012unified} trains a feature transformation on MSRA dataset, hence it obtains poor performance for the detection of multiple objects. In PR curves, our method presents better stability with increased recall. While in ROC curves, our method and DRFI~\cite{wang2017salient} achieve the best performance and almost the same AUC score, outperforming the rest approaches.

\subsubsection{Results on complex scene images}
The ECSSD dataset contains images with complicated background and also objects of varying size. From Fig.~\ref{fig:pr_curve} (c), Fig.~\ref{fig:roc} (c) and Table~\ref{tab:res} (c), we can see that our method achieves the highest overlapping ratio and AUC score, and is outperformed by DRFI~\cite{wang2017salient} in terms of weighted F-measure and mean absolute error. In PR curves, our method performs similarly to SMD~\cite{peng2017salient}, while in ROC curves, DRFI~\cite{wang2017salient} and our method are the best two among the state-of-the-arts. The result demonstrates that our method is competitive under complex scene. Approaches such as HS~\cite{yan2013hierarchical}, HCT~\cite{kim2016salient}, MR~\cite{yang2013saliency} and RBD~\cite{zhu2014saliency} that depend on cues like contrast bias and center bias fail to maintain good performance.

\subsubsection{Visual \& efficiency comparison}
\begin{figure*}[!th]
\centering
\begin{tabular}{ccccccccccccccc}
    \includegraphics[width=0.062\textwidth]{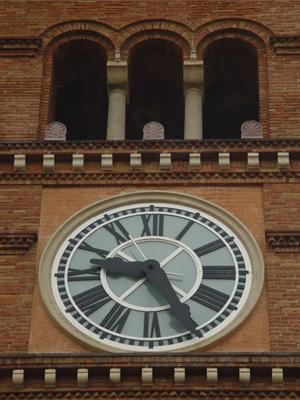} &
    \includegraphics[width=0.062\textwidth]{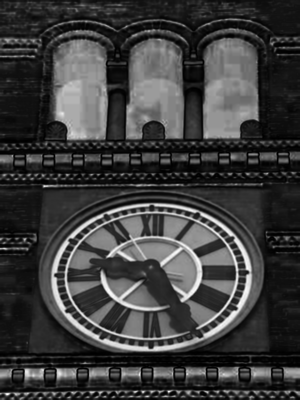} &
    \includegraphics[width=0.062\textwidth]{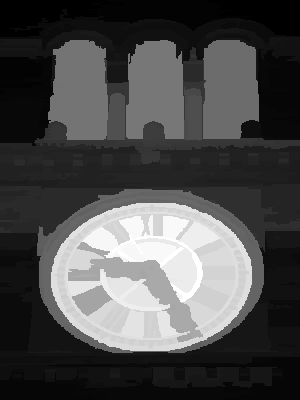} &
    \includegraphics[width=0.062\textwidth]{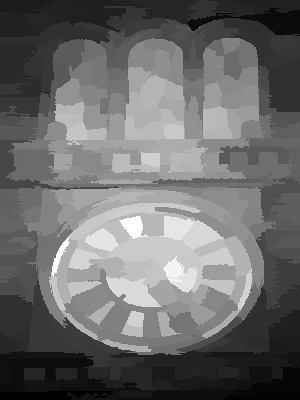} &
    \includegraphics[width=0.062\textwidth]{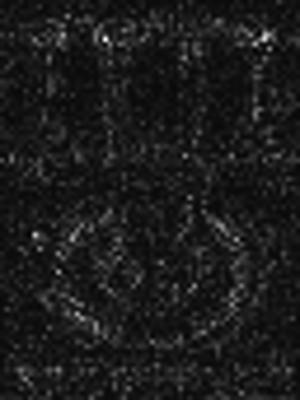} &
    \includegraphics[width=0.062\textwidth]{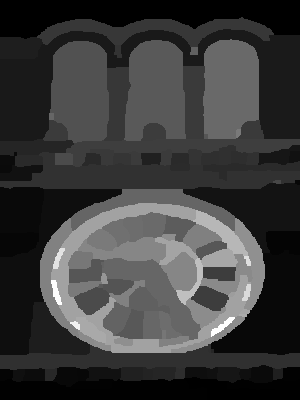} &
    \includegraphics[width=0.062\textwidth]{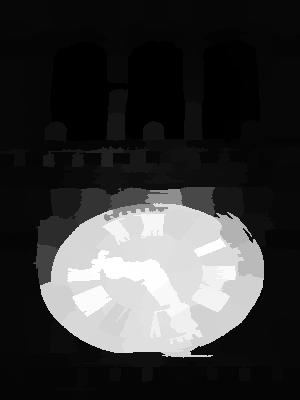} &
    \includegraphics[width=0.062\textwidth]{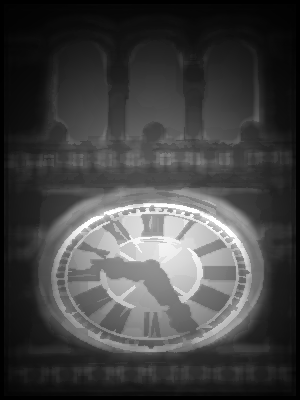} &
    \includegraphics[width=0.062\textwidth]{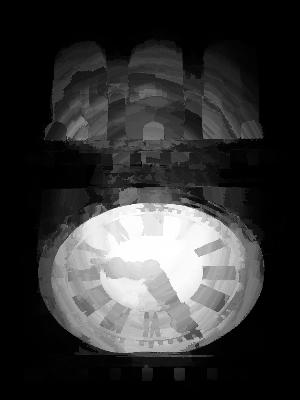} &
    \includegraphics[width=0.062\textwidth]{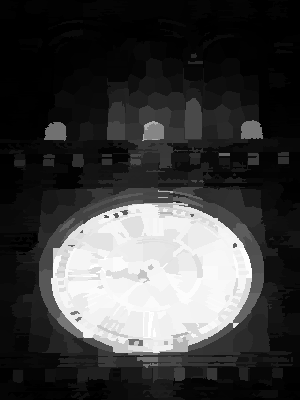} &
    \includegraphics[width=0.062\textwidth]{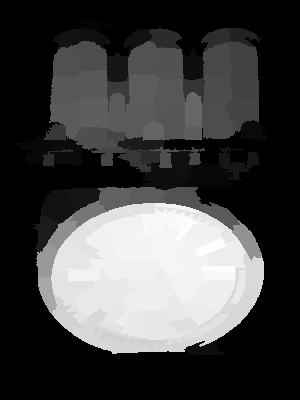} &
    \includegraphics[width=0.062\textwidth]{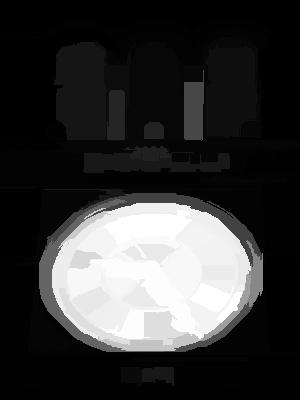} &
    \includegraphics[width=0.062\textwidth]{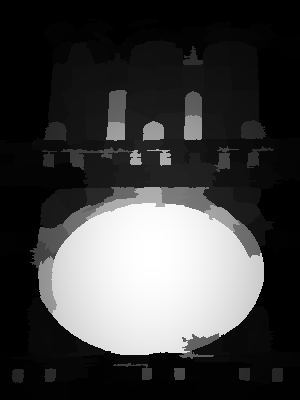} &
    \includegraphics[width=0.062\textwidth]{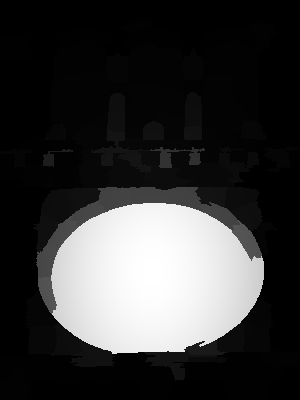} &
    \includegraphics[width=0.062\textwidth]{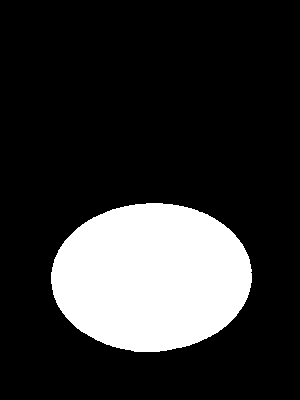} \\

    \includegraphics[width=0.062\textwidth]{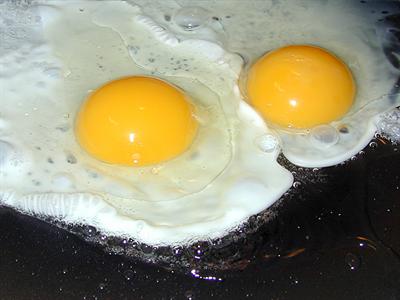} &
    \includegraphics[width=0.062\textwidth]{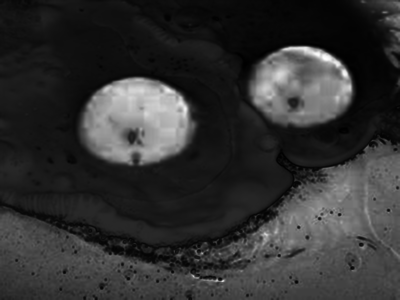} &
    \includegraphics[width=0.062\textwidth]{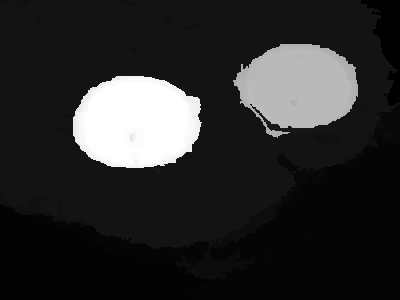} &
    \includegraphics[width=0.062\textwidth]{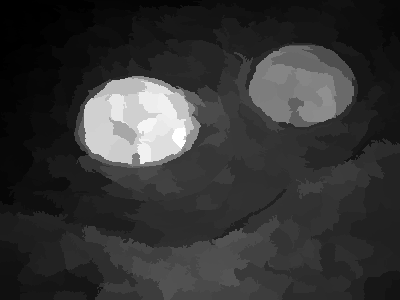} &
    \includegraphics[width=0.062\textwidth]{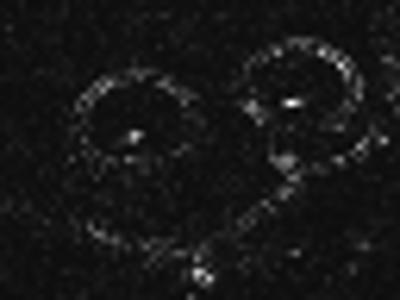} &
    \includegraphics[width=0.062\textwidth]{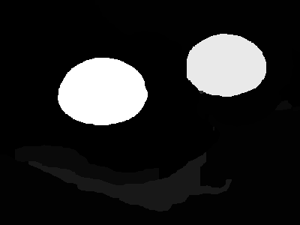} &
    \includegraphics[width=0.062\textwidth]{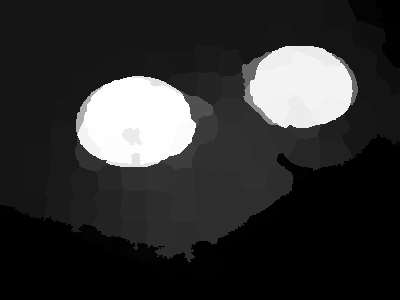} &
    \includegraphics[width=0.062\textwidth]{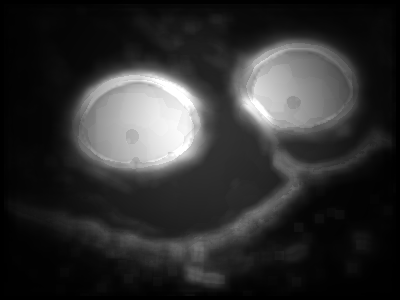} &
    \includegraphics[width=0.062\textwidth]{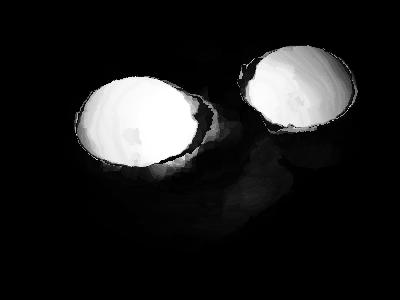} &
    \includegraphics[width=0.062\textwidth]{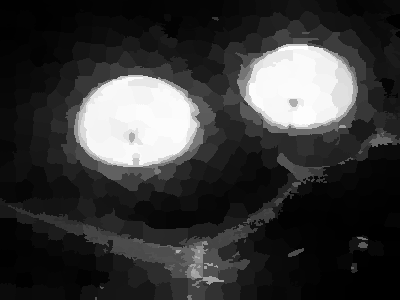} &
    \includegraphics[width=0.062\textwidth]{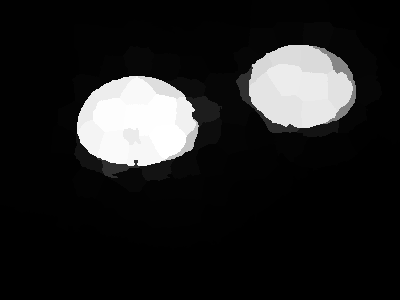} &
    \includegraphics[width=0.062\textwidth]{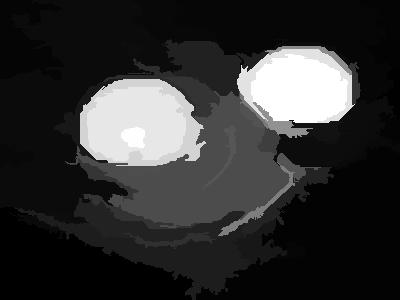} &
    \includegraphics[width=0.062\textwidth]{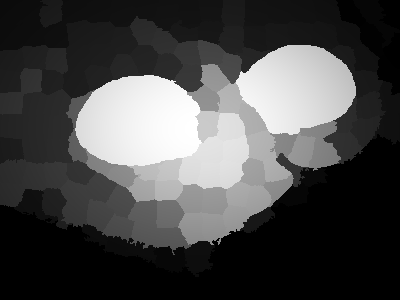} &
    \includegraphics[width=0.062\textwidth]{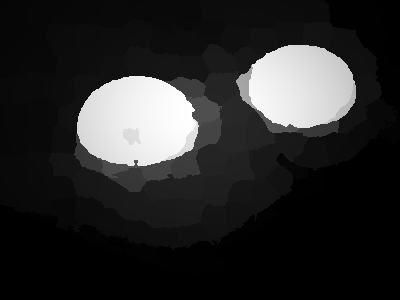} &
    \includegraphics[width=0.062\textwidth]{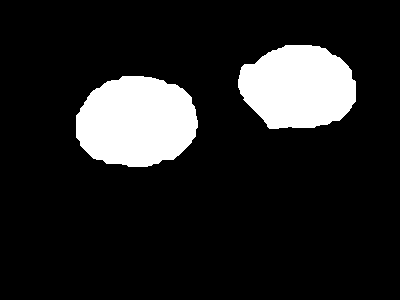} \\

    \includegraphics[width=0.062\textwidth]{figs/1626_rgb.jpg} &
    \includegraphics[width=0.062\textwidth]{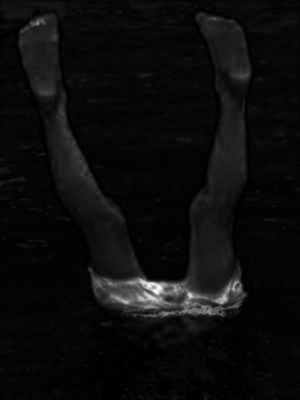} &
    \includegraphics[width=0.062\textwidth]{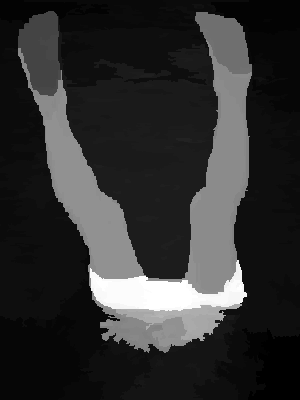} &
    \includegraphics[width=0.062\textwidth]{figs/1626_ulr.png} &
    \includegraphics[width=0.062\textwidth]{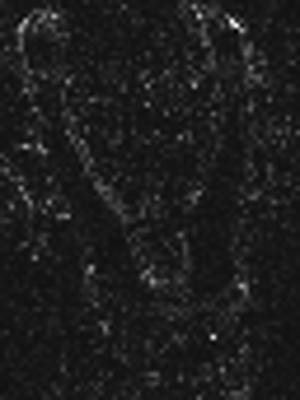} &
    \includegraphics[width=0.062\textwidth]{figs/1626_slr.png} &
    \includegraphics[width=0.062\textwidth]{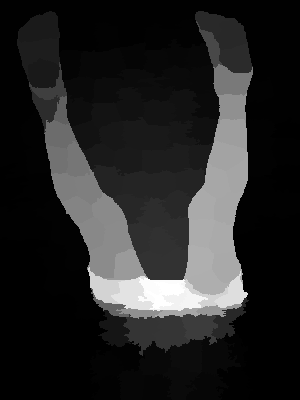} &
    \includegraphics[width=0.062\textwidth]{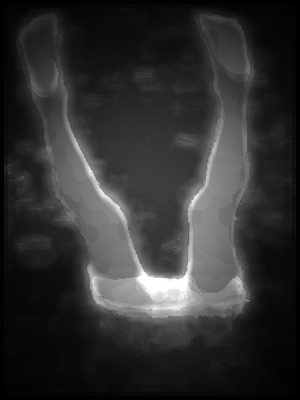} &
    \includegraphics[width=0.062\textwidth]{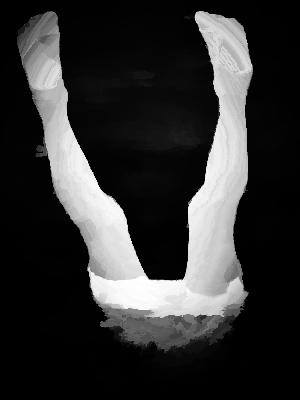} &
    \includegraphics[width=0.062\textwidth]{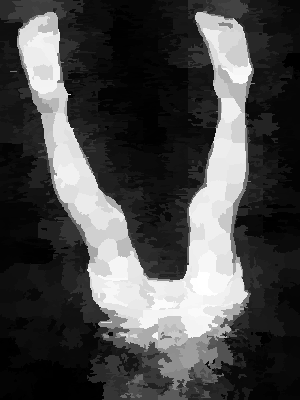} &
    \includegraphics[width=0.062\textwidth]{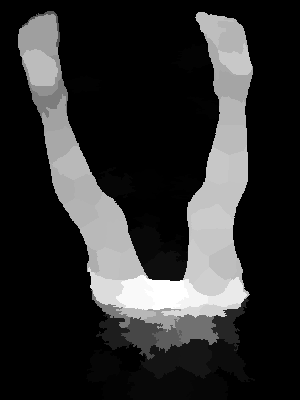} &
    \includegraphics[width=0.062\textwidth]{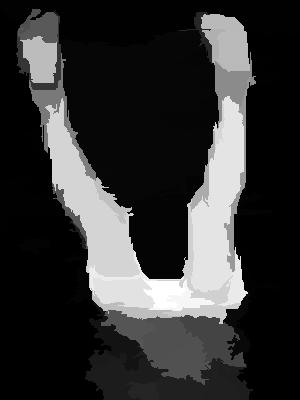} &
    \includegraphics[width=0.062\textwidth]{figs/1626_smd.png} &
    \includegraphics[width=0.062\textwidth]{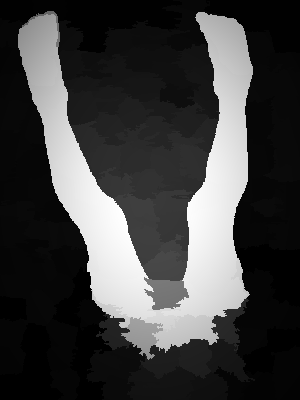} &
    \includegraphics[width=0.062\textwidth]{figs/1626_gt.png} \\

    \includegraphics[width=0.062\textwidth]{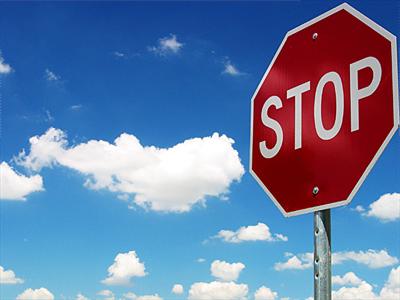} &
    \includegraphics[width=0.062\textwidth]{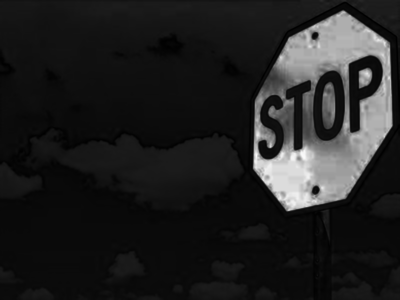} &
    \includegraphics[width=0.062\textwidth]{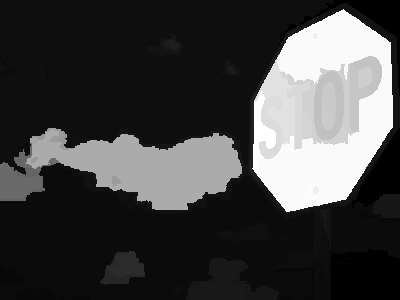} &
    \includegraphics[width=0.062\textwidth]{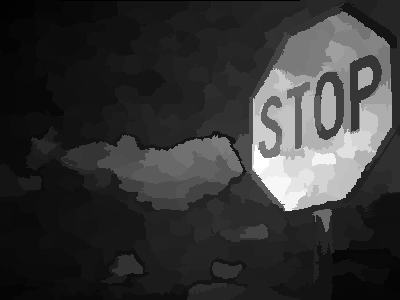} &
    \includegraphics[width=0.062\textwidth]{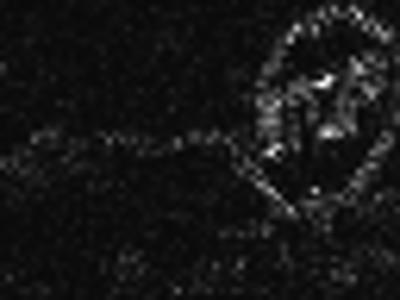} &
    \includegraphics[width=0.062\textwidth]{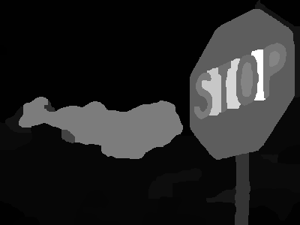} &
    \includegraphics[width=0.062\textwidth]{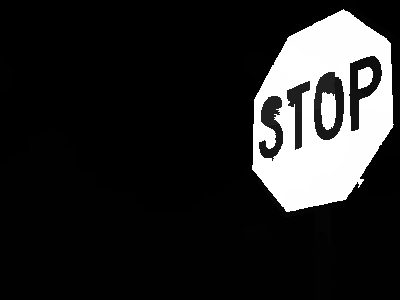} &
    \includegraphics[width=0.062\textwidth]{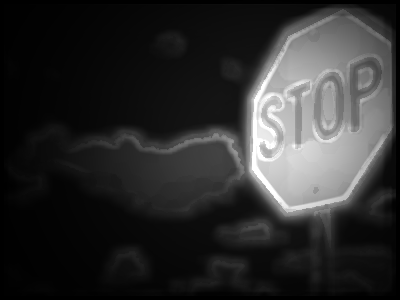} &
    \includegraphics[width=0.062\textwidth]{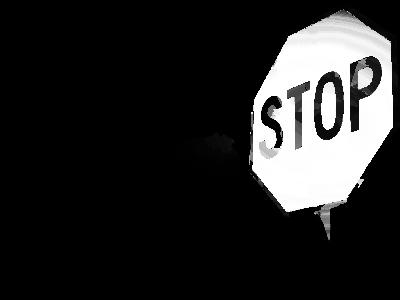} &
    \includegraphics[width=0.062\textwidth]{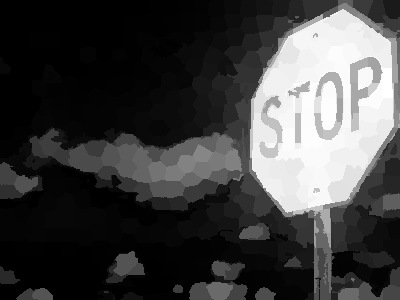} &
    \includegraphics[width=0.062\textwidth]{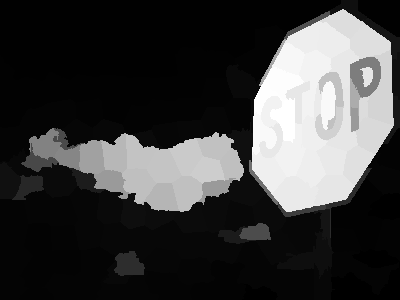} &
    \includegraphics[width=0.062\textwidth]{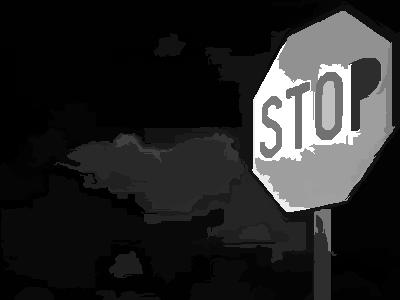} &
    \includegraphics[width=0.062\textwidth]{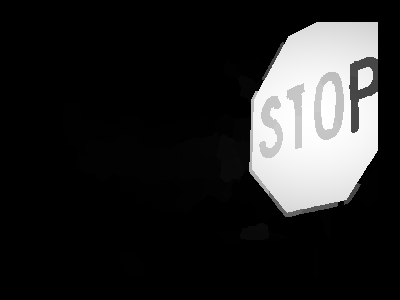} &
    \includegraphics[width=0.062\textwidth]{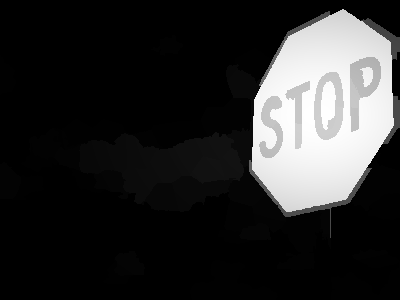} &
    \includegraphics[width=0.062\textwidth]{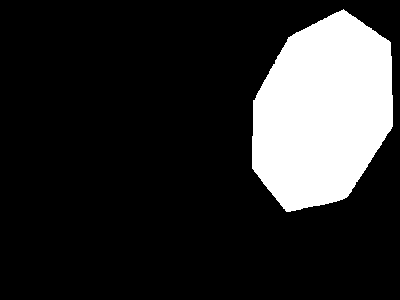} \\

    \includegraphics[width=0.062\textwidth]{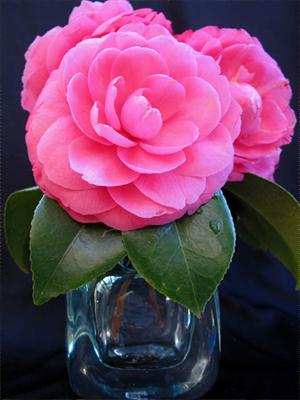} &
    \includegraphics[width=0.062\textwidth]{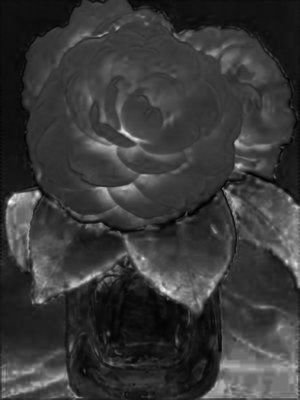} &
    \includegraphics[width=0.062\textwidth]{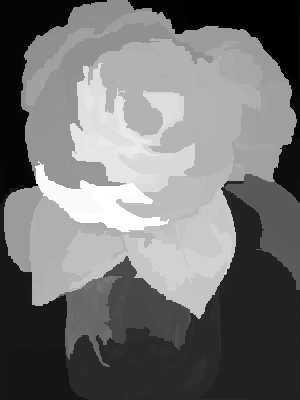} &
    \includegraphics[width=0.062\textwidth]{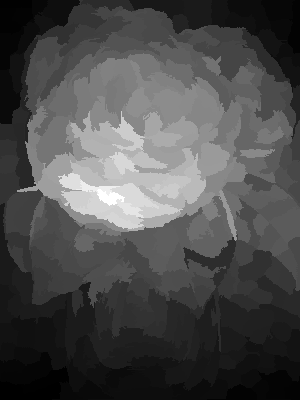} &
    \includegraphics[width=0.062\textwidth]{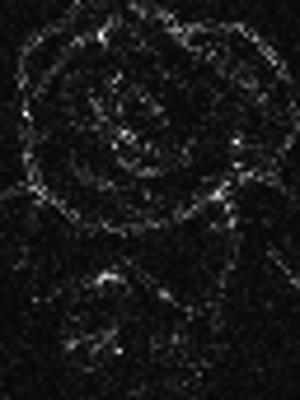} &
    \includegraphics[width=0.062\textwidth]{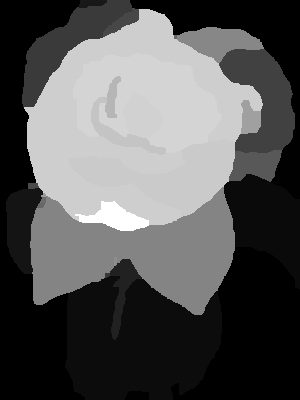} &
    \includegraphics[width=0.062\textwidth]{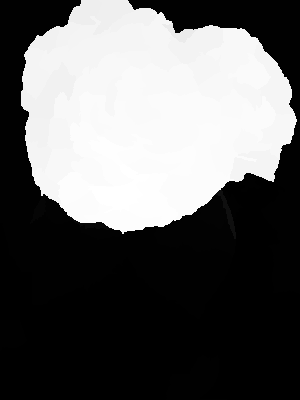} &
    \includegraphics[width=0.062\textwidth]{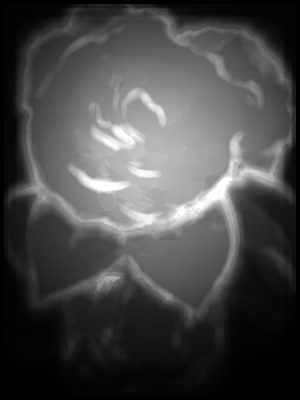} &
    \includegraphics[width=0.062\textwidth]{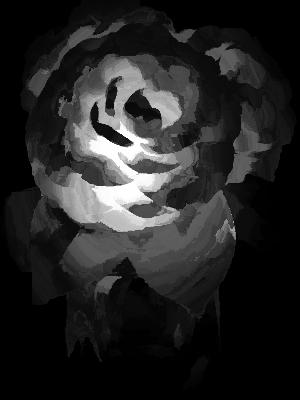} &
    \includegraphics[width=0.062\textwidth]{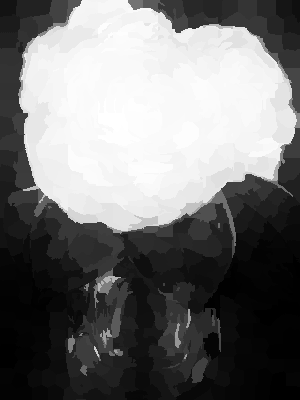} &
    \includegraphics[width=0.062\textwidth]{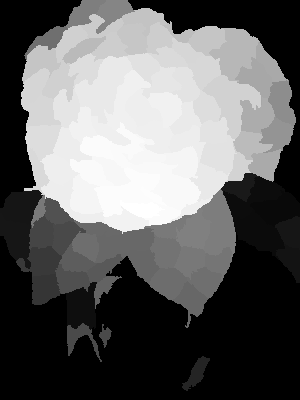} &
    \includegraphics[width=0.062\textwidth]{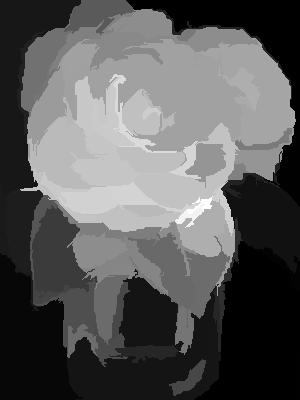} &
    \includegraphics[width=0.062\textwidth]{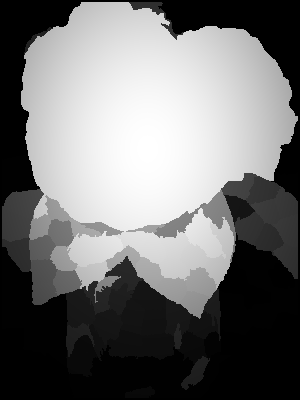} &
    \includegraphics[width=0.062\textwidth]{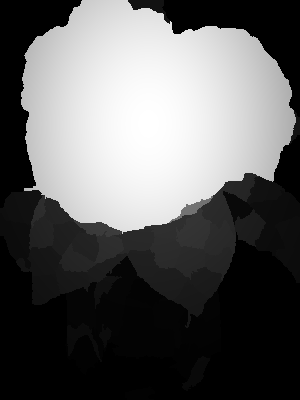} &
    \includegraphics[width=0.062\textwidth]{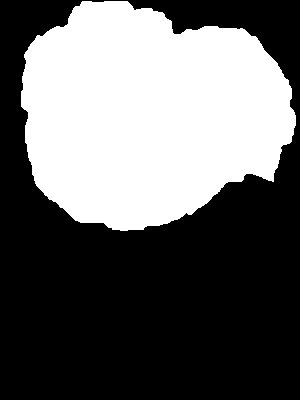} \\

    \includegraphics[width=0.062\textwidth]{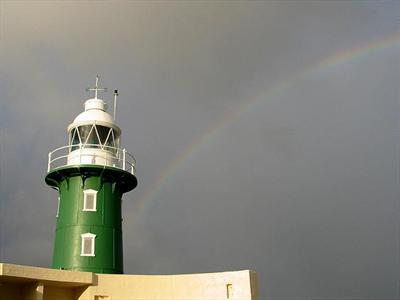} &
    \includegraphics[width=0.062\textwidth]{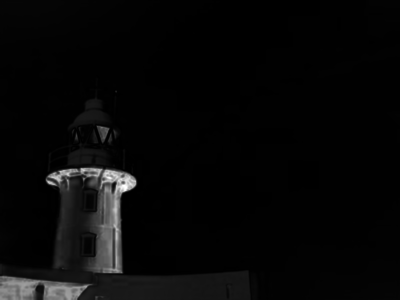} &
    \includegraphics[width=0.062\textwidth]{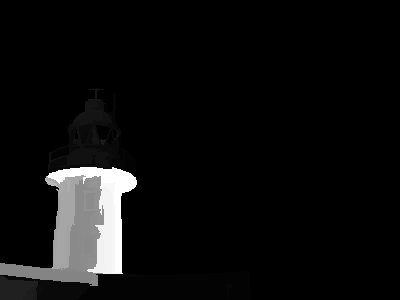} &
    \includegraphics[width=0.062\textwidth]{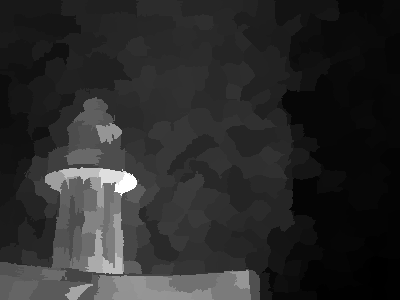} &
    \includegraphics[width=0.062\textwidth]{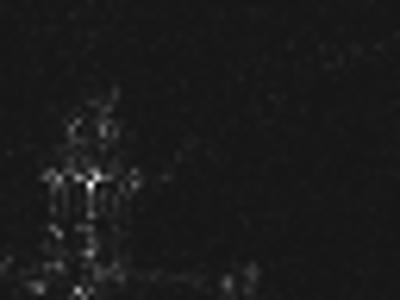} &
    \includegraphics[width=0.062\textwidth]{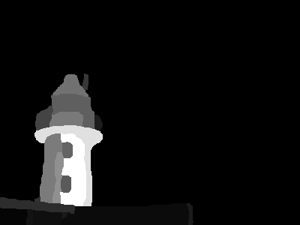} &
    \includegraphics[width=0.062\textwidth]{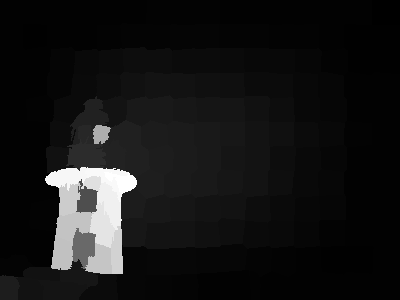} &
    \includegraphics[width=0.062\textwidth]{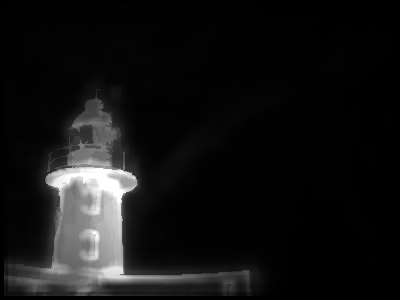} &
    \includegraphics[width=0.062\textwidth]{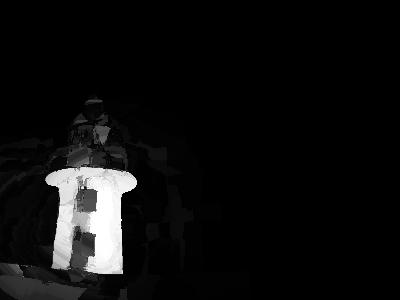} &
    \includegraphics[width=0.062\textwidth]{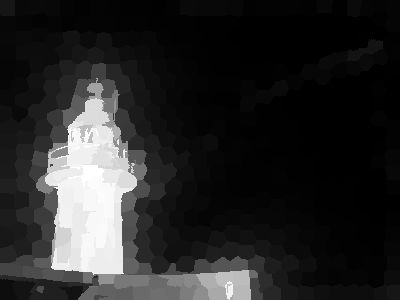} &
    \includegraphics[width=0.062\textwidth]{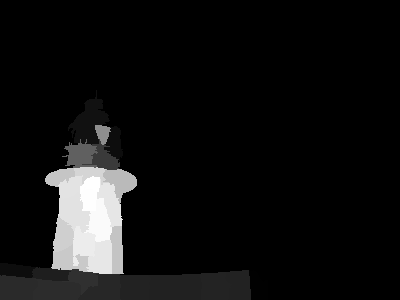} &
    \includegraphics[width=0.062\textwidth]{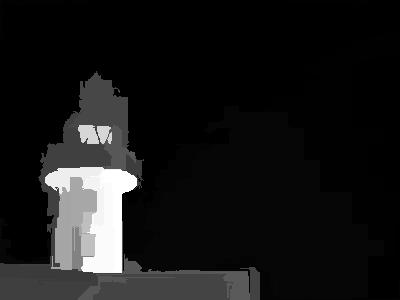} &
    \includegraphics[width=0.062\textwidth]{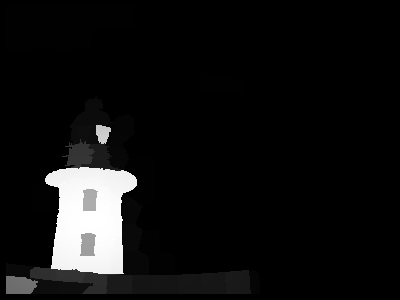} &
    \includegraphics[width=0.062\textwidth]{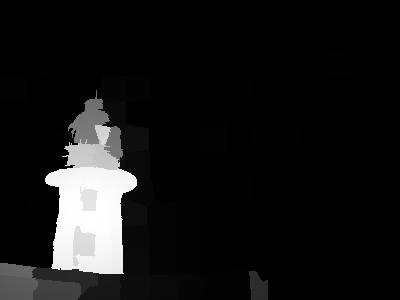} &
    \includegraphics[width=0.062\textwidth]{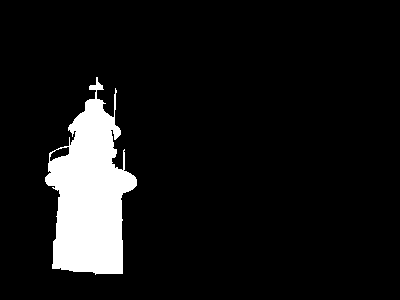} \\

    \includegraphics[width=0.062\textwidth]{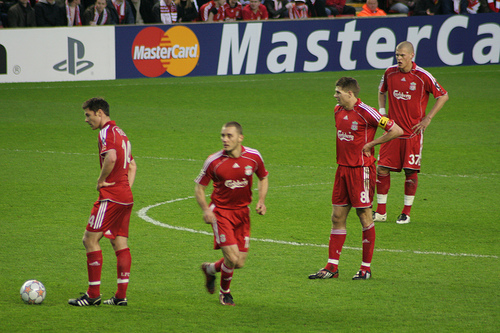} & \includegraphics[width=0.062\textwidth]{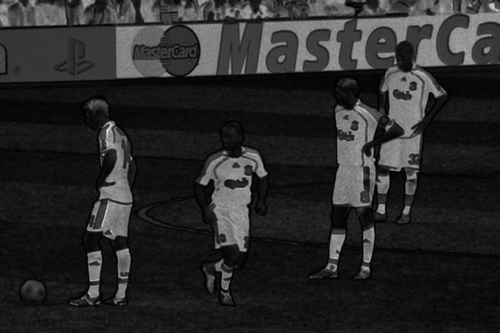} & \includegraphics[width=0.062\textwidth]{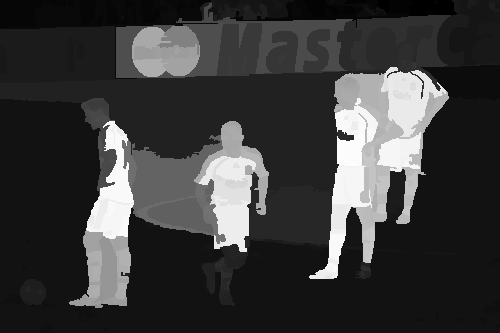} & \includegraphics[width=0.062\textwidth]{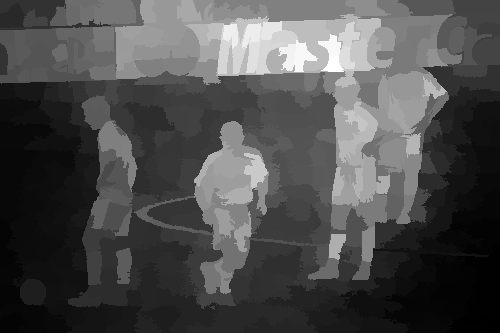} &
    \includegraphics[width=0.062\textwidth]{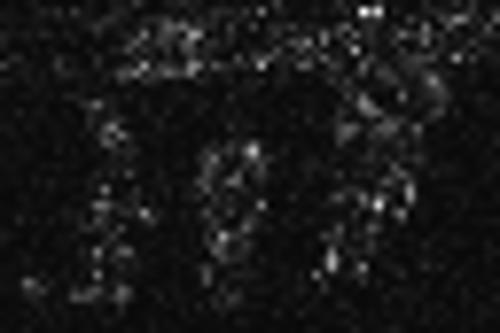} & \includegraphics[width=0.062\textwidth]{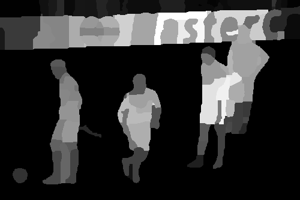} &
    \includegraphics[width=0.062\textwidth]{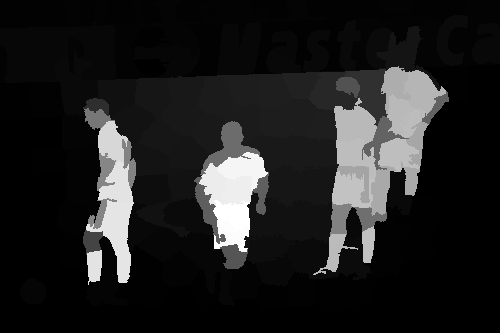} &
    \includegraphics[width=0.062\textwidth]{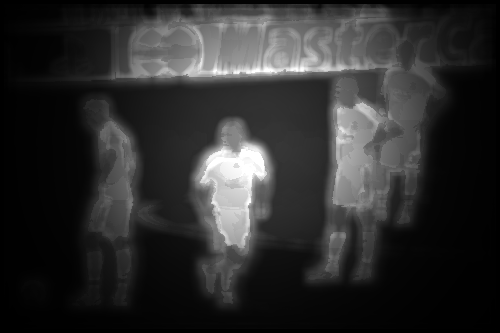} &
    \includegraphics[width=0.062\textwidth]{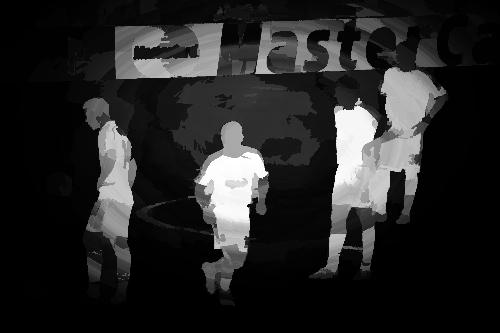} &
    \includegraphics[width=0.062\textwidth]{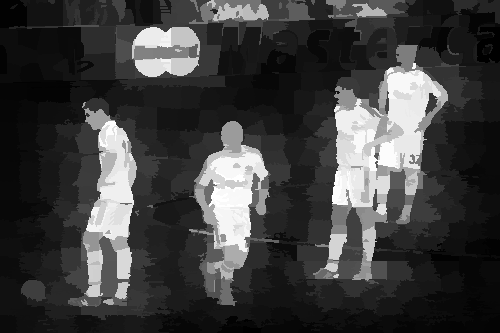} &
    \includegraphics[width=0.062\textwidth]{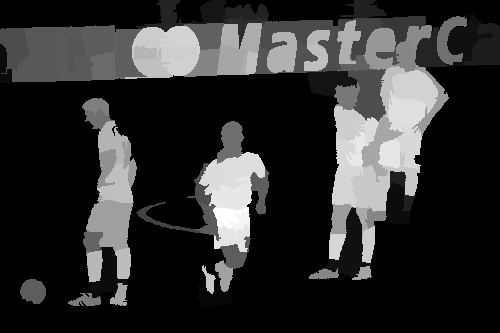} &
    \includegraphics[width=0.062\textwidth]{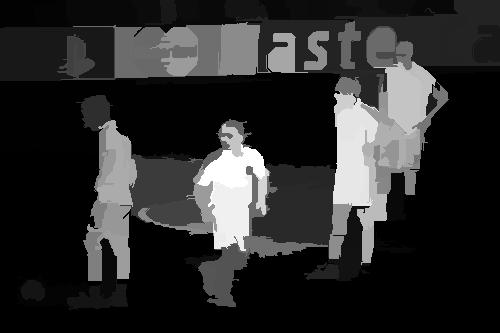} &
    \includegraphics[width=0.062\textwidth]{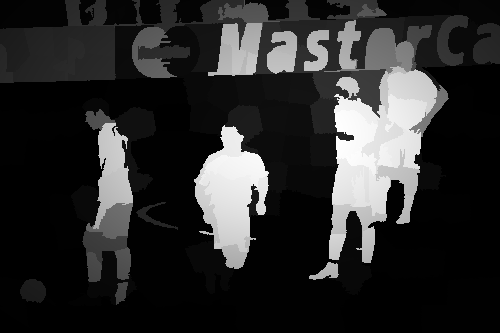} &
    \includegraphics[width=0.062\textwidth]{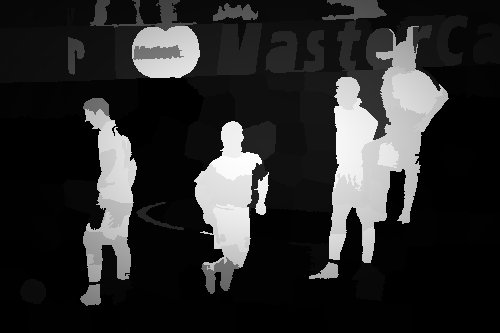} &
    \includegraphics[width=0.062\textwidth]{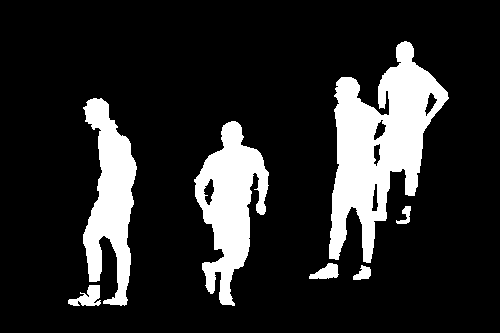} \\

    \includegraphics[width=0.062\textwidth]{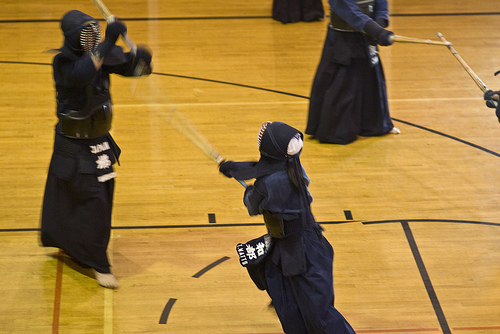} & \includegraphics[width=0.062\textwidth]{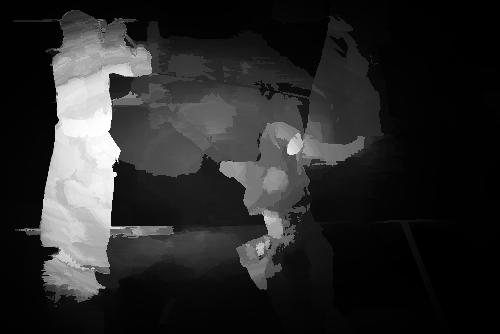} & \includegraphics[width=0.062\textwidth]{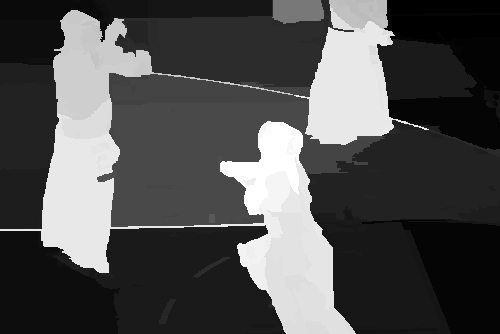} & \includegraphics[width=0.062\textwidth]{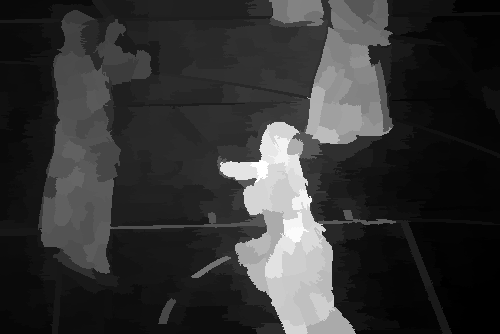} &
    \includegraphics[width=0.062\textwidth]{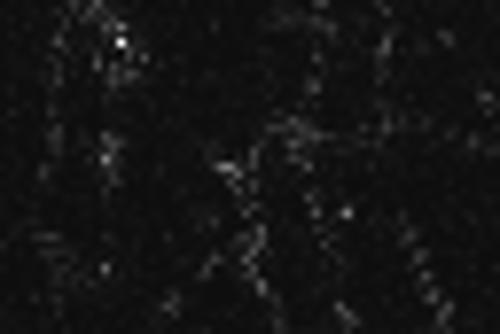} & \includegraphics[width=0.062\textwidth]{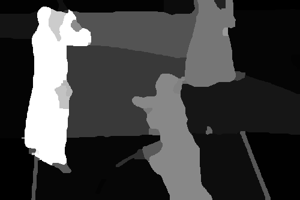} &
    \includegraphics[width=0.062\textwidth]{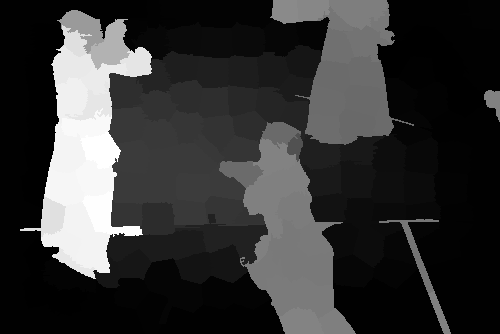} &
    \includegraphics[width=0.062\textwidth]{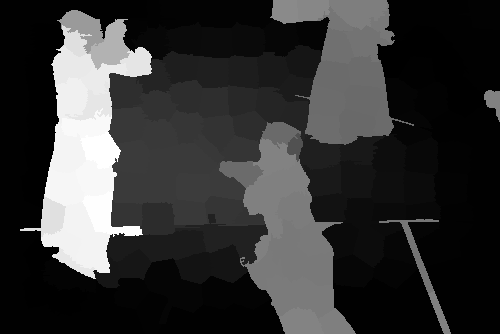} &
    \includegraphics[width=0.062\textwidth]{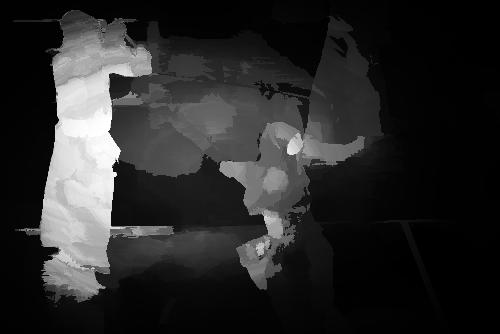} &
    \includegraphics[width=0.062\textwidth]{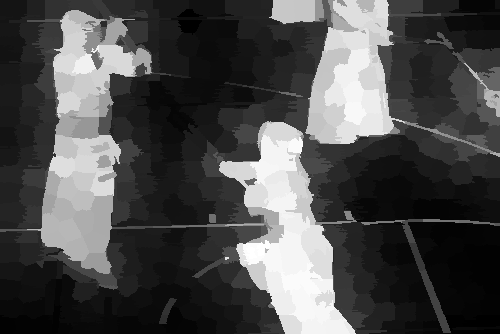} &
    \includegraphics[width=0.062\textwidth]{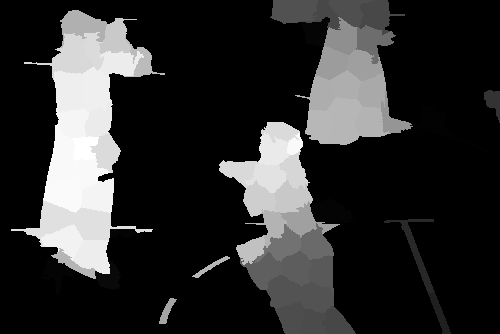} &
    \includegraphics[width=0.062\textwidth]{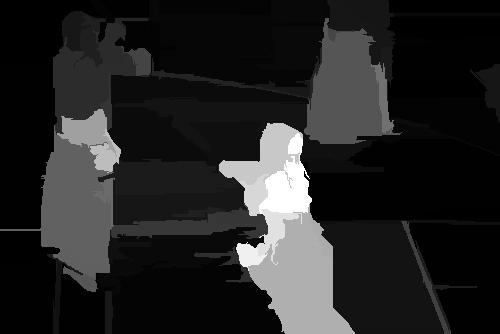} &
    \includegraphics[width=0.062\textwidth]{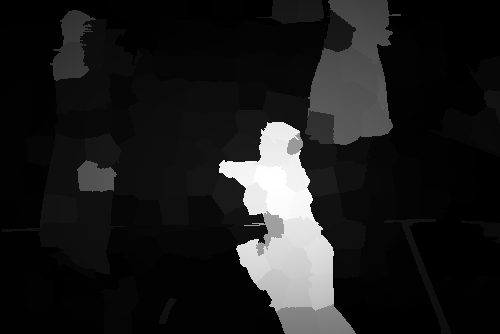} &
    \includegraphics[width=0.062\textwidth]{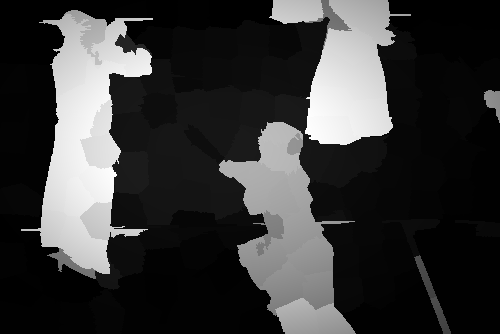} &
    \includegraphics[width=0.062\textwidth]{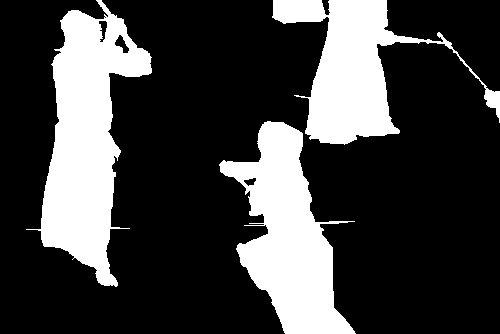} \\

    \includegraphics[width=0.062\textwidth]{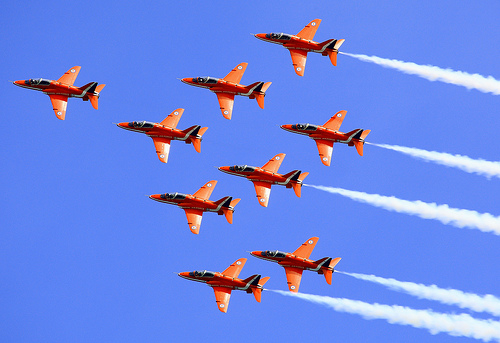} & \includegraphics[width=0.062\textwidth]{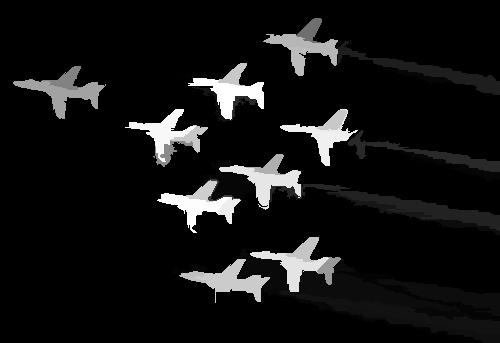} & \includegraphics[width=0.062\textwidth]{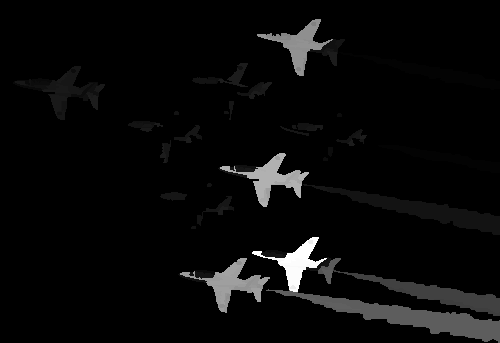} & \includegraphics[width=0.062\textwidth]{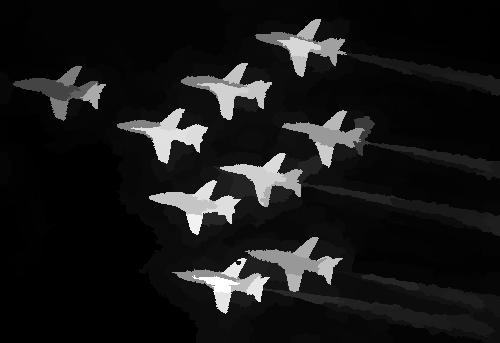} &
    \includegraphics[width=0.062\textwidth]{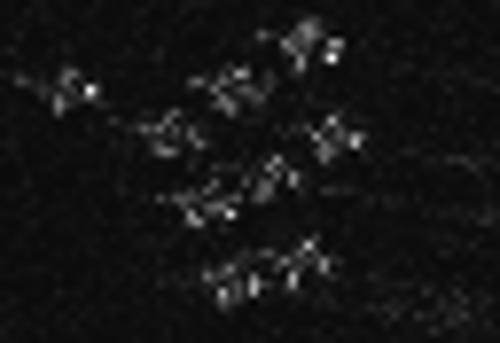} & \includegraphics[width=0.062\textwidth]{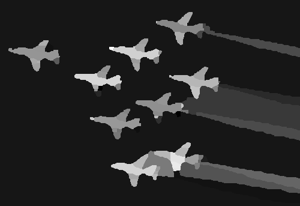} &
    \includegraphics[width=0.062\textwidth]{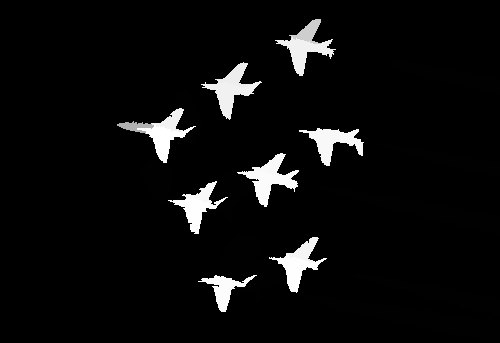} &
    \includegraphics[width=0.062\textwidth]{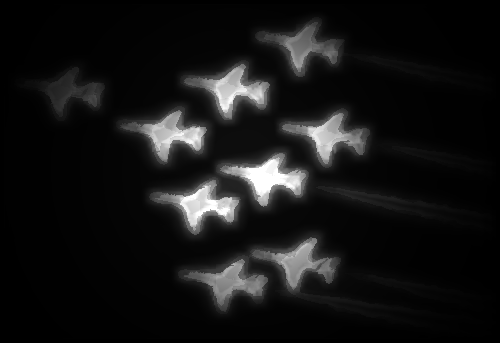} &
    \includegraphics[width=0.062\textwidth]{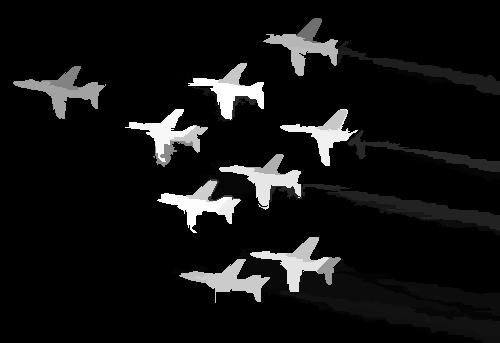} &
    \includegraphics[width=0.062\textwidth]{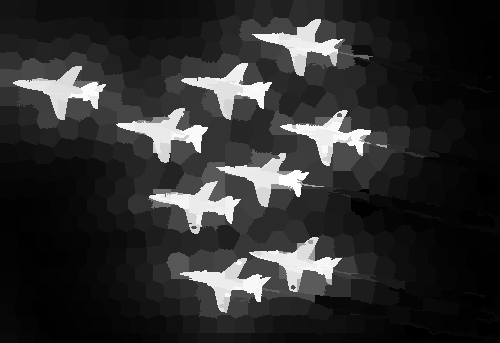} &
    \includegraphics[width=0.062\textwidth]{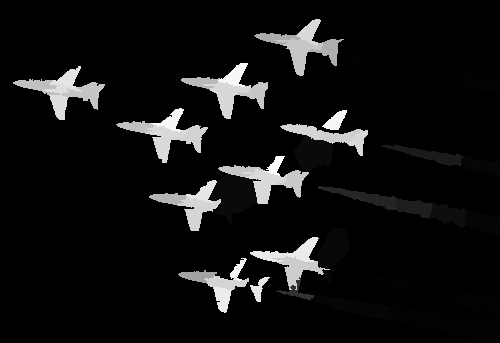} &
    \includegraphics[width=0.062\textwidth]{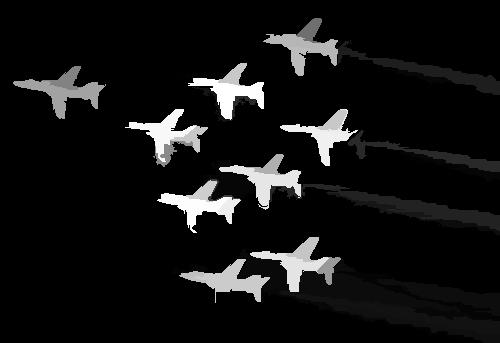} &
    \includegraphics[width=0.062\textwidth]{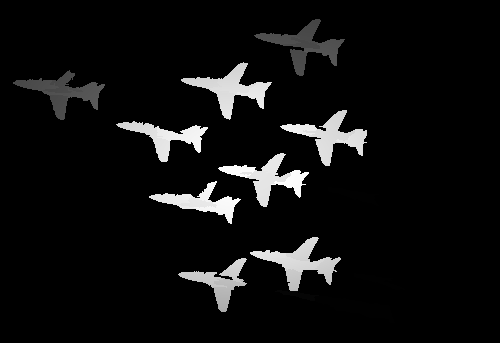} &
    \includegraphics[width=0.062\textwidth]{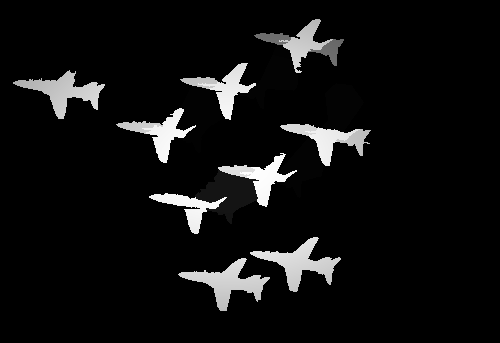} &
    \includegraphics[width=0.062\textwidth]{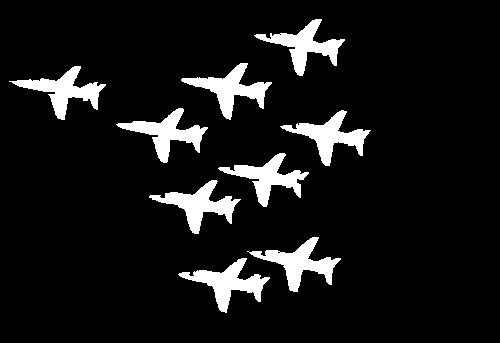} \\

    \includegraphics[width=0.062\textwidth]{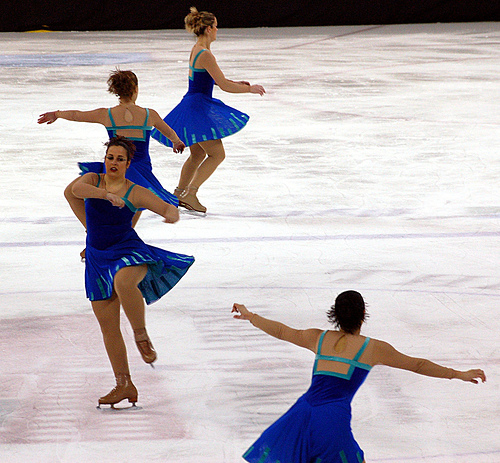} & \includegraphics[width=0.062\textwidth]{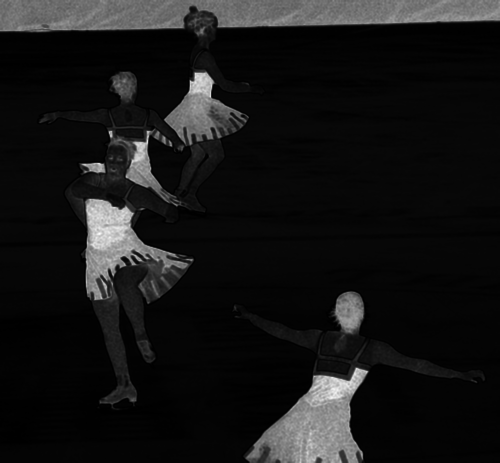} & \includegraphics[width=0.062\textwidth]{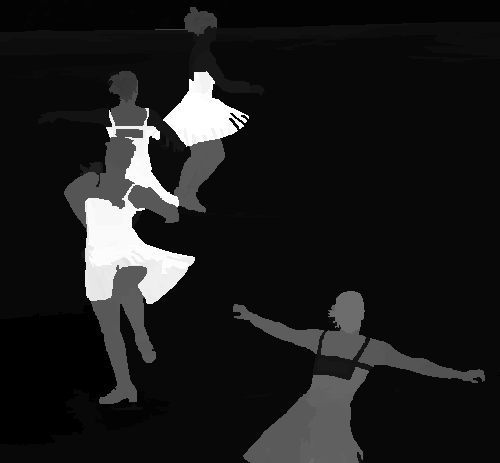} & \includegraphics[width=0.062\textwidth]{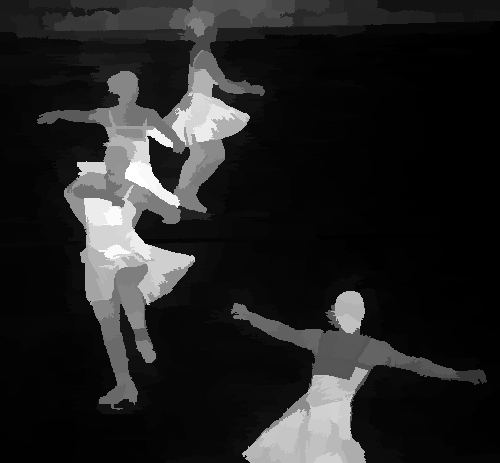} &
    \includegraphics[width=0.062\textwidth]{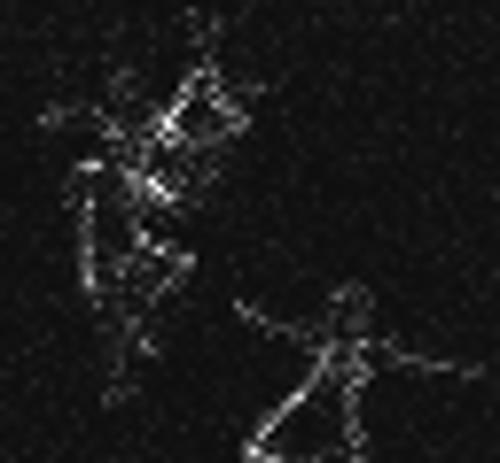} & \includegraphics[width=0.062\textwidth]{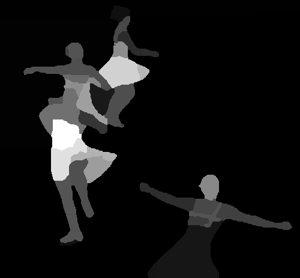} &
    \includegraphics[width=0.062\textwidth]{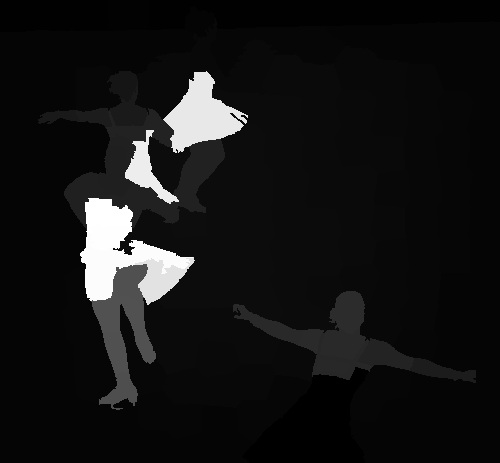} &
    \includegraphics[width=0.062\textwidth]{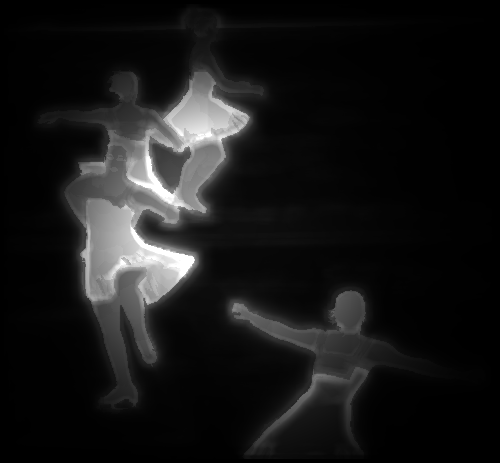} &
    \includegraphics[width=0.062\textwidth]{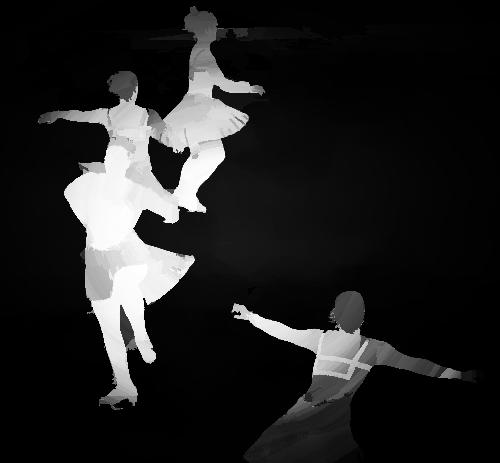} &
    \includegraphics[width=0.062\textwidth]{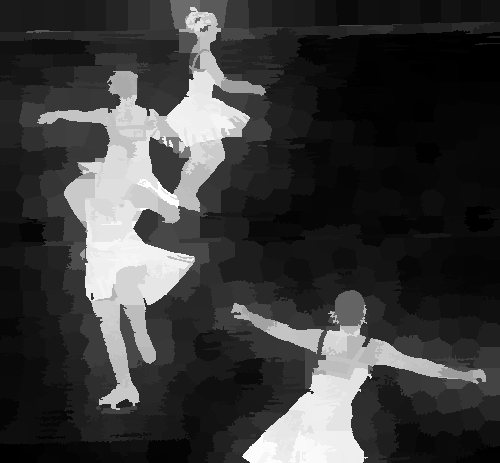} &
    \includegraphics[width=0.062\textwidth]{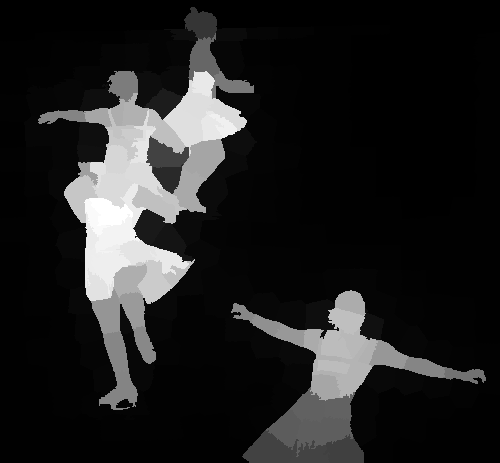} &
    \includegraphics[width=0.062\textwidth]{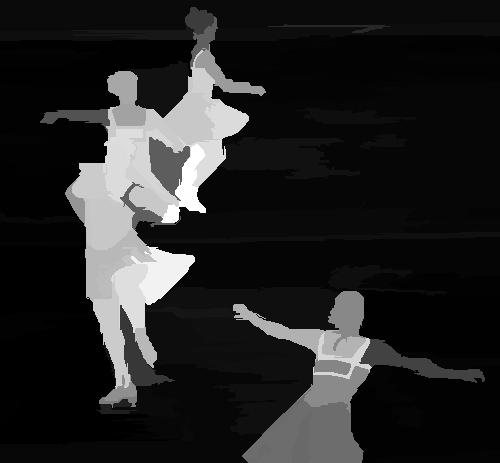} &
    \includegraphics[width=0.062\textwidth]{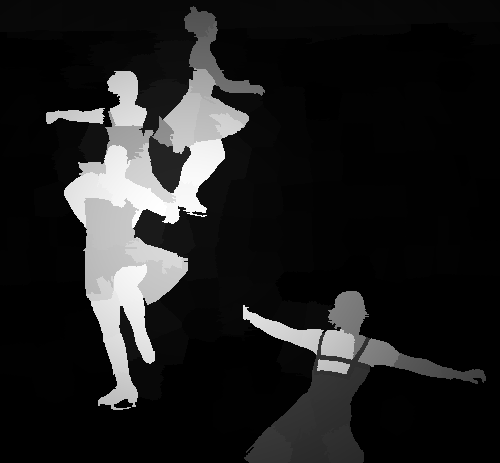} &
    \includegraphics[width=0.062\textwidth]{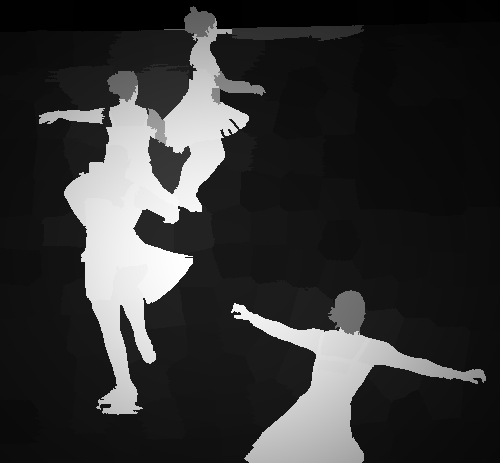} &
    \includegraphics[width=0.062\textwidth]{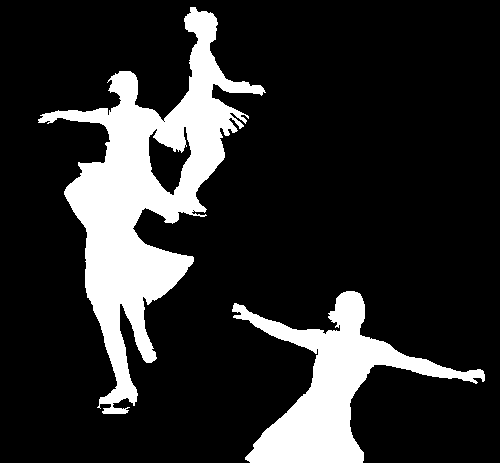} \\

    \includegraphics[width=0.062\textwidth]{figs/0057_rgb.jpg} & \includegraphics[width=0.062\textwidth]{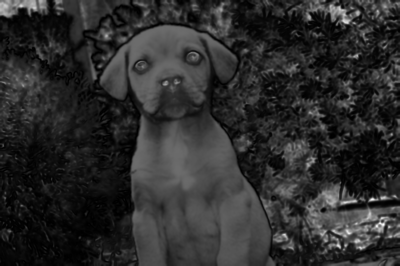} & \includegraphics[width=0.062\textwidth]{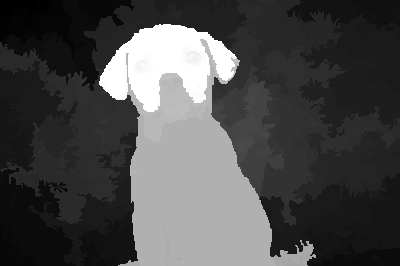} & \includegraphics[width=0.062\textwidth]{figs/0057_ulr.png} &
    \includegraphics[width=0.062\textwidth]{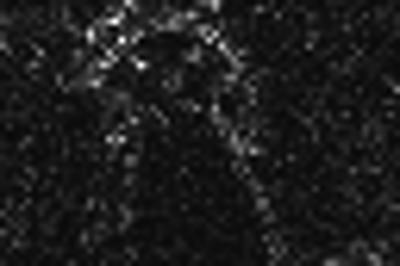} & \includegraphics[width=0.062\textwidth]{figs/0057_slr.png} &
    \includegraphics[width=0.062\textwidth]{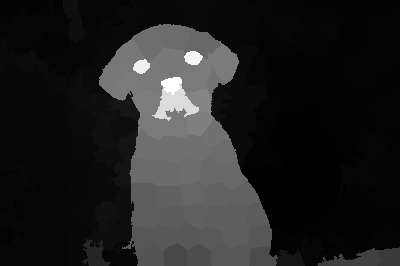} &
    \includegraphics[width=0.062\textwidth]{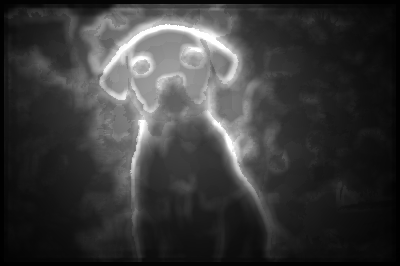} &
    \includegraphics[width=0.062\textwidth]{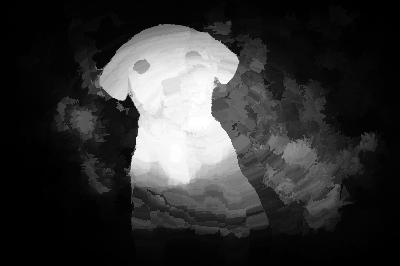} &
    \includegraphics[width=0.062\textwidth]{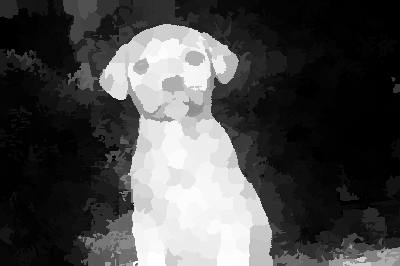} &
    \includegraphics[width=0.062\textwidth]{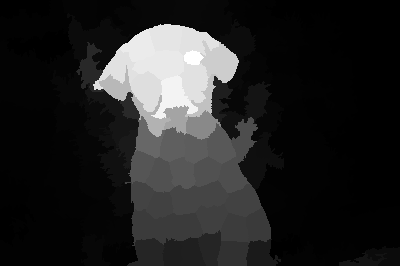} &
    \includegraphics[width=0.062\textwidth]{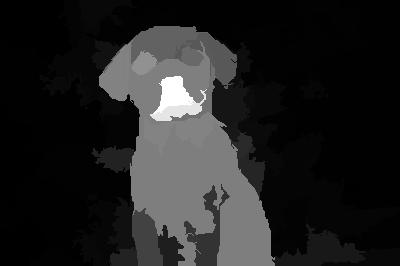} &
    \includegraphics[width=0.062\textwidth]{figs/0057_smd.png} &
    \includegraphics[width=0.062\textwidth]{figs/0057_lrbl.png} &
    \includegraphics[width=0.062\textwidth]{figs/0057_gt.png} \\

    \includegraphics[width=0.062\textwidth]{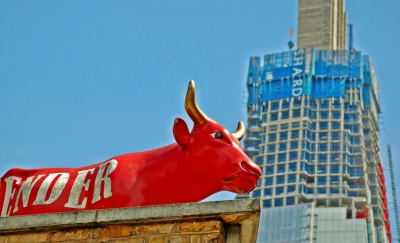} & \includegraphics[width=0.062\textwidth]{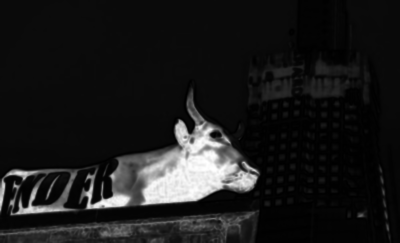} & \includegraphics[width=0.062\textwidth]{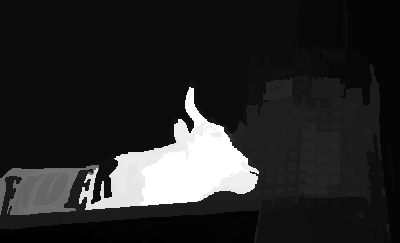} & \includegraphics[width=0.062\textwidth]{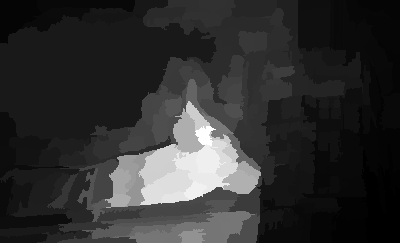} &
    \includegraphics[width=0.062\textwidth]{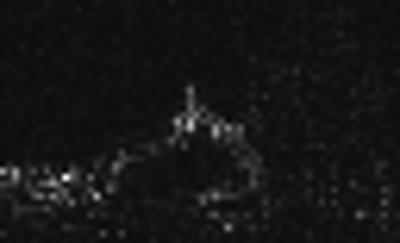} & \includegraphics[width=0.062\textwidth]{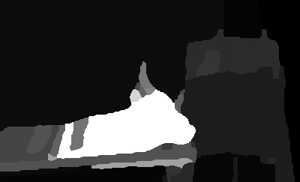} &
    \includegraphics[width=0.062\textwidth]{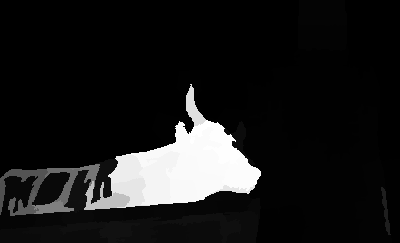} &
    \includegraphics[width=0.062\textwidth]{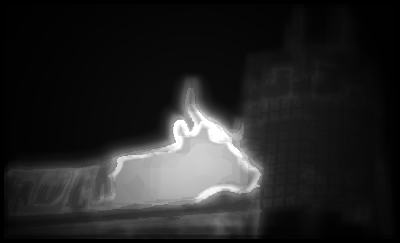} &
    \includegraphics[width=0.062\textwidth]{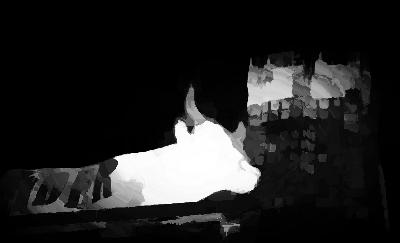} &
    \includegraphics[width=0.062\textwidth]{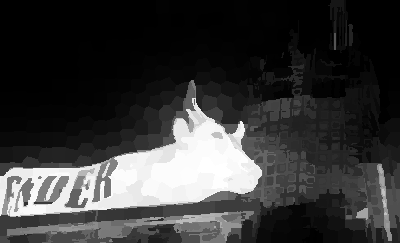} &
    \includegraphics[width=0.062\textwidth]{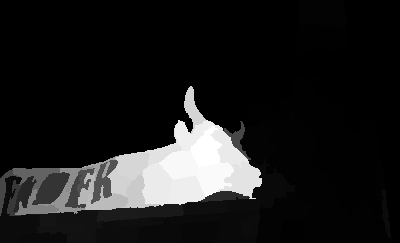} &
    \includegraphics[width=0.062\textwidth]{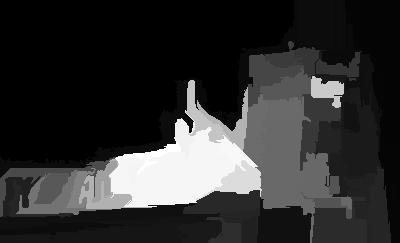} &
    \includegraphics[width=0.062\textwidth]{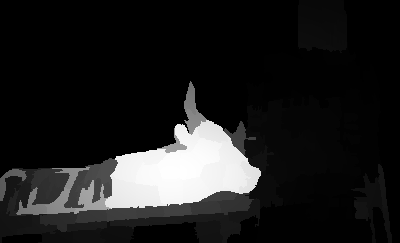} &
    \includegraphics[width=0.062\textwidth]{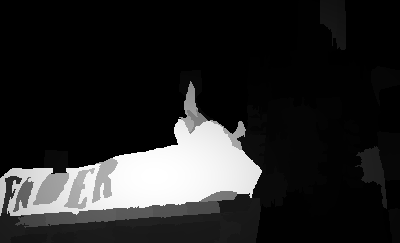} &
    \includegraphics[width=0.062\textwidth]{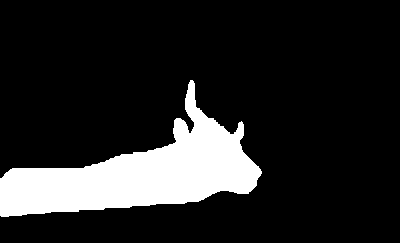} \\

    \includegraphics[width=0.062\textwidth]{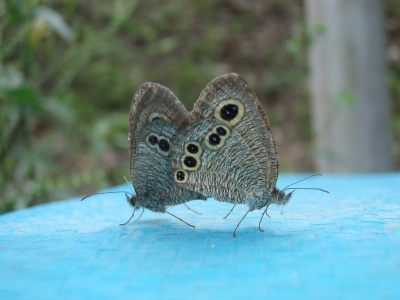} & \includegraphics[width=0.062\textwidth]{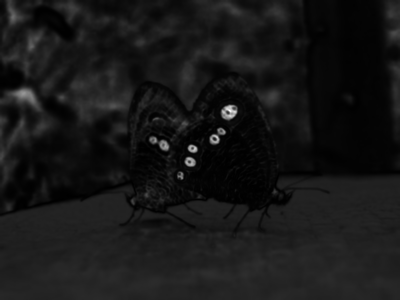} & \includegraphics[width=0.062\textwidth]{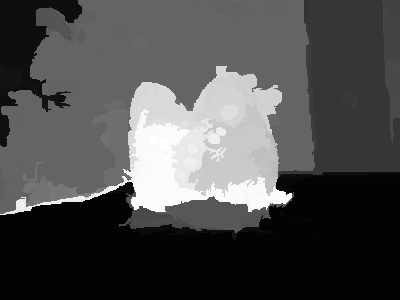} & \includegraphics[width=0.062\textwidth]{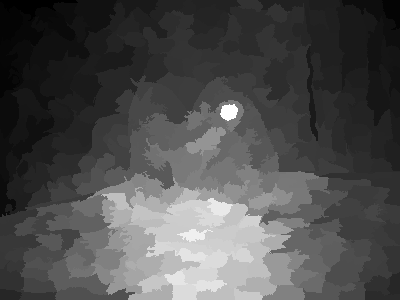} &
    \includegraphics[width=0.062\textwidth]{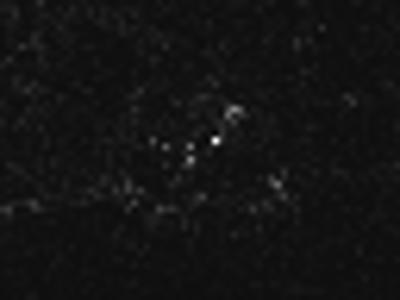} & \includegraphics[width=0.062\textwidth]{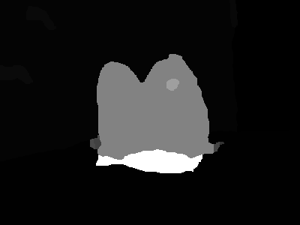} &
    \includegraphics[width=0.062\textwidth]{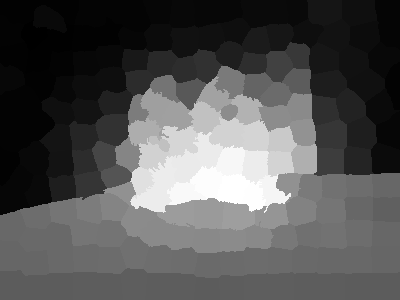} &
    \includegraphics[width=0.062\textwidth]{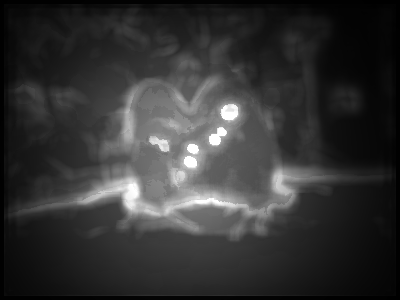} &
    \includegraphics[width=0.062\textwidth]{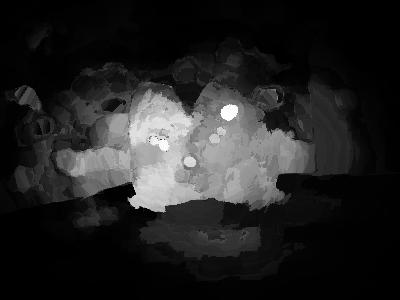} &
    \includegraphics[width=0.062\textwidth]{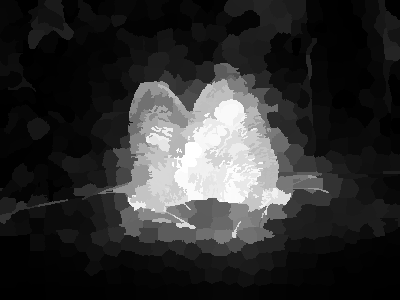} &
    \includegraphics[width=0.062\textwidth]{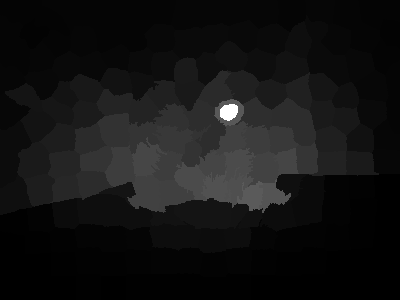} &
    \includegraphics[width=0.062\textwidth]{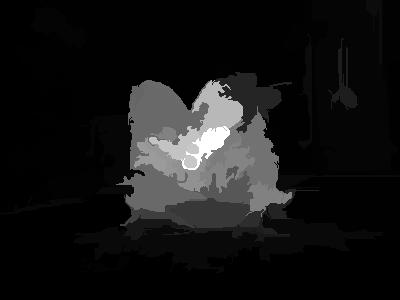} &
    \includegraphics[width=0.062\textwidth]{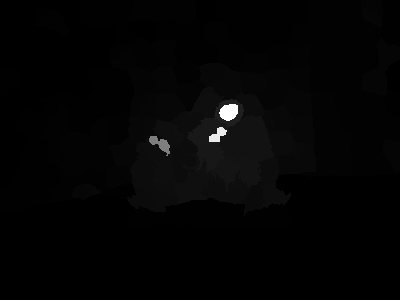} &
    \includegraphics[width=0.062\textwidth]{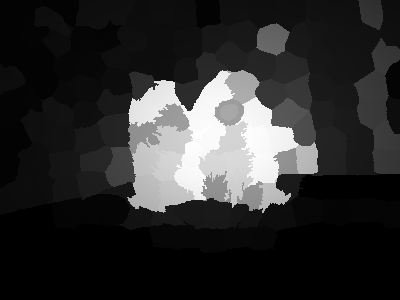} &
    \includegraphics[width=0.062\textwidth]{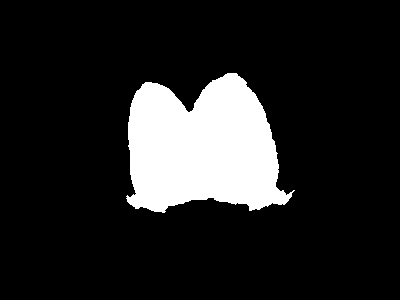} \\

    \includegraphics[width=0.062\textwidth]{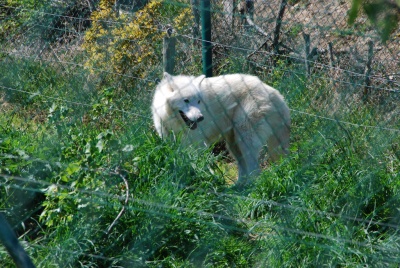} & \includegraphics[width=0.062\textwidth]{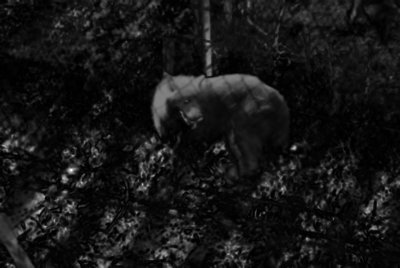} & \includegraphics[width=0.062\textwidth]{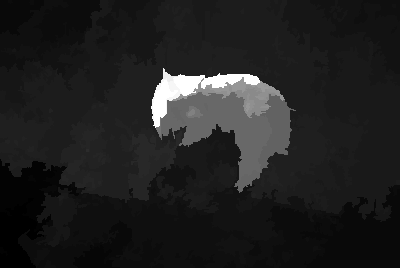} & \includegraphics[width=0.062\textwidth]{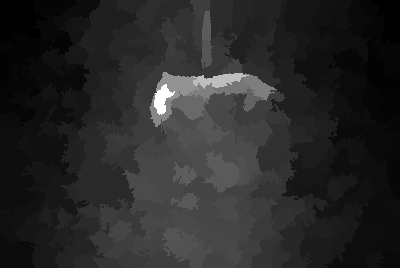} &
    \includegraphics[width=0.062\textwidth]{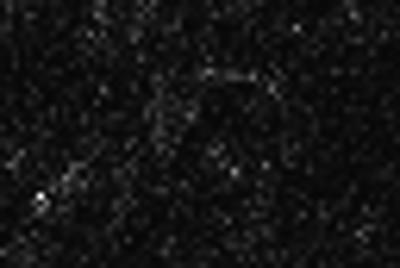} & \includegraphics[width=0.062\textwidth]{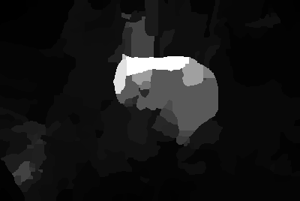} &
    \includegraphics[width=0.062\textwidth]{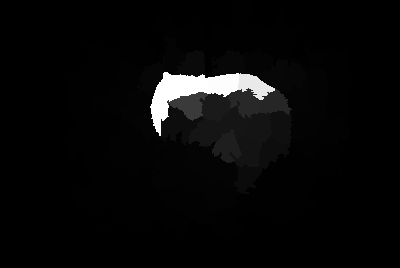} &
    \includegraphics[width=0.062\textwidth]{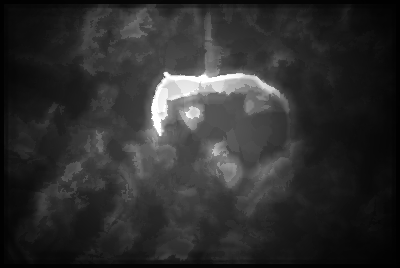} &
    \includegraphics[width=0.062\textwidth]{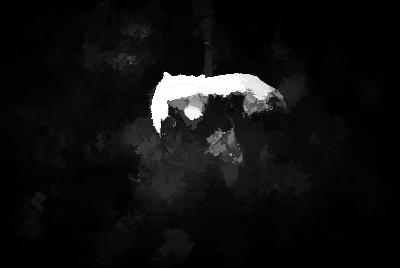} &
    \includegraphics[width=0.062\textwidth]{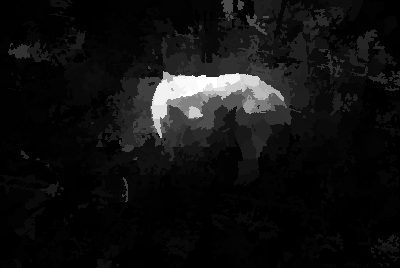} &
    \includegraphics[width=0.062\textwidth]{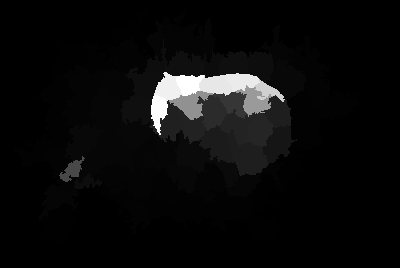} &
    \includegraphics[width=0.062\textwidth]{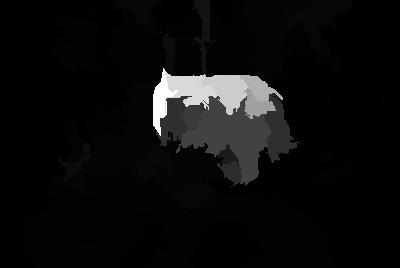} &
    \includegraphics[width=0.062\textwidth]{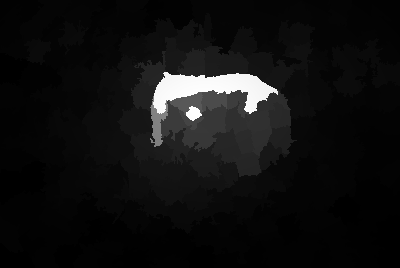} &
    \includegraphics[width=0.062\textwidth]{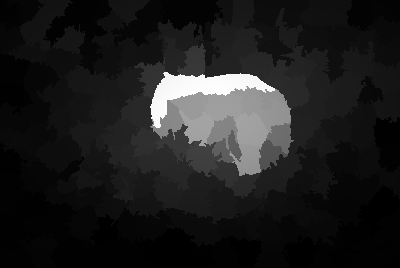} &
    \includegraphics[width=0.062\textwidth]{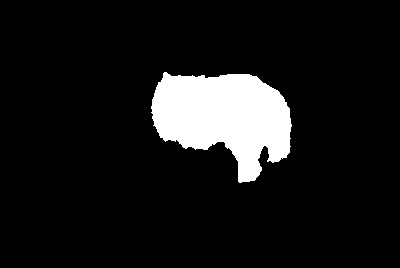} \\

    Image & FT & ULR & SS & HS & SLR & MR & PCA & DSR & HCT & RBD & DRFI & SMD & {\color{red}Ours} & GT

\end{tabular}
\caption{Visible comparison of saliency maps generated by different methods. We select six images from the MSRA10K dataset, four from the iCoSeg dataset and four from the ECSSD dataset, which are arranged sequentially.}
\label{fig:vis}
\end{figure*}

To have an intuitive concept of the performance, we provide a visual comparison of detection result with images selected from the three benchmark datasets, which are diverse in object size, complexity of background and number of objects, as listed in Fig.~\ref{fig:vis}. We can see that our method works well under most cases, and is capable of providing a relatively entire detection. As analyzed above, frequency-tuned method FT~\cite{achanta2009frequency} tends either to filter out part of object or to preserve part of background. Basic low-rank matrix recovery methods like SLR~\cite{zou2013segmentation} and ULR~\cite{shen2012unified} are not robust enough to background and fail to provide a uniform saliency map. Approaches depending on prior cues such as HC~\cite{yan2013hierarchical}, HCT~\cite{kim2016salient}, MR~\cite{yang2013saliency} and RBD~\cite{zhu2014saliency} are more likely to miss object parts that are adjacent to image boundary. Finally, time consumption for all methods is provided in Table~\ref{tab:time}, which demonstrates the efficiency of our method.

\setlength{\tabcolsep}{0.5pt}
\begin{table*} [!ht]
    \centering
    \caption{Average time consumption for each method to process an image in MSRA10K dataset.}
    \begin{tabular}{c|c|c|c|c|c|c|c|c|c|c|c|c|c}
        \hline
        Methods ~& \textbf{Ours} & SMD & DRFI & RBD & HCT & DSR & PCA & MR & SLR & SS & ULR & HS & FT

        \\ \hline
        Time(s) & 0.83 & 1.59 & 9.06 & 0.20 & 4.12 & 10.2 & 4.43 & 1.84 & 22.80 & 0.05 & 15.62 & 0.53 & 0.07 \\ \hline
        Code    &  M+C & M+C  & M+C  & M+C  & M    & M+C  & M+C  & M+C  & M+C   & M    & M+C   & EXE  & C \\ \hline
    \end{tabular}
    \label{tab:time}
\end{table*}

\setlength{\tabcolsep}{1pt}
\begin{figure*}[!ht]
    \centering
    \begin{tabular}{ccc}
        \includegraphics[width=0.32\textwidth]{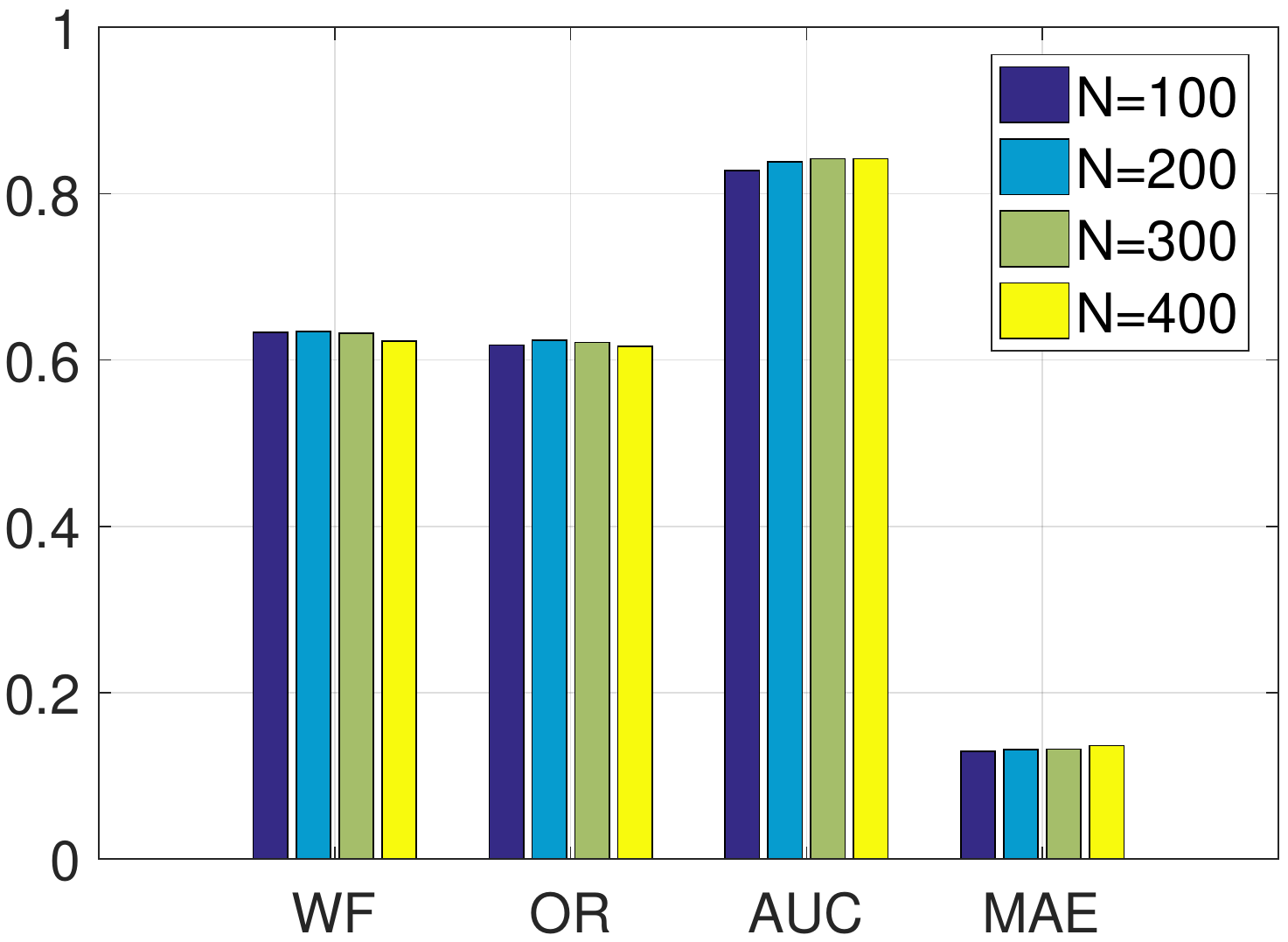} & \includegraphics[width=0.32\textwidth]{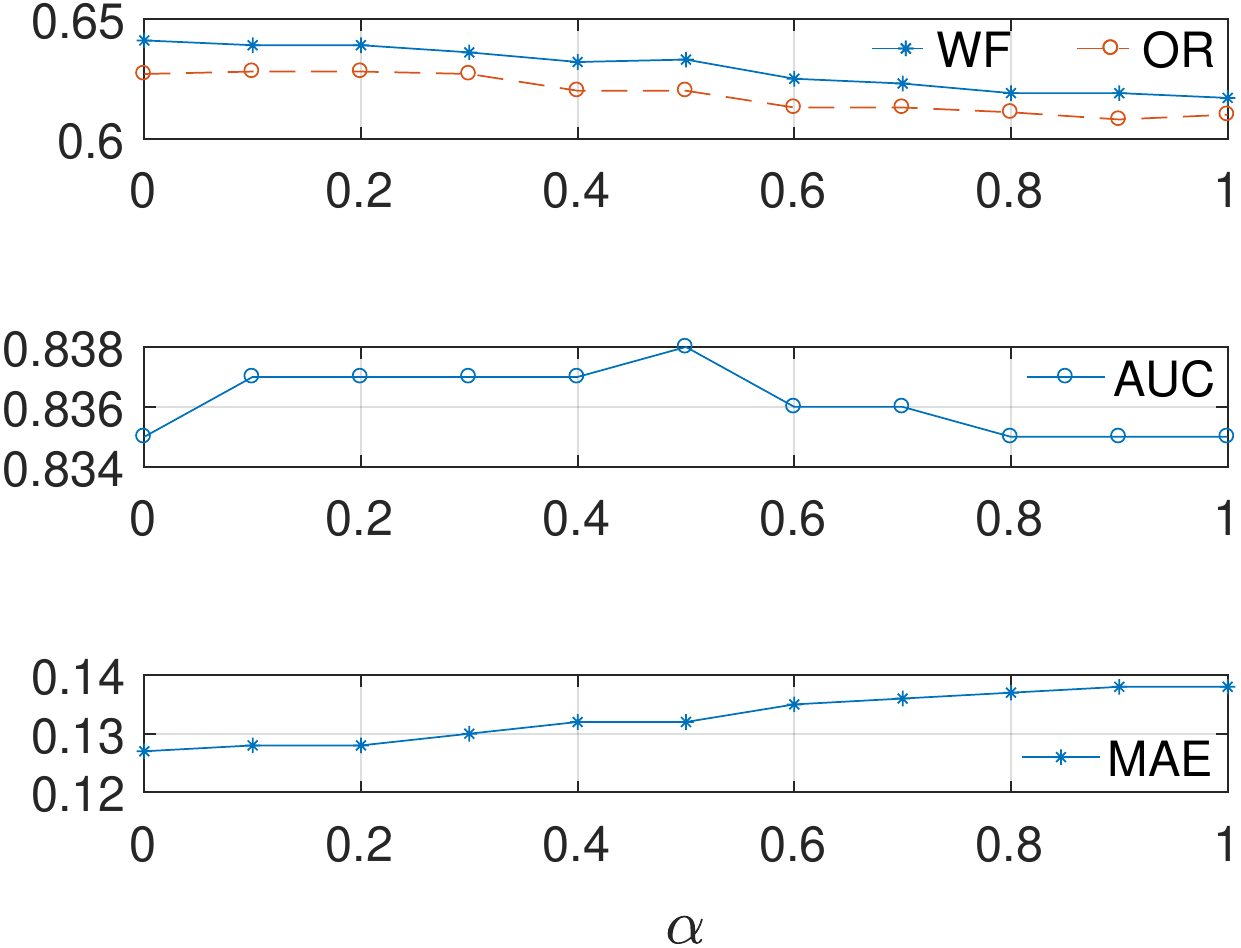} & \includegraphics[width=0.32\textwidth]{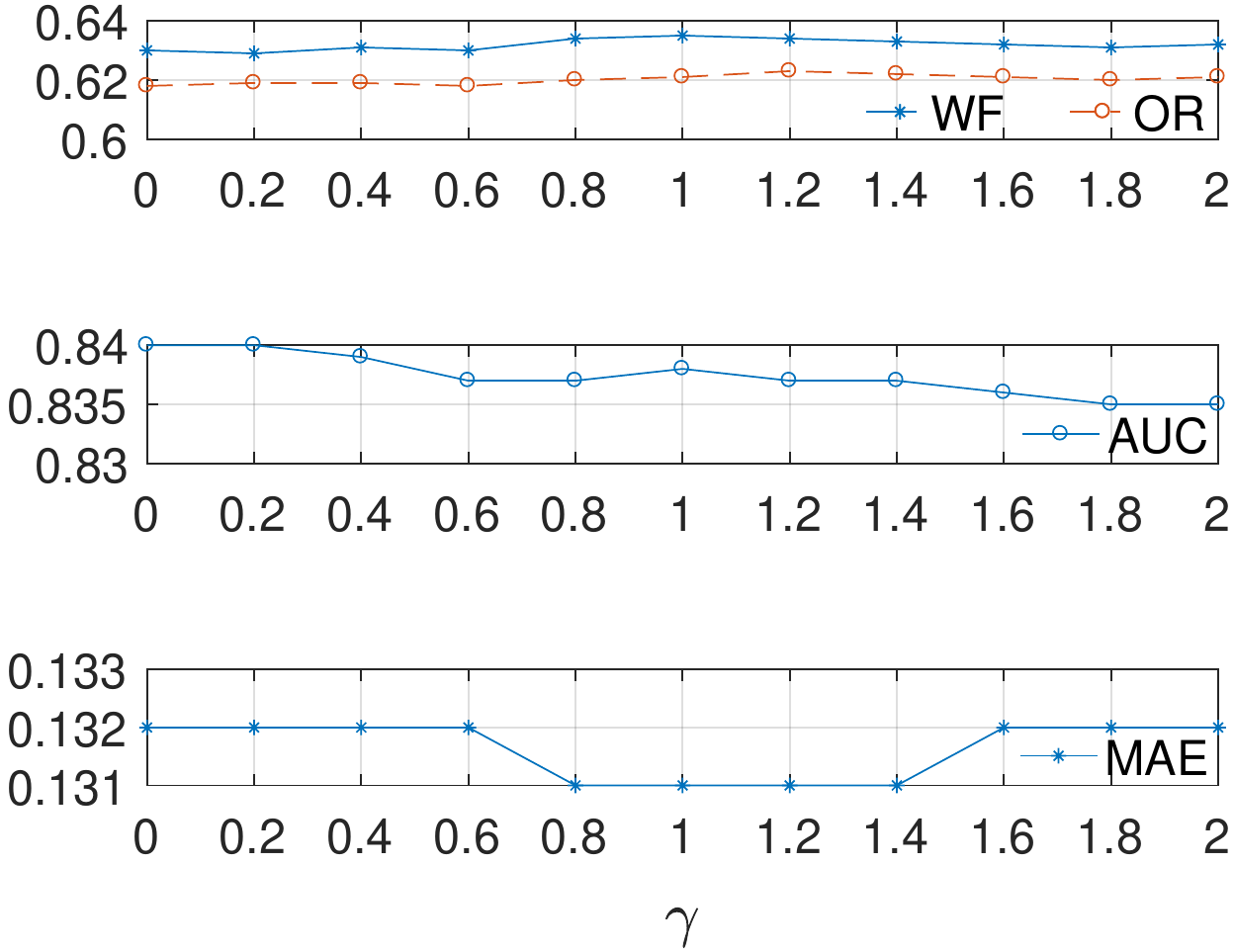}\\
        (a) & (b) & (c)
    \end{tabular}
    \caption{Parametric sensitivity analysis: (a) shows the variation of WF, OR, AUC, MAE w.r.t. $N$ by fixing $\alpha=0.35,~\gamma=1.1$. (b) shows the variation of WF, OR, AUC, MAE w.r.t. $\alpha$ by fixing $N=200,~\gamma=1.1$. (c) shows the variation of WF, OR, AUC, MAE w.r.t. $\gamma$ by fixing $N=200,~\alpha=0.35$.}
    \label{fig:sensi_c}
\end{figure*}

\setlength{\tabcolsep}{1pt}
\begin{figure*}[!ht]
    \centering
    \begin{tabular}{ccc}
        \includegraphics[width=0.32\textwidth]{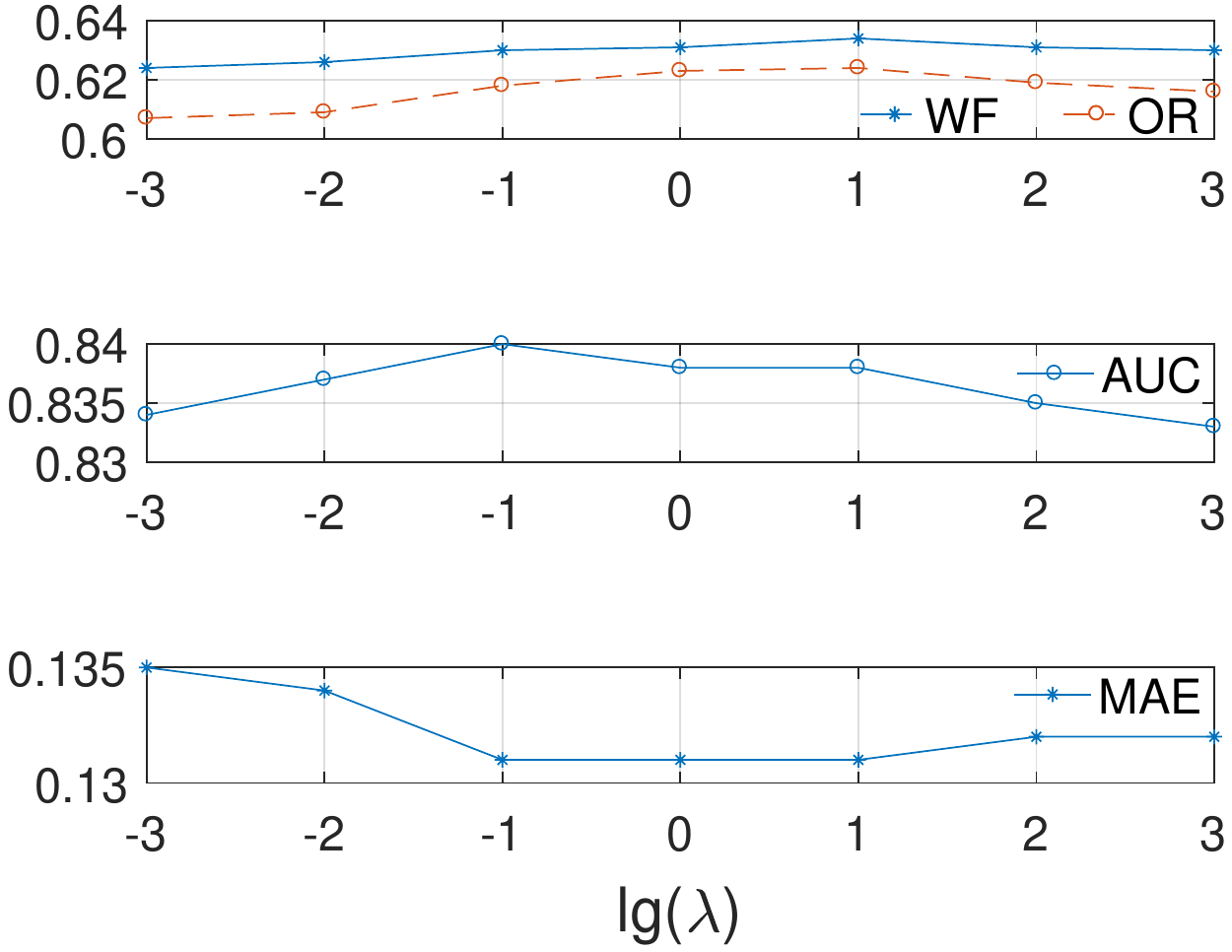} & \includegraphics[width=0.32\textwidth]{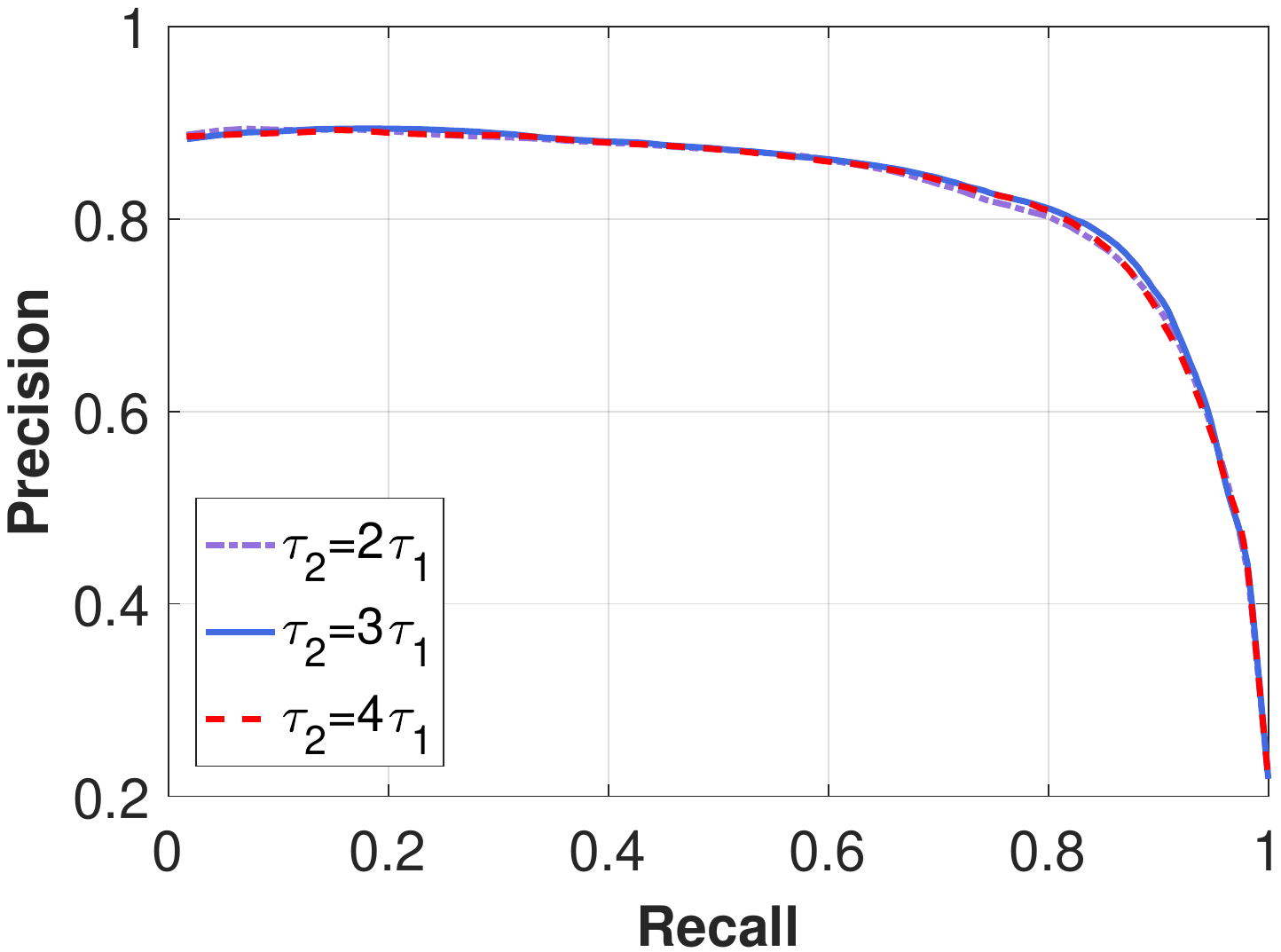} & \includegraphics[width=0.32\textwidth]{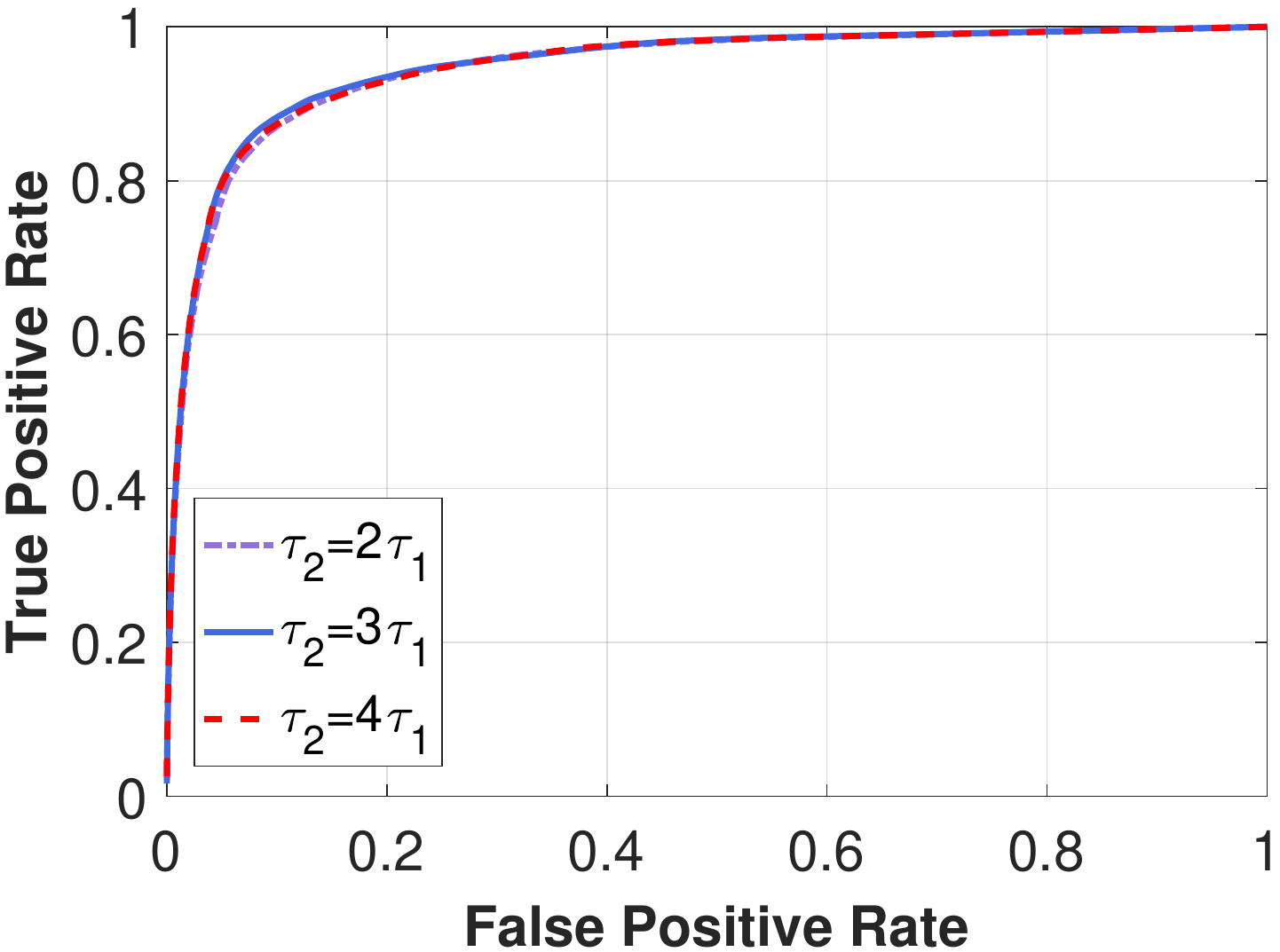}\\
        (a) & (b) & (c)
    \end{tabular}
    \caption{Parametric sensitivity analysis: (a) shows the variation of WF, OR, AUC, MAE w.r.t. $\lambda$. (b) shows the PR curve of different thresholding strategies. (c) shows the ROC curve of different thresholding strategies.}
    \label{fig:sensi_f}
\end{figure*}

\subsection{Analysis of Parameters} \label{sec:param_sensi}

\subsubsection{Parameters in coarse module}
In our coarse module, the algorithm takes three parameters, i.e., the number of super-pixels $N$ in over-segmentation, regularization parameters $\alpha,~\gamma$. We examine the sensitivity of our model to changes of $N,~\alpha,~\gamma$ on iCoSeg dataset as an example. The analysis is conducted by tuning one parameter while fixing another two. The performance changes in terms of WF, OR, AUC, MAE are shown in Fig.~\ref{fig:sensi_c}. For $N$, we observe that similar results are achieved by varying $N$ and $N=200$ is a good trade-off between efficiency and performance, as larger $N$ requires more expensive computation. Besides, we observe that when $\gamma$ is fixed ($\gamma=1.1$), the WF, OR and MAE performance decreases while the AUC performance initially increases, spikes within a range of $\alpha$ from $0.4$ to $0.5$, and then decreases. Thus, we choose the optimal $\alpha=0.35$. When $\alpha$ is fixed ($\alpha=0.35$), the WF and OR performance initially increases, spikes within a range of $\gamma$ from $0.8$ to $1.2$. The AUC performance initially maintains and then decreases, and the MAE performance initially maintains, increases within a range of $\gamma$ from $0.6$ to $1.4$, and then decreases. Thus, we choose the optimal $\gamma=1.1$.

\subsubsection{Parameters in refining module}
In our fine module, the main parameter is the regularization parameter $\lambda$. The sensitivity in terms of WF, OR, AUC, MAE is shown in Fig.~\ref{fig:sensi_f} (a). We observe that the WF, OR performance initially increases, spikes within a range of $\lambda$ from $1$ to $100$, and then decreases. The AUC performance initially increases, spikes within a range of $\lambda$ from $0.01$ to $1$, and then decreases. The MAE performance initially increases, spikes at $0.01$, and then maintains. The results illustrate that compared a small $\lambda$, the model performs worse with a lack of label information from those samples (including both confident ones and tough ones). When $\lambda$ is large, the performance suffers from an obvious drop, which may be caused by over-fitting the confident samples. Therefore, we choose $\lambda=10$ in our method.

Moreover, we also examine the sensitivity of our model to the changes of different thresholding strategies in our refining module. We fix the lower threshold, i.e., we set $\tau_1$ as the average value of coarse saliency, and test varying $\tau_2$. PR curves and ROC curves of $\tau_2=2\tau_1,~\tau_2=3\tau_1$ and $\tau_2=4\tau_1$ are shown in Fig.~\ref{fig:sensi_f} (b) and (c). We observe that our method performs similarly under the three strategies, which demonstrates its robustness.

\section{Conclusion}
\label{sec:conclusion}
In this paper, we present a coarse-to-fine saliency detection architecture that first estimates a coarse saliency map using a novel LRMR model and then refines the obtained coarse saliency map using a learning scheme. Compared with state-of-the-art approaches, our method can efficiently detect salient objects with enhanced object boundaries, even in the scenario of multiple objects. We also show that our fine-tuning scheme can be easily imposed on previous LRMR-based methods to significantly improve their detection accuracy.


\setlength{\tabcolsep}{1.2pt}


\section*{Acknowledgment}
This work was supported in part by the Key Program for International S\&T Cooperation Projects of China (No. 2016YFE0121200), in part by the National Natural Science Foundation of China (No. 61571205, No. 61772220).

\section*{References}
\bibliography{ref}
\end{document}